\documentclass[aps,prb,onecolumn,superscriptaddress,floatfix,notitlepage,11pt,tightenlines]{revtex4-1}
\makeatletter
\AtBeginDocument{\let\LS@rot\@undefined}
\makeatother

\usepackage{color}
\usepackage{dcolumn} 
\usepackage{bm} 
\usepackage{graphicx}
\usepackage{multirow} 
\usepackage{pifont} 
\usepackage{epsfig}
\usepackage{verbatim}
\usepackage{amsfonts}
\usepackage{amsmath} 
\usepackage{amssymb}
\usepackage{subfigure}
\usepackage{mathtools}
\usepackage{braket}
\usepackage{float}
\usepackage{pdfpages}
\usepackage{pdflscape}
\usepackage[version=4]{mhchem}
\usepackage{array}
\usepackage[columnwise]{lineno}
\usepackage{mathrsfs}
\usepackage{epigraph}
\usepackage[mathcal]{euscript}
\usepackage{textcomp}
\usepackage{threeparttable}
\usepackage{enumitem}
\usepackage{caption}
\usepackage{url}
\usepackage{comment}
\usepackage{float}
\usepackage{wrapfig}
\usepackage{xcolor,listings} 
\usepackage[nolist]{acronym}
\usepackage[titletoc,toc,title]{appendix}
\usepackage[bf,raggedright]{titlesec}
\usepackage{textcomp}
\usepackage[left=1.0in,right=1.0in,top=1.0in,bottom=1.0in]{geometry}
\usepackage{indentfirst}	

\newcommand{\etal}{\emph{et~al.}}
\newcommand{\etc}{\emph{etc.}}
\newcommand{\ie}{\emph{i.e.}}

\newcommand{\si}{\text{Supporting Information}}

\newcommand{\al}{\alpha}
\newcommand{\be}{\beta}

\newcommand{\la}{\lambda}

\newcommand{\ze}{\zeta}
\newcommand{\om}{\omega}

\newcommand{\ga}{\gamma}
\newcommand{\sig}{\sigma}

\newcommand{\MAE}{\overline{\text{MAE}}}
\newcommand{\MAEe}{\text{MAE/e}}
\newcommand{\AKEEe}{\text{AKEE/e}}
\newcommand{\MAKEE}{\overline{\text{MAKEE}}}

\newcommand{\NIAD}{\overline{\text{NIAD}}}

\DeclareMathOperator{\erf}{erf}

\renewcommand{\Re}{\operatorname{Re}}

\newcommand{\mathematica}{\textsc{Mathematica}}
\newcommand{\matlab}{\textsc{MATLAB}}

\newcommand{\nvidia}{\textsc{NVIDIA}}

\setenumerate[1]{label=\thesection.\arabic*.}

\newcommand{\mr}{\multirow}

\newcommand*\pFqskip{8mu}
\catcode`,\active
\newcommand*\pFq{\begingroup
	\catcode`\,\active
	\def ,{\mskip\pFqskip\relax}%
	\dopFq
}
\catcode`\,12
\def\dopFq#1#2#3#4{%
	\Phi\biggl[#1\biggl|#2\genfrac..{0pt}{}{#3}{#4}\biggr]%
	\endgroup
}

\begin{document}

\author{Mohammad Mostafanejad}
\email{smostafanejad@vt.edu}
\affiliation{Department of Chemistry, Virginia Tech, Blacksburg, Virginia 24061, USA}
\affiliation{Molecular Sciences Software Institute, Blacksburg, Virginia 24060, USA}

\title{Unification of popular artificial neural network activation functions}

\acresetall	
\begin{abstract}
    We present a unified representation of the most popular neural network activation
    functions. Adopting Mittag-Leffler functions of fractional calculus, we propose a 
    flexible and compact functional form that is able to interpolate between various 
    activation functions and mitigate common problems in training neural networks such 
    as vanishing and exploding gradients. The presented gated representation extends the
    scope of fixed-shape activation functions to their adaptive counterparts whose shape 
    can be learnt from the training data. The derivatives of the proposed functional 
    form can also be expressed in terms of Mittag-Leffler functions making it a suitable
    candidate for gradient-based backpropagation algorithms. By training multiple 
    neural networks of different complexities on various datasets with different sizes,
    we demonstrate that adopting a unified gated representation of activation functions
    offers a promising and affordable alternative to individual built-in implementations
    of activation functions in conventional machine learning frameworks.
\end{abstract}
\maketitle

\acresetall	
\section{Introduction}\label{SEC:INTRODUCTION}
Activation functions are one of the key building blocks in \acp{ANN} that
control the richness of the neural response and determine the accuracy,
efficiency and performance\cite{Clevert:2015:ARXIV} of multilayer neural networks
as universal approximators.\cite{Hornik:1989:359} Due to their
biological links\cite{Abbott:2001:BOOK,Haykin:1999:BOOK} and optimization
performance, saturating activation functions\cite{Gulcehre:2016:4457}
such as logistic Sigmoid and hyperbolic tangent \cite{Costarelli:2013:72}
were commonly adopted in early neural networks. Nevertheless, both activation
functions suffer from the vanishing gradient problem.\cite{Hochreiter:1998:107}
Later studies on image classification using \aclp{RBM}\cite{Nair:2010:807}
and \aclp{DNN}\cite{Glorot:2011:315} demonstrated that \acp{RELU} can mitigate
the vanishing gradient problem and improve the performance of neural networks.
Furthermore, the sparse coding produced by \acp{RELU} not only creates a more robust
and disentangled feature representation but also accelerates the learning
process.\cite{Glorot:2011:315}

The computational benefits and the current popularity of \acp{RELU} should
be taken with a grain of salt due to their notable disadvantages such as bias
shift,\cite{Clevert:2015:ARXIV} ill-conditioned parameter scaling\cite{Glorot:2011:315}
and dying \ac{RELU}.\cite{Lu:2020:1671} Furthermore, the unbounded nature of
\acp{RELU} for positive inputs, while potentially helpful for training \aclp{DNN},
can aggravate the exploding gradient problem in
\aclp{RNN}.\cite{Bengio:1994:157,Pascanu:2013:1310} In order to address the
dying \ac{RELU} and the vanishing/exploding gradient problems, a multitude of \ac{RELU}
variants have been proposed\cite{Maas:2013:JMLR,Qiu:2017:ARXIV,Liu:2019:ARXIV2,
  He:2015:1026,Jin:2015:ARXIV} but none has managed to consistently outperform
the vanilla \acp{RELU} in a wide range of experiments.\cite{Xu:2015:ARXIV} Alternative
activation functions such as \acp{ELU}\cite{Clevert:2015:ARXIV} and \acp{SELU}\cite{Klambauer:2017:972}
have also been proposed to build upon the benefits of \ac{RELU} and its variants
and provide more robustness and resistance to the input noise. Yet, among the
existing slew of activation functions in the literature,\cite{DasGupta:1992:615,
  Apicella:2021:14,Duch:1999:163} no activation function seems to offer global
superiority across all modalities and application domains.

Trainable activation functions,\cite{Chen:1996:627,Guarnieri:1999:672,Piazza:1993:1401,
  Piazza:1992:343} whose functional form is learnt from the training data, offer a more
flexible option than their fixed-shape counterparts. In order be able to fine-tune the
shape of activation functions during backpropagation,\cite{Rumelhart:1986:533}
partial derivatives of activation functions with respect to unknown learning parameters
are required. It is important to note that some trainable activation functions can also
be replaced by simpler multilayer feed-forward subnetworks with constrained parameters and
classical fixed-shape activation functions.\cite{Apicella:2021:14} The ability to
replace a trainable activation function with a simpler sub-neural network highlights
a deep connection between the choice of activation functions and performance of
neural networks. As such, pre-setting the best possible trainable activation function
parameters or fine-tuning the experimental settings\cite{Smith:2018:ARXIV} such as
data preprocessing methods, gradient and weight clipping,\cite{Pascanu:2013:1310}
optimizers,\cite{Polyak:1964:1,Nesterov:1983:543,Duchi:2011:2121,Hinton:2012:Coursera,
Kingma:2014:ARXIV} regularization methods such as $L_1$, $L_2$ and drop
out,\cite{Hinton:2012:ARXIV} \ac{BN},\cite{Ioffe:2015:448} learning rate
scheduling,\cite{Smith:2018:ARXIV,Senior:2013:6724} (mini-)batch size, or network
design\cite{Hagan:2014:BOOK} variables such as depth (number of layers) and width
(number of neurons per layer) of the neural network as well as weight initialization
methods\cite{Glorot:2010:249,He:2015:1026,Montavon:2012:BOOK} becomes an important
but challenging task. Several strategies such as neural architecture
search\cite{Liu:2019:ARXIV} and network design space design\cite{Radosavovic:2020:arxiv}
have been proposed to assist the automation of the network design\cite{Hagan:2014:BOOK}
process but they have to deal with an insurmountable computational cost barrier
for practical applications.

In this manuscript, we take a theoretical neuroscientific standpoint\cite{Abbott:2001:BOOK}
towards activation functions by emphasizing the existing connections among them from
a mathematical perspective. As such, we resort to the expressive power of rational
functions as well as higher transcendental special functions of fractional calculus to
propose a unified gated representation of activation functions. The presented functional
form is conformant with the outcome of a semi-automated search, performed by
Ramachandran \etal,\cite{Ramachandran:2017:ARXIV} in order to find the optimal
functional form of activation functions over a pre-selected set of functions. The
unification of activation functions offers several significant benefits: It requires
fewer lines of code to be implemented and leads to less confusion in dealing with a
wide variety of empirical guidelines on activation functions because individual
activation functions correspond to special parameter sets in the gated representation.
The proposed functional form is closed under differentiation making it a suitable choice
for an efficient implementation of backpropagation algorithms commonly used for
training \acp{ANN}. Finally, the unified functional can be adopted as a fixed-shape
or trainable activation function or both when training neural networks. In other words,
one can access different activation functions or interpolate between them by fixing
or varying a set of parameters in the gated functional representation, respectively.

The manuscript is organized as follows: In Sec.~\ref{SEC:THEORY}, we introduce
Mittag-Leffler functions of one- and two-parameters and discuss their important
analytical and numerical properties. Next, we use Mittag-Leffler functions to
create a gated representation that can unify a set of most commonly used activation
functions. Section~\ref{SEC:COMPDETAILS} delineates the computational details of
our experiments presented in Sec.~\ref{SEC:RESULTS}, where we provide numerical
evidence for the efficiency and accuracy of the proposed functional form.
Concluding remarks and future directions are presented in Sec.~\ref{SEC:CONCLUSION}.

\section{Theory}\label{SEC:THEORY}
\subsection{Mittag-Leffler functions of one- and two-parameters}
Mittag-Leffler functions, sometimes referred to as ``\textit{the queen of functions
in fractional calculus}'',\cite{Mainardi:2007:269,Mainardi:2020:1359} are one of the most
important higher transcendental functions that play a fundamental role in fractional
calculus.\cite{Samko:1993:BOOK,2019:FCHANDBOOK,Podlubny:1999:BOOK} The interested reader
is referred to Refs.~\citenum{2019:FCHANDBOOK}, \citenum{Herrmann:2018:BOOK} and
\citenum{Gorenflo:2020:BOOK} for a survey of scientific and engineering applications.
The one-parameter Mittag-Leffler function is defined
as\cite{Gorenflo:2020:BOOK,Kochubei:2019a:BOOK}
\begin{equation}\label{EQ:MLF1}
  E_\al(z) = \sum_{k=0}^{\infty} \frac{z^k}{\Gamma(\al k + 1)},
  \qquad  \al \in \mathbb{C},
\end{equation}
where $\mathbb{C}$ denotes the set of complex numbers. For all values of $\Re(\al)>0$, the
series in Eq.~\ref{EQ:MLF1} converges everywhere in the complex plane and the one-parameter
Mittag-Leffler function becomes an entire function of the complex variable
$z$.\cite{Gorenflo:2020:BOOK} However, when $\Re(\al)<0$, the series in Eq.~\ref{EQ:MLF1}
diverges everywhere on $\mathbb{C}\ \! \backslash\ \! \lbrace0\rbrace$. As
$\al \rightarrow 0^+$, the Mittag-Leffler function can be expressed as\cite{Gorenflo:2020:BOOK}
\begin{equation}\label{EQ:GEOMETRIC}
  E_{0}(\pm z) \sim \frac{1}{1 \mp z}, \qquad |z| < 1.
\end{equation}
Although Mittag-Leffler series has a finite radius of convergence, the restriction
in Eq.~\ref{EQ:GEOMETRIC} can be lifted and the asymptotic geometric series form can be
adopted as a part of the definition of Mittag-Leffler function for
$\al=0$\cite{Berberan-Santos:2005:265} The aforementioned definition seems
to be consistent with the implementation of Mittag-Leffler function in \mathematica\
13.2.\cite{Mathematica:2022:SOFTWARE} Note that for $x>0$ and $0\leq \al \leq 1$,
the one-parameter Mittag-Leffler function with negative arguments, $E_\al(-x)$,
is a completely monotonic\cite{Pollard:1948:1115} function with no real
zeros.\cite{Gorenflo:2020:BOOK}

The two-parameter Mittag-Leffler function can be similarly defined as
\begin{equation}\label{EQ:MLF2}
  E_{\al,\be}(z) = \sum_{k=0}^{\infty} \frac{z^k}{\Gamma(\al k + \be)},
  \qquad  \text{where} \quad \Re(\al) > 0, \quad \text{and} \quad \be \in \mathbb{C}.
\end{equation}
The exponential form of Mittag-Leffler function, $E_1(z) = E_{1,1}(z)$, has no zeros
in the complex plane. Nonetheless, for all $m\in\mathbb{N}$, where $\mathbb{N}$ is the
set of natural numbers, $E_{1,-m}$ has its only $m+1$-order zero located at $z=0$. All
zeros of $E_2(z)$ are simple and can be found on the negative real semi-axis. For a more
detailed discussion on the distribution of zeros and the asymptotic properties of
Mittag-Leffler functions, see Ref.~\citenum{Gorenflo:2020:BOOK}.

Parallel to the study of analytic properties, the realization of accurate and
efficient numerical methods for calculating Mittag-Leffler functions is still an open and
active area of research.\cite{Karniadakis:2019:BOOK,Gorenflo:2020:BOOK} In particular, the
existence of free, open-source and accessible software for computing Mittag-Leffler
functions is key to their usability in practical applications. We must note that the code
base and programmatic details of recent updates to the implementation of Mittag-Leffler
functions in \mathematica\ are not publicly available for further analysis in this
manuscript. Nonetheless, several open-source modules for numerical computation of
Mittag-Leffler functions are available in the public domain. Gorenflo
\etal\cite{Gorenflo:2002:491} have proposed an algorithm for computing two-parameter
Mittag-Leffler functions that is suitable for use in \mathematica. Podlubny's algorithm
is implemented in \matlab\cite{MATLAB:2022:SOFTWARE} which allows the computation of 
Mittag-Leffler functions with arbitrary accuracy.\cite{Podlubny:2012:MATHWORKS} Garrappa 
has proposed an efficient method for calculating one- and two-parameter Mittag-Leffler 
functions using hyperbolic path integral transform and quadrature.\cite{Garrappa:2015:1350}
Both \matlab\cite{Garrappa:2015:MATHWORKS} and Python\cite{Hinsen:2017:GITHUB} implementations
of Garrappa's algorithm are also available in the public domain. Zeng and Chen have also
constructed global Pad\'e approximations for the special cases of parameters 
$0 < \al \leq 1$ and $\be \geq \al$, based on the complete monoticity of 
$E_{\al,\be}(-x)$.\cite{Zeng:2015:1492} Another powerful feature of Mittag-Leffler functions
is their relation to other higher transcendental special functions such as hypergeometric,
Wright, Meijer $G$ and Fox $H$-functions\cite{Gorenflo:2020:BOOK,Olver:2010:NISTBOOK,
Mathai:1973:BOOK,Abramowitz:1964:BOOK,Mathai:2010:BOOK} which allows for more general 
analytic and efficient numerical computations. For example, \mathematica\ automatically 
simplifies the one-parameter Mittag-Leffler functions with non-negative (half-)integer
$\al$ to (sum of) generalized hypergeometric functions.\cite{Wolfram:2022:MLF}

In addition to the algorithm complexity and implementation specifics, the total number of
activation functions in a neural network can strongly affect its runtime on computing
accelerators such as \acp{GPU}. The neural network architecture\cite{Hagan:2014:BOOK} is
also a major factor in determining the computational cost.\cite{Radosavovic:2020:arxiv} We
will consider the impact of these factors in our numerical experiments in Sec.~\ref{SEC:RESULTS}.

\subsection{Gated representation of activation functions}
In order to unify the most common classical fixed-shape activation functions,
listed in a recent survey,\cite{Apicella:2021:14} we propose the following functional
form
\begin{equation}\label{EQ:FUNC}
  x \pFq{x}{\gamma}{\al_1\ \be_1\ f}{\al_2\ \be_2\ g} := x \bigg\lbrace  x ^{\ga-1}
  \left( \frac{E_{\al_1,\be_1}\left[f(x)\right]}{E_{\al_2,\be_2}\left[g(x)\right]} \right) \bigg\rbrace,
\end{equation}
where the gate function, $\Phi[f(x), g(x)]$, is a binary composition of two ``well-behaved''
functional mappings $f, g: \mathbb{R} \rightarrow \mathbb{R}$ and is responsible for
generating a (non-)linear neural response. Here, $\mathbb{R}$ denotes the set of real numbers.
The gated representation in Eq.~\ref{EQ:FUNC} incorporates the functional form obtained from
an automated search over a set of a pre-selected functions\cite{Ramachandran:2017:ARXIV}
and is consistent with the functional form of popular activation functions such as
\ac{RELU} and Swish. Throughout this manuscript, we restrict ourselves to
$\ga \ge 0$, $\Re(\al) > 0$ and $\be \in \mathbb{R}$.

Table \ref{TAB:PARAMS} presents a shortlist of popular fixed-shape classical activation
functions that are accessible to the proposed gated representation as special cases
via different sets of parameters.

\begin{table*}[!htbp]
  \centering
  \setlength{\tabcolsep}{3pt}
  \setlength{\extrarowheight}{1pt}
  \caption{Special cases of gate function $\Phi$ in Eq.~\ref{EQ:FUNC}}
  \label{TAB:PARAMS}
  \begin{tabular}{lcccccccc}
    \hline\hline
    Activation Function            & Argument      & $\gamma$ & $\al_1$ & $\al_2$ & $\be_1$ & $\be_2$ & $f(x)$        & $g(x)$   \\
    \hline
    ReLU$^a$                       & $x$           & 1        & 1       & 1       & 1       & 1       & 0             & 0        \\
    Sigmoid                        & $x$           & 0        & 0       & 1       & 1       & 1       & $-e^{-x}$     & 0        \\
    Swish                          & $x$           & 1        & 0       & 1       & 1       & 1       & $-e^{-c x}$   & 0        \\
    Softsign                       & $x$           & 1        & 0       & 1       & 1       & 1       & $-|x|$        & 0        \\
    Hyperbolic Tangent             & $x$           & 1        & 2       & 2       & 2       & 1       & $x^2$         & $x^2$    \\
    Mish                           & $\log(1+e^x)$ & 2        & 2       & 2       & 2       & 1       & $x^2$         & $x^2$
    \\
    \mr{2}{*}{Bipolar Sigmoid}$^b$ & $x$           & 0        & 0       & 0       & 1       & 1       & $-e^{-x}$     & $e^{-x}$ \\
                                   & $x/2$         & 1        & 2       & 2       & 2       & 1       & $x^2$         & $x^2$    \\
    GELU                           & $x$           & 1        & $1/2$   & 1       & 1       & 1       & $ x/\sqrt{2}$ & $x^2/2$  \\
    \hline\hline
  \end{tabular}
  \begin{tablenotes}
    \scriptsize
    \item \qquad \qquad \qquad \qquad \quad $^a$ The set of parameters are pertinent to $x \ge 0$, otherwise $\Phi=0$.
    \item \qquad \qquad \qquad \qquad \quad $^b$ At least two representations exist for the Bipolar Sigmoid.
  \end{tablenotes}
\end{table*}

\begin{figure}[!htbp]
  \centering
  \subfigure[\hspace{-25pt}]{\includegraphics[scale=0.85]{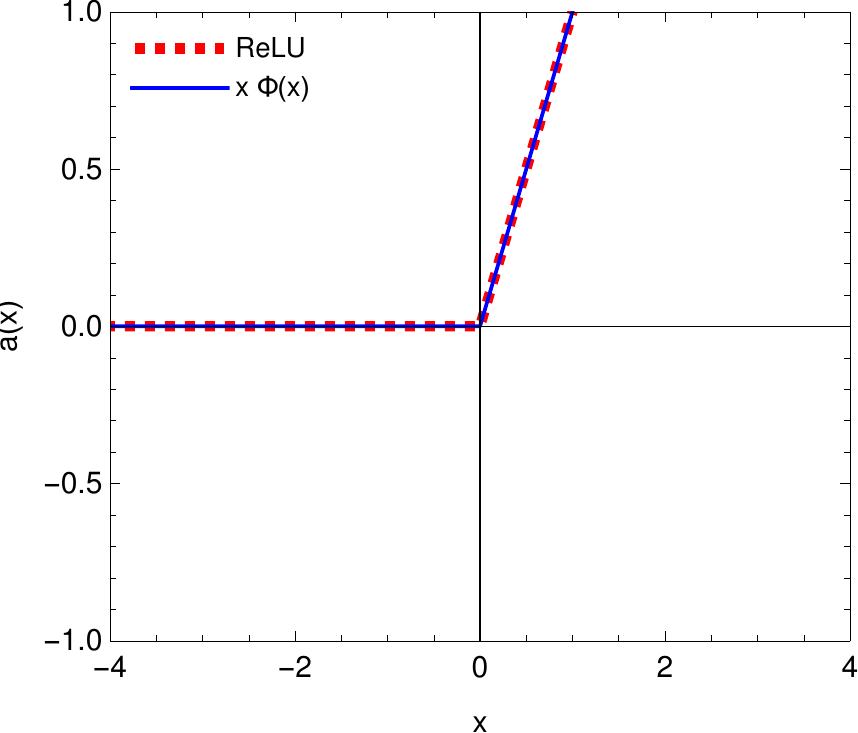}\label{FIG:RELU}}
  \hspace{3pt}
  \subfigure[\hspace{-25pt}]{\includegraphics[scale=0.85]{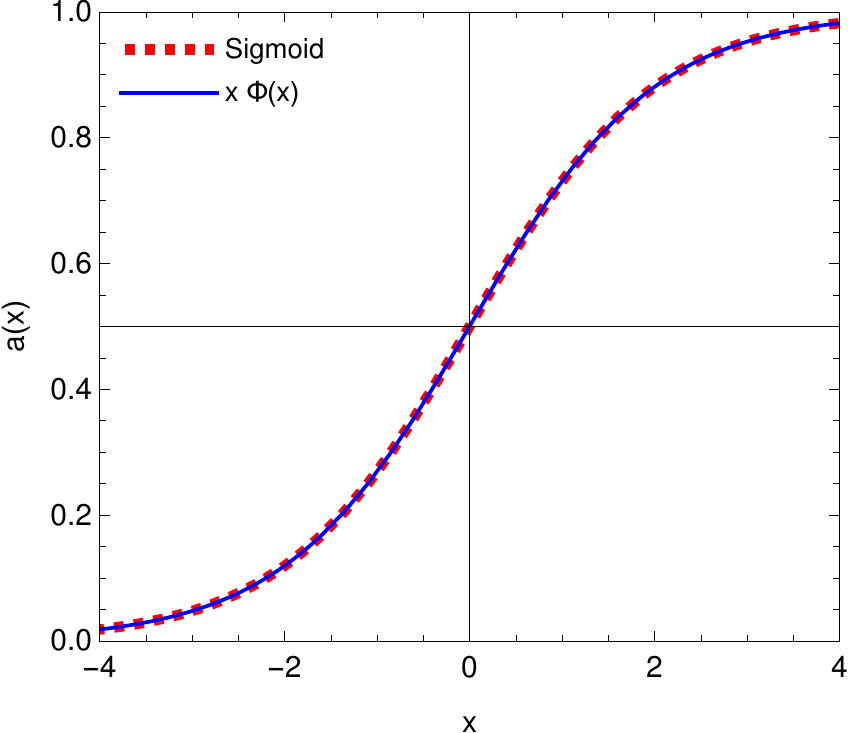}\label{FIG:SIGMOID}}
  \\
  \subfigure[\hspace{-25pt}]{\includegraphics[scale=0.85]{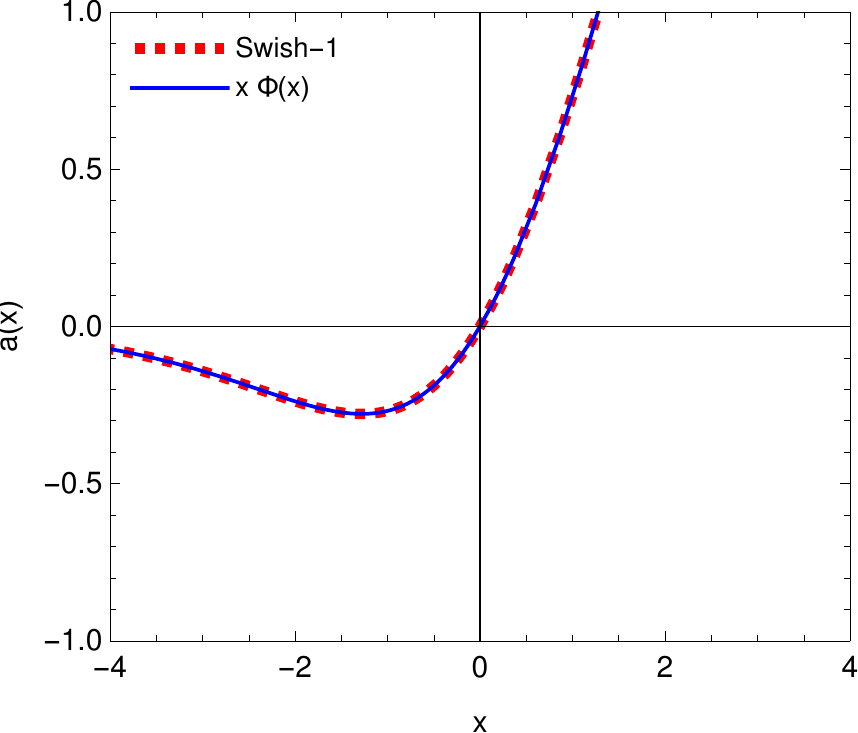}\label{FIG:SWISH1}}
  \hspace{3pt}
  \subfigure[\hspace{-25pt}]{\includegraphics[scale=0.85]{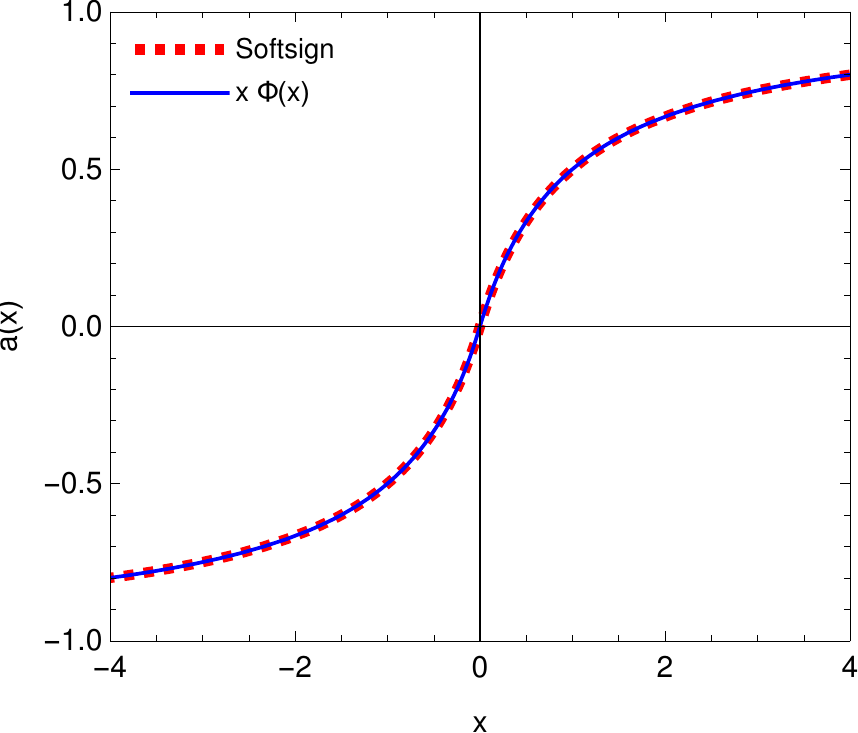}\label{FIG:SOFTSIGN}}
  \\
  \subfigure[\hspace{-25pt}]{\includegraphics[scale=0.85]{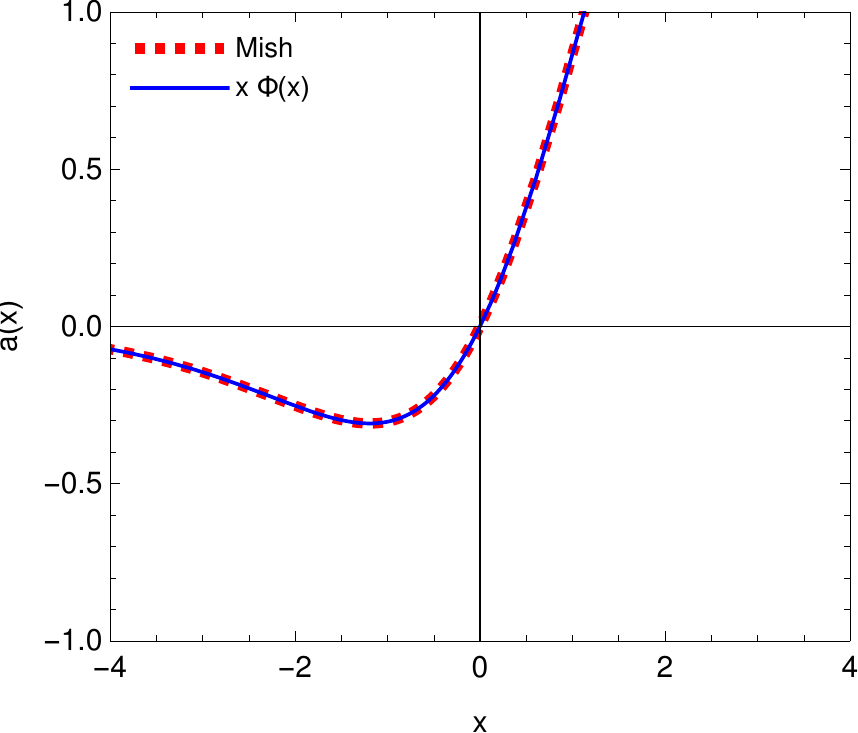}\label{FIG:MISH}}
  \hspace{3pt}
  \subfigure[\hspace{-25pt}]{\includegraphics[scale=0.85]{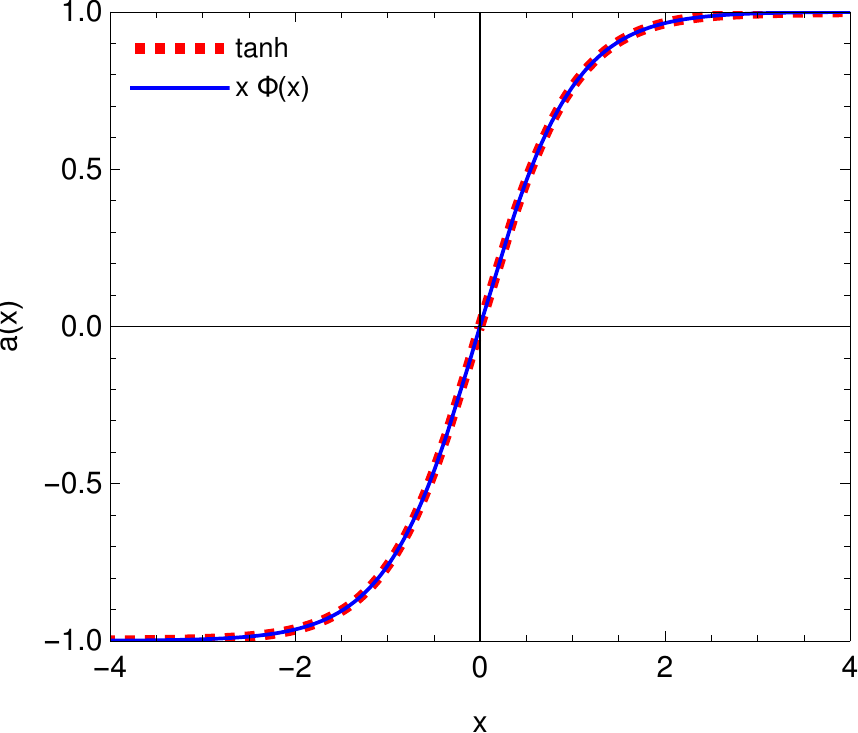}\label{FIG:TANH}}
  \caption{Plots of built-in and gated representation of various activation functions}
\end{figure}

\begin{figure}[!tbph]
  \centering
  \includegraphics{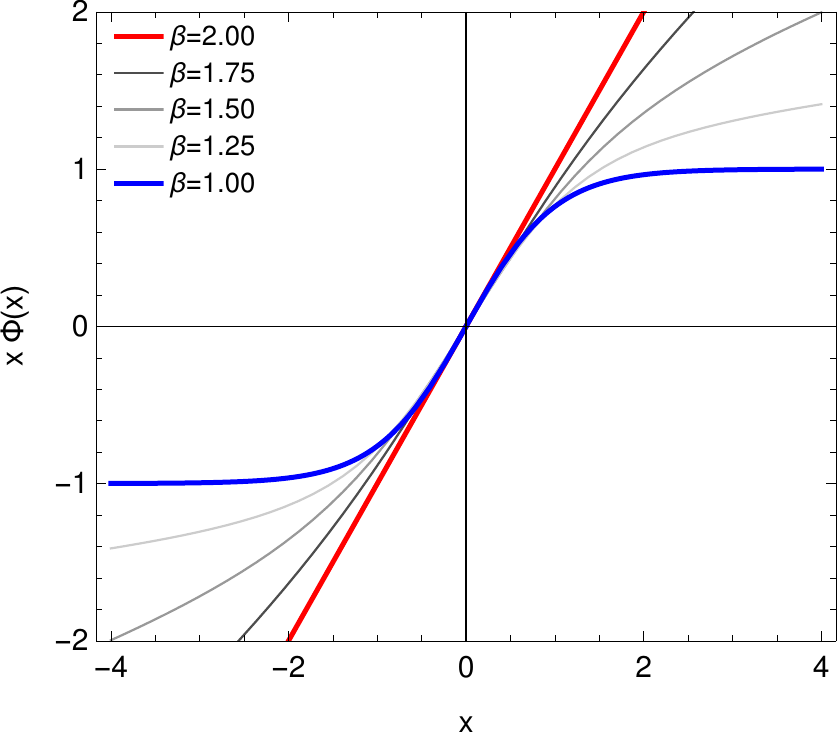}
  \caption{The interpolation of $x \pFq{x}{1}{2~ 2\ x^2}{2\ \be\ x^2}$ between linear and
    hyperbolic tangent functions}
  \label{FIG:FLEX}
\end{figure}

The \ac{RELU} is commonly represented in a piecewise functional form as $\max(0,x)$.
In order to mimic this behavior, the gate function in Eq.~\ref{EQ:FUNC} should reduce
to identity for $x>0$ and zero otherwise. The former condition is satisfied when $\ga=1$
and $E_{\al_1,\be_1}\left[f(x)\right]/E_{\al_2,\be_2}\left[g(x)\right] = 1$, for which
$\pFq{x}{1}{\al\ \be\ f}{\al\ \be\ f} = 1$ is the trivial case. Plots of \ac{RELU}
activation function and its gated representation are shown in Fig.~\ref{FIG:RELU}. Note
that the y-axis label, $a(x)$, collectively refers to activation functions regardless
of their functional form.
The gate functional for Sigmoid activation function, $\sig(x)$, takes $f(x) = -e^{-x}$ and
$g(x) = 0$ to yield
\begin{equation}\label{EQ:FSIGMOID}
  x \Phi\bigg[x \bigg| 0\
    \begin{matrix}
      0 & 1 & -e^{-x} \\
      1 & 1 & 0
    \end{matrix}\bigg] = \frac{1}{1+e^{-x}} = \sig(x).
\end{equation}
Plots of Sigmoid activation function and its gated representation are shown in
Fig.~\ref{FIG:SIGMOID}. The Sigmoid gate function in Eq.~\ref{EQ:FSIGMOID} can be
morphed into that of Swish by setting $\ga = 1$ and $f(x) = e^{-c x}$ to obtain
\begin{equation}\label{EQ:FSWISH}
  x \Phi\bigg[x \bigg| 1\
    \begin{matrix}
      0 & 1 & -e^{-c x} \\
      1 & 1 & 0
    \end{matrix}\bigg] = x\ \! \sig(c x),
\end{equation}
where $c$ is a trainable parameter. For $c=1$, the resulting activation function
in Eq.~\ref{EQ:FSWISH} is referred to as Swish-1.\cite{Ramachandran:2017:ARXIV}
Plots of Swish-1 activation function and its gated variant are shown in Fig.~\ref{FIG:SWISH1}.
Setting $f(x) = -|x|$ in the gate function, one can convert Swish into Softsign, defined as
\begin{equation}\label{EQ:FSOFTSIGN}
  x \Phi\bigg[x \bigg| 1\
    \begin{matrix}
      0 & 1 & -|x| \\
      1 & 1 & 0
    \end{matrix}\bigg] = \frac{x}{1+|x|}.
\end{equation}
Plots of Softsign activation function and its gated representation are illustrated in
Fig.~\ref{FIG:SOFTSIGN}. The gate functional for the hyperbolic tangent activation function
takes $f(x) = g(x) = x^2$ to yield
\begin{equation}\label{EQ:FTANH}
  x \Phi\bigg[x \bigg| 1\
    \begin{matrix}
      2 & 2 & x^2 \\
      2 & 1 & x^2
    \end{matrix}\bigg] = \tanh(x).
\end{equation}
Plots of hyperbolic tangent activation function and its gated representation are
presented in Fig.~\ref{FIG:TANH}. As mentioned in Sec.~\ref{SEC:INTRODUCTION}, the
gated functional form in Eq.~\ref{EQ:FUNC} arms us with significant variational
flexibility-- In addition to accessing a set of fixed-shape activation functions via
setting the gate function parameters, we can also interpolate between different functional
forms by varying those parameters over a finite domain. Figure \ref{FIG:FLEX} illustrates
an example where by fixing all parameters in the gated representation of hyperbolic
tangent except $\be_2$, one can smoothly interpolate between linear $(\be_2 = 2)$ and
hyperbolic tangent $(\be_2 = 1)$ activation functions. Thus, it is possible to tune
the saturation behavior of gated representation of saturating functions such as hyperbolic
tangent and mitigate their vanishing/exploding gradient problem in a controlled
fashion.\cite{Nair:2010:807,Glorot:2010:249,Glorot:2011:315} Furthermore, one can turn
$\be_2$ (or in principle, any other parameter) into a trainable parameter and allow the
hosting neural network to learn its optimal value from the training data.

Our unification strategy can go beyond the aforementioned list of fixed-shape or trainable
activation functions. For instance, Mish\cite{Misra:2019:2019} can be obtained by passing
Softplus, $\log(1+e^x)$, to hyperbolic tangent gate function as an argument and
setting $\ga=2$ to get
\begin{equation}\label{EQ:FMISH}
  x \Phi\bigg[ \log(1+e^x) \bigg| 2\
    \begin{matrix}
      2 & 2 & x^2 \\
      2 & 1 & x^2
    \end{matrix}\bigg] = x\ \! \tanh\big[\log(1+e^x) \big].
\end{equation}
The Bipolar Sigmoid function can also be expressed by using the hyperbolic tangent gate
function and passing a scaled linear function as an argument
\begin{equation}\label{EQ:BIPOLARSIGMOID}
  x \Phi\bigg[\frac{x}{2} \bigg| 1\
    \begin{matrix}
      2 & 2 & x^2 \\
      2 & 1 & x^2
    \end{matrix}\bigg] = \tanh\left(\frac{x}{2}\right).
\end{equation}
Equivalently, one can also express the Bipolar Sigmoid function with a different set of
paramenters and arguments in the gate function as
\begin{equation}\label{EQ:BIPOLARSIGMOID2}
  x \Phi\bigg[x \bigg| 0\
    \begin{matrix}
      0 & 1 & -e^{-x} \\
      0 & 1 & e^{-x}
    \end{matrix}\bigg] = \frac{1-e^{-x}}{1+e^{-x}}.
\end{equation}
Setting $f(x) = \tfrac{x}{\sqrt{2}}$ and $g(x) = \tfrac{x^2}{2}$, the \ac{GELU}
activation function can also be written as
\begin{equation}\label{EQ:GELU}
  \frac{x}{2} \Phi\bigg[x \bigg| 1\
    \begin{matrix}
      \frac{1}{2} & 1 & \frac{x}{\sqrt{2}} \\
      1           & 1 & \frac{x^2}{2}
    \end{matrix}\bigg] = \frac{x}{2} \bigg[ 1+\erf\left(\frac{x}{\sqrt{2}}\right)\bigg],
\end{equation}
where $\erf(\cdot)$ is the error function.\cite{Abramowitz:1964:BOOK,Olver:2010:NISTBOOK}

\section{Computational details}\label{SEC:COMPDETAILS}
In order to compare the efficiency and performance of gated activation functions and
their counterparts from Table \ref{TAB:PARAMS}, we design four sets of image classification
experiments involving multiple neural network architectures and datasets of different
sizes and complexities. In particular, we train the classical LeNet-5 neural 
network\cite{Lecun:1998:2278} on the \ac{MNIST}\cite{LeCun:1988:MNIST} and 
CIFAR-10\cite{Krizhevsky:2009:CIFAR10} datasets as well as ShuffleNet-v2 and ResNet-101
neural networks on the ImageNet-1k dataset.\cite{ImageNet:2015:211} 

The first two sets of experiments involve replacing all three element-wise \ac{RELU} 
activation functions in LeNet-5 architecture (Fig.~\ref{FIG:LENET5}) with their 
counterparts from Table \ref{TAB:PARAMS}. In each experiment, we run an ensemble of twenty 
independent sessions and train the LeNet-5 neural network on the \ac{MNIST} and CIFAR-10 
datasets for 10 and 20 epochs, respectively. In order to help the fairness of the comparisons
between different experiments, we randomly initialize the network parameters using ``1234'' as 
a seed to ensure individual training sessions in each ensemble start with the same set 
of parameters. All training sessions are performed in-memory,\cite{Wolfram:2022:largedatasets}
with a batch size of 64 and in single-precision. The average results are rounded to 
four significant digits and reported in Tables \ref{TAB:MNIST} and \ref{TAB:CIFAR10}.

In the next set of experiments, we replace the \ac{RELU} activation function in the
ShuffleNet-v2's terminal convolutional layer (Fig.~\ref{FIG:SHUFFLENETV2}) with its 
counterparts from Table \ref{TAB:PARAMS}. We also modify the ResNet-101 architecture by
replacing a pair of \ac{RELU} activation functions in the last bottleneck block 
(Fig.~\ref{FIG:RESNET101}) with various functions from Table \ref{TAB:PARAMS}.
Each experiment is an ensemble of three independent sessions where we train the 
ShuffleNet-v2 and ResNet-101 neural networks on the ImageNet-1k dataset with batch sizes
of 1024 and 128, respectively. He initialization\cite{He:2015:1026} method and mixed precision
are used for training both neural networks and the best average performance results 
presented in Tables \ref{TAB:SHUFFLENET} and \ref{TAB:RESNET101}, respectively. 
All training sessions pertinent to the ShuffleNet-v2 and ResNet-101 neural 
networks are performed out-of-core, in which batches of data are transferred to the 
neural network on the \ac{GPU} on-the-fly.\cite{Wolfram:2022:largedatasets} We stop
the training when the absolute change in the macro-average value of F1-score falls below 
0.001 for at least ten consecutive epochs.

The \ac{ADAM} optimizer is used for all training experiments with the stability
parameter, the first and the second moment exponential decay rates set to
$\epsilon=10^{-5}$, $\be_1=0.9$ and $\be_2=0.999$, respectively. The initial
learning rate is set to 0.001 and automatically modified by the program.
All computations are performed using Wolfram 
\mathematica\ 13.2.\cite{Mathematica:2022:SOFTWARE} Bipolar Sigmoid and
\ac{GELU} are excluded from our studies as their hosting neural networks fail to
converge without deviating from the selected default settings in \mathematica\ and
further modifications to avoid the divergence. In this manuscript, we do not make any
attempts to optimize the performance of the neural networks by fine-tuning various 
hyperparameters. 

Three hardware platforms are adopted for performing the computations:
a single laptop armed with a \nvidia\ GeForce GTX 1650 \ac{GPU}, a Supermicro workstation
with 2 $\times$ \nvidia\ A100 80GB PCIe \acp{GPU} and a \nvidia\ DGX \ac{HPC} cluster 
node with 8 $\times$ \nvidia\ A100 80GB PCIe \acp{GPU}. The resulting data from the 
first setup can be found in the \si. The trajectory of all training instances are 
recorded in byte representation files that can be instantly reproduced in \mathematica.
Furthermore, all input scripts and output logs alongside a simple \mathematica\ code
snippet are also provided to assist the readers in reproducing the results.

\section{Results and discussion}\label{SEC:RESULTS}
In order to quantify the impacts of various activation functions on the performance
of LeNet-5, ShuffleNet-v2 and ResNet-101 classifiers during training and validation,
we focus on a variety of numerical metrics such as loss, accuracy, precision, recall
and F1-score. The training loss is measured via multi-class cross-entropy
which is defined as\cite{Wilmott:2019:BOOK,Geron:2017:BOOK}
\begin{equation}\label{EQ:CROSSENTROPY}
  \mathscr{L} = -\sum_{i=1}^{N} \sum_{k=1}^{K} y_{i,k} \ln(\hat{y}_{i,k}).
\end{equation}
For each data point (image) $i$ in a dataset of size $N$, cross-entropy can measure how well
the estimated probabilities, $\hat{y}_{i,k}$ for each class $k$, where
$k \in \lbrace 1, 2, 3, \dots, K \rbrace$, match those of the target class labels,
$y_{i,k}$. Compared with other loss functions, cross-entropy can also improve the
convergence rate of the optimization process in our study by aggressively penalizing
the incorrect predictions and generating larger gradients.\cite{Geron:2017:BOOK}
Accuracy is defined as the fraction of the number of times that a classifier is correct
in its predictions.\cite{Wilmott:2019:BOOK} Since accuracy is not an appropriate metric for
imbalanced datasets,\cite{Geron:2017:BOOK} such as ImageNet-1k,\cite{Luccioni:2022:ARXIV}
we also consider precision and recall as metrics for the classification task. 
Precision and recall are the ratios of the number of correctly predicted positive 
classes to the total number of instances that are predicted as or are indeed positive, 
respectively. In theory, we are interested in classifiers that have high precision and
recall values. In practice, however, one adopts the harmonic mean of precision and recall
(called F1-score) in order to take into account the tradeoff between the aforementioned
two metrics. We report the macro-average values of precision, recall and F1-score
for all studied multi-class classification tasks, notwithstanding our knowledge about the 
(im)balanced nature of the class distributions in \ac{MNIST}, CIFAR-10 or ImageNet-1k 
datasets. We also consider training runtime and processing rate, as recommended by 
Ref.~\citenum{Ma:2018:122}, to better reflect the effects of various activation functions 
on the computational cost and complexity of each neural network.

\subsection{Training LeNet-5 on MNIST dataset}\label{SUBSEC:LENETMNIST}
\begin{figure}[!tbph]
  \centering
  \includegraphics{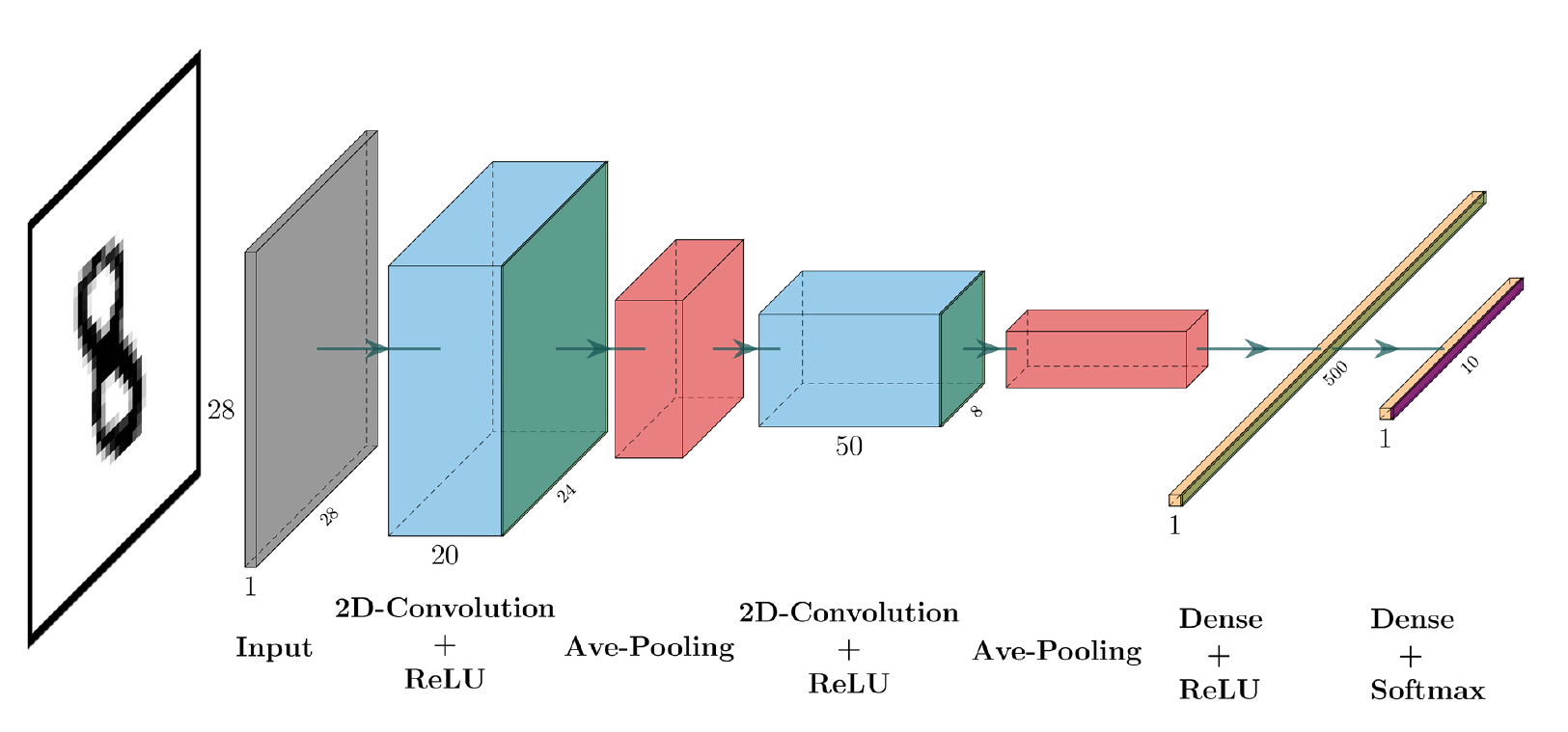}
  \caption{LeNet-5 neural network architecture}
  \label{FIG:LENET5}
\end{figure}

The \ac{MNIST} dataset consists of 60,000 training and 10,000 testing grayscale images
of hand-written digits ($0, 1, 2, \dots, 9$) that are normalized and centered to a
28$\times$28 fixed size. We pulled the \ac{MNIST} dataset from the Wolfram Data
Repository\cite{LeCun:2016:WOLFRAM}. Individual training sessions (excluding that of the
baseline) involve replacing all three element-wise \ac{RELU} activation layers in the
LeNet-5 architecture (Fig.~\ref{FIG:LENET5}) with their counterparts from Table \ref{TAB:PARAMS}.

Table \ref{TAB:MNIST} shows the performance results for training/validation of LeNet-5 
neural network on the \ac{MNIST} dataset. For each activation function, there are two 
entries: the first entry refers to the results of built-in activation functions in 
\mathematica\ 13.2 and the second one corresponds to those of gated activation functions.

\begin{table*}[!htbp]
  \centering
  \setlength{\tabcolsep}{3pt}
  \setlength{\extrarowheight}{1pt}
  \caption{The best performance metrics and timings pertinent to training and testing
    of LeNet-5 neural network on \ac{MNIST} dataset with various activation functions$^{a}$}
  \label{TAB:MNIST}
  \begin{tabular}{lccc}
    \hline\hline
    Activation Function$^b$ & Loss            & Accuracy (\%) & Wall-Clock Time (s) \\
    \hline
    \mr{2}{*}{Sigmoid}      & 0.0270 (0.0290) & 99.13 (99.13) & 21.93               \\
                            & 0.0270 (0.0290) & 99.13 (99.13) & 30.39               \\[3pt]
    \mr{2}{*}{Swish-1}      & 0.0411 (0.0295) & 98.73 (98.98) & 28.38               \\
                            & 0.0422 (0.0295) & 98.69 (98.99) & 30.60               \\[3pt]
    \mr{2}{*}{Softsign}     & 0.0138 (0.0324) & 99.56 (99.00) & 28.39               \\
                            & 0.0137 (0.0315) & 99.56 (99.03) & 28.61               \\[3pt]
    \mr{2}{*}{$\tanh$}      & 0.0146 (0.0329) & 99.52 (98.95) & 20.77               \\
                            & 0.0126 (0.0325) & 99.58 (99.01) & 29.72               \\[3pt]
    \mr{2}{*}{Mish}         & 0.0279 (0.0299) & 99.14 (99.07) & 24.38               \\
                            & 0.0339 (0.0304) & 98.96 (99.03) & 40.73               \\[3pt]
    \hline
    ReLU$^c$                & 0.0144 (0.0292) & 99.54 (99.16) & 23.78               \\
    \hline\hline
  \end{tabular}
  \begin{tablenotes}
    \scriptsize
    \item \qquad \qquad \qquad \quad $^a$ All results are ensemble averages over 20
    independent training and testing experiments \\ \qquad \qquad \qquad \qquad
    performed on a Supermicro workstation with \nvidia\ A100 80GB PCIe \acp{GPU}.
    \item \qquad \qquad \qquad \quad $^b$ The test results are given in parentheses.
    The first and second rows in each activation \\ \qquad \qquad \qquad \qquad
    function entry correspond to the built-in and gated representations, respectively.
    \item \qquad \qquad \qquad \quad $^c$ The LeNet-5 neural network architecture with
    ReLU activation functions is taken as \\ \qquad \qquad \qquad \qquad baseline architecture.
  \end{tablenotes}
\end{table*}

Table \ref{TAB:MNIST} reveals that the average validation accuracy of LeNet-5 classifier
on the \ac{MNIST} test set is not significantly sensitive towards the choice of
activation functions in the element-wise layers. In particular, the validation accuracy
of LeNet-5 classifier armed with Sigmoid activation function is slightly smaller from that
of the baseline neural network with \acp{RELU}. Furthermore, choosing other activation
functions such as Softsign and Mish seem to further deteriorate the corresponding average
validation accuracies compared with that of \acp{RELU} in the baseline LeNet-5 architecture.
Plots of training/validation loss and accuracy versus epochs can be found in \si.

Our main interest in Table \ref{TAB:MNIST} is in the average values of total wall-clock
time spent on training LeNet-5 on \ac{MNIST} dataset using a Supermicro Workstation with
\nvidia\ A100 80GB PCIe \acp{GPU}. The average timings reveal that the added cost of
calculating one- or two-parameter Mittag-Leffler functions in the gated representation
of activation functions is small compared with their built-in variants implemented
in \mathematica\ 13.2 program package.\cite{Wolfram:2022:elementwiselayer} Specifically,
the largest measured time gap is observed between the built-in and gated representations
of Mish which mainly stems from the overhead of calculating Softplus and passing it as
an argument to the two-parameter Mittag-Leffer function for each neural response in
the element-wise activation layers. On the other hand, the computational time gap
between built-in and gated representation of Softsign is very small on average.

\subsection{Training LeNet-5 on CIFAR-10 dataset}\label{SUBSEC:LENETCIFAR10}
Table \ref{TAB:CIFAR10} presents the classification performance results 
for training/validation of LeNet-5 neural network on CIFAR-10 dataset which contains 
50,000 training and 10,000 test images from 10 object classes (airplane, automobile,
bird, cat, deer, dog, frog, horse, ship, and truck). Each data point in the CIFAR-10 
dataset is a 32$\times$32 RGB image.\cite{Krizhevsky:2009:WOLFRAM}

\begin{table*}[!htbp]
  \centering
  \setlength{\tabcolsep}{3pt}
  \setlength{\extrarowheight}{1pt}
  \caption{The best performance metrics and timings pertinent to training and testing
    of the LeNet-5 neural network on CIFAR-10 dataset with various activation functions$^{a}$}
  \label{TAB:CIFAR10}
  \begin{tabular}{lccc}
    \hline\hline
    Activation Function$^b$ & Loss            & Accuracy (\%) & Wall-Clock Time (s) \\
    \hline
    \mr{2}{*}{Sigmoid}      & 0.7353 (1.0151) & 74.79 (65.39) & 44.26               \\
                            & 0.7353 (1.0152) & 74.78 (65.38) & 54.97               \\[3pt]
    \mr{2}{*}{Swish-1}      & 0.7323 (0.9016) & 74.55 (69.39) & 54.44               \\
                            & 0.7322 (0.9018) & 74.55 (69.40) & 60.77               \\[3pt]
    \mr{2}{*}{Softsign}     & 0.6380 (0.9297) & 77.89 (68.64) & 53.09               \\
                            & 0.6280 (0.9313) & 78.26 (68.56) & 54.52               \\[3pt]
    \mr{2}{*}{$\tanh$}      & 0.8129 (0.9766) & 71.63 (66.79) & 45.58               \\
                            & 0.8015 (0.9738) & 72.05 (66.90) & 56.09               \\[3pt]
    \mr{2}{*}{Mish}         & 0.6842 (0.8914) & 76.18 (69.40) & 48.37               \\
                            & 0.6841 (0.8913) & 76.18 (69.44) & 78.09               \\[3pt]
    \hline
    ReLU$^c$                & 0.7299 (0.9279) & 74.57 (68.24) & 45.08               \\
    \hline\hline
  \end{tabular}
  \begin{tablenotes}
    \scriptsize
    \item \qquad \qquad \qquad \quad $^a$ All results are ensemble averages over 20
    independent training and testing experiments \\ \qquad \qquad \qquad \qquad
    performed on a Supermicro workstation with \nvidia\ A100 80GB PCIe \acp{GPU}.
    \item \qquad \qquad \qquad \quad $^b$ The test results are given in parentheses.
    The first and second rows in each activation \\ \qquad \qquad \qquad \qquad
    function entry correspond to the built-in and gated representations, respectively.
    \item \qquad \qquad \qquad \quad $^c$ The LeNet-5 neural network architecture with
    ReLU activation functions is taken as \\ \qquad \qquad \qquad \qquad baseline architecture.
  \end{tablenotes}
\end{table*}

All results correspond to the average of 20 individual training sessions, each running for
20 epochs. A comparison of the average accuracy values in Table \ref{TAB:CIFAR10} with
those in Table \ref{TAB:MNIST} reflects the more intricate nature of the CIFAR-10 dataset
which requires deeper neural networks, more advanced architectural design and training
strategies. The interested reader is referred to a study on ``Convolutional Deep Belief
Neural Networks on CIFAR-10''\cite{Krizhevsky:2010:UNPUBLISHED} and ImageNet competition
for a chronological survey of the efforts on this topic.\cite{ImageNet:2015:211}
Table \ref{TAB:CIFAR10} reveals that the validation accuracy of LeNet-5 neural network
can be improved by replacing \acp{RELU} with any other activation function considered
in this study with the exception of Sigmoid and hyperbolic tangent. In particular,
replacing \acp{RELU} with Swish-1 or Mish yields the largest improvement in the validation
accuracy of approximately 1.2 \%.

The average timings for training LeNet-5 neural network on CIFAR-10 dataset show similar
trends to those presented in Table \ref{TAB:MNIST}. Among all studied variants of the LeNet-5
architecture, those with built-in and gated Mish activation functions show the largest wall-clock 
time difference of approximately 30 seconds. Note that the timings in Table \ref{TAB:CIFAR10} 
are roughly twice their counterparts in Table \ref{TAB:MNIST} due to the adopted number of 
training epochs (20 for CIFAR-10 compared with 10 for \ac{MNIST}). The average timings 
reported in both tables suggest that a unified implementation of the most popular 
activation functions using Mittag-Leffler functions is possible at an affordable computational cost compared with 
their individual built-in implementations. The gap between the built-in 
and gated representations of activation functions can be further reduced as more
efficient algorithms and implementations of special functions such as Mittag-Leffler
function become available. A comparison of the aforementioned timings obtained using a 
Supermicro workstation with \nvidia\ A100 80GB PCIe \acp{GPU} (Tables \ref{TAB:MNIST} 
and \ref{TAB:CIFAR10}) with those computed by a laptop with a \nvidia\ GeForce GTX 
1650 \ac{GPU} (see \si) demonstrates that the availability of more powerful 
computing accelerators can also be an important factor for training \acp{ANN} with large
numbers of gated activation functions.

\subsection{Training ShuffleNet-v2 on ImageNet-1k dataset}\label{SUBSEC:SHUFFLEIMAGENET}
\begin{figure}[!tbph]
  \centering
  \includegraphics{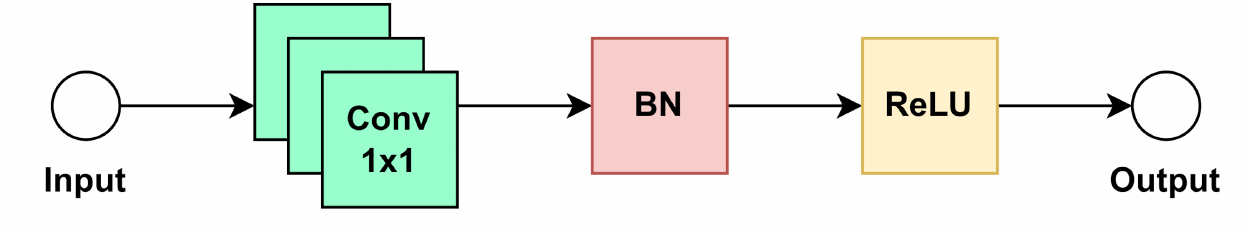}
  \caption{The final (target) convolution layer in the ShuffleNet-v2 architecture}
  \label{FIG:SHUFFLENETV2}
\end{figure}

The ImageNet-1k dataset consists of 1,281,167 training, 50,000 validation and
100,000 test RGB images within 1000 categories.\cite{ImageNet:2015:211} All images are
cropped and resized to 224$\times$224 pixels during preprocessing.
Table \ref{TAB:SHUFFLENET} shows the performance results for training/validation of
ShuffleNet-v2\cite{Ma:2018:122} on the ImageNet-1k dataset where individual entries
correspond to the ensemble average of three independent training experiments.
\begin{table*}[!htbp]
  \centering
  \setlength{\tabcolsep}{3pt}
  \setlength{\extrarowheight}{1pt}
  \caption{The best performance metrics and timings for training and testing ShuffleNet-v2
    neural network on the ImageNet-1k dataset using various activation functions$^{a}$}
  \label{TAB:SHUFFLENET}
  \resizebox{\textwidth}{!}{
    \begin{tabular}{lccccccc}
      \hline\hline
      \mr{2}{*}{Activation Function$^b$} & \mr{2}{*}{Loss} & Accuracy      & Precision     & Recall        & F1 Score      & Processing Rate & Wall-Clock Time \\
                                         &                 & (\%)          & (\%)          & (\%)          & (\%)          & (images/s)      & (h)             \\
      \hline
      \mr{2}{*}{Sigmoid}                 & 0.8648 (1.923)  & 76.49 (59.55) & 76.32 (62.24) & 76.42 (59.55) & 76.35 (59.41) & 466             & 20.42           \\
                                         & 0.8885 (1.919)  & 75.95 (59.33) & 75.77 (62.02) & 75.87 (59.33) & 75.80 (59.16) & 363             & 23.12           \\[3pt]
      \mr{2}{*}{Swish-1}                 & 0.7114 (1.860)  & 80.36 (61.91) & 80.25 (63.92) & 80.32 (61.91) & 80.27 (61.84) & 437             & 17.63           \\
                                         & 0.7113 (1.874)  & 80.37 (61.98) & 80.26 (63.89) & 80.32 (61.98) & 80.28 (61.82) & 348             & 20.44           \\[3pt]
      \mr{2}{*}{SoftSign}                & 0.7836 (1.845)  & 78.52 (60.96) & 78.37 (63.13) & 78.46 (60.96) & 78.40 (60.84) & 413             & 22.70           \\
                                         & 0.8114 (1.823)  & 77.87 (60.92) & 77.72 (62.99) & 77.81 (60.92) & 77.74 (60.72) & 313             & 26.88           \\[3pt]
      \mr{2}{*}{$\tanh$}                 & 0.8288 (1.882)  & 77.46 (60.53) & 77.32 (62.73) & 77.41 (60.53) & 77.34 (60.35) & 346             & 27.38           \\
                                         & 0.8488 (1.867)  & 77.02 (60.50) & 76.87 (62.78) & 76.97 (60.50) & 76.90 (60.33) & 336             & 24.33           \\[3pt]
      \mr{2}{*}{Mish}                    & 0.7545 (1.825)  & 79.37 (61.99) & 79.25 (64.10) & 79.32 (61.99) & 79.27 (61.87) & 307             & 26.48           \\
                                         & 0.7057 (1.859)  & 80.47 (62.05) & 80.35 (63.90) & 80.41 (62.05) & 80.37 (61.90) & 360             & 24.22           \\[3pt]
      \hline
      ReLU$^c$                           & 0.7456 (1.669)  & 79.78 (63.50) & 79.67 (64.80) & 79.72 (63.50) & 79.68 (63.33) & 302             & 20.06           \\
      \hline\hline
    \end{tabular}
  }
  \begin{tablenotes}
    \scriptsize
    \item $^a$ All results are ensemble averages over 3 independent training and testing experiments
    performed on a \nvidia\ DGX \\ \quad \ac{HPC} cluster node with \nvidia\ A100 80GB PCIe \acp{GPU}.
    \item $^b$ The test results are given in parentheses. The first and second rows in each activation
    function entry correspond to \\ \quad the built-in and gated representations, respectively.
    \item $^c$ The ShuffleNet-v2 neural network (Ref.~\citenum{Ma:2018:122}) with \acs{RELU} activation functions is
    taken as baseline architecture.
  \end{tablenotes}
\end{table*}
Each experiment involves replacing the \ac{RELU} activation function in the ShuffleNet-v2's
final convolutational layer (Fig.~\ref{FIG:SHUFFLENETV2}) with one of its counterparts
from Table \ref{TAB:PARAMS}.

Table \ref{TAB:SHUFFLENET} provides numerical evidence of overfitting where ShuffleNet-v2
performs significantly better in memorizing the training data than generalizing to 
the unseen data during validation (F1-score of 75-80\% vs. 59-63\%). The best validation 
loss and accuracy values are obtained by the ShuffleNet-v2 baseline architecture with \ac{RELU} 
activation functions. However, substituting the \ac{RELU} activation function with other
variants from Table \ref{TAB:PARAMS} deteriorates the validation loss and accuracy by at 
least 9\% and 2\%, respectively. Similar trends are observed for validation precision,
recall and F-score macro-averages across Table \ref{TAB:SHUFFLENET} which also highlight
the imbalanced nature of the ImageNet-1k dataset\cite{Luccioni:2022:ARXIV}-- one of the main 
reasons behind choosing F1-score as our early stopping convergence criterion for training
and performance metric for analysis. Table \ref{TAB:SHUFFLENET} demonstrates that the top
three performers based on validation F1-score are the ShuffleNet-v2 neural networks 
with \ac{RELU} (63.3\%), Mish (61.9\%), and Swish-1 (61.8\%) activation functions, respectively.

The performance results presented in Table \ref{TAB:SHUFFLENET} can also be impacted by 
system-dependent factors such as out-of-core data transfer bandwidth, core affinity,
task binding and distribution over sockets, \etc, commonly encountered in shared \ac{HPC} 
cluster environments.\cite{Ma:2018:122} As such, we complement our data in 
Table \ref{TAB:SHUFFLENET}  with two additional temporal metrics to quantify the 
efficiency of ShuffleNet-v2 neural network with various activation functions: the total 
wall-clock runtime and processing rate. The availability of 80GB of \ac{GPU} memory 
allowed us to adopt larger batch sizes (1024 vs. 4 images/batch) to facilitate achieving higher
processing rates (approximately, 470 images/s) during training compared with those 
reported in Ref.~\citenum{Ma:2018:122} (190 images/s). Nonetheless, Table \ref{TAB:SHUFFLENET}
illustrates that the processing rates of gated activation functions are in general lower
than their built-in counterparts due to higher memory access costs and computational complexity.
An exception is gated Mish activation function which is 53 images/s faster than its built-in
variant. Notably, the processing rate of training ShuffleNet-v2 with built-in hyperbolic tangent
is only 10 images/s faster than that of its gated counterpart. Both observations are 
corroborated by shorter total training wall-clock runtimes for gated Mish and hyperbolic 
tangent functions compared with those of their built-in variants. Furthermore, training 
ShuffleNet-v2 with Sigmoid and SoftSign activation functions show the largest 
differences in processing rates (of about 100 images/s) between their built-in and gated 
variants. The highest processing rates, however, are obtained by the built-in variants of 
Sigmoid (466 images/s), Swish-1 (437 images/s) and SoftSign (413 images/s) activation 
functions, respectively. The minimum average runtime of 17 hours is measured for training 
ShuffleNet-v2 neural network with built-in Swish-1 activation function which is approximately
3 hours shorter than those required by \ac{RELU}, built-in Sigmoid or gated Swish-1 activation
functions.

\subsection{Training ResNet-101 on ImageNet-1k dataset}\label{SUBSEC:RESNETIMAGENET}
The performance results of training/validation of ResNet-101\cite{He:2016:770} on the
ImageNet-1k dataset are shown in Table \ref{TAB:RESNET101} where each entry 
corresponds to the average of three independent training sessions.

\begin{figure}[!tbph]
  \centering
  \includegraphics{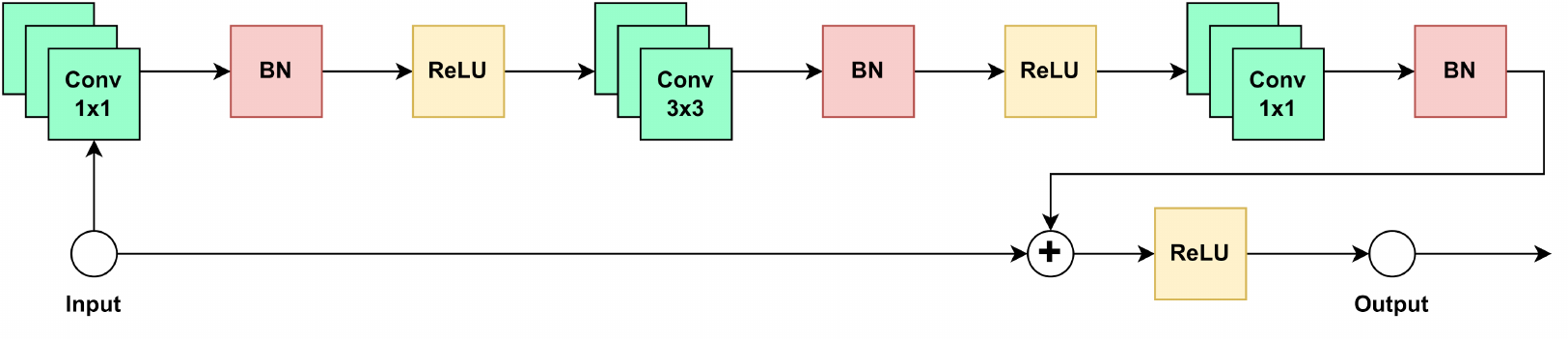}
  \caption{The bottleneck design with an identity shortcut in the ResNet-101 architecture}
  \label{FIG:RESNET101}
\end{figure}

\begin{table*}[!htbp]
  \centering
  \setlength{\tabcolsep}{3pt}
  \setlength{\extrarowheight}{1pt}
  \caption{The best performance metrics and timings for training and testing ResNet-101
    neural network on the ImageNet-1k dataset using various activation functions$^{a}$}
  \label{TAB:RESNET101}
  \resizebox{\textwidth}{!}{
    \begin{tabular}{lccccccc}
      \hline\hline
      \mr{2}{*}{Activation Function$^b$} & \mr{2}{*}{Loss} & Accuracy      & Precision     & Recall        & F1 Score      & Processing Rate & Wall-Clock Time \\
                                         &                 & (\%)          & (\%)          & (\%)          & (\%)          & (images/s)      & (d)             \\
      \hline
      \mr{2}{*}{Sigmoid}                 & 0.0700 (2.120)  & 98.08 (67.57) & 98.08 (68.71) & 98.07 (67.57) & 98.07 (67.40) & 143             & 3.1             \\
                                         & 0.1268 (1.937)  & 96.34 (67.45) & 96.34 (68.65) & 96.33 (67.29) & 96.33 (67.29) & 82              & 4.6             \\[3pt]
      \mr{2}{*}{Swish-1}                 & 0.0637 (2.118)  & 98.28 (67.76) & 98.28 (68.90) & 98.28 (67.76) & 98.28 (67.60) & 89              & 4.8             \\
                                         & 0.0412 (2.382)  & 98.81 (67.82) & 98.81 (68.79) & 98.81 (67.82) & 98.81 (67.66) & 92$^d$          & 8.3$^d$         \\[3pt]
      \mr{2}{*}{SoftSign}                & 0.1691 (1.832)  & 95.16 (67.47) & 95.14 (68.81) & 95.14 (67.47) & 95.14 (67.33) & 140             & 2.5             \\
                                         & 0.1710 (1.791)  & 95.07 (67.39) & 95.06 (68.71) & 95.06 (67.39) & 95.06 (67.27) & 82              & 4.2             \\[3pt]
      \mr{2}{*}{$\tanh$}                 & 0.0632 (2.181)  & 98.26 (67.82) & 98.25 (68.66) & 98.25 (67.82) & 98.25 (67.58) & 95              & 5.8             \\
                                         & 0.2184 (1.714)  & 93.75 (67.26) & 93.73 (68.62) & 93.73 (67.26) & 93.73 (67.10) & 77              & 3.0             \\[3pt]
      \mr{2}{*}{Mish}                    & 0.0382 (2.450)  & 98.88 (68.02) & 98.88 (68.89) & 98.88 (68.02) & 98.88 (67.76) & 103$^d$         & 8.2$^d$         \\
                                         & 0.0741 (2.103)  & 97.96 (67.79) & 97.95 (68.80) & 97.95 (67.79) & 97.95 (67.61) & 81              & 5.6             \\[3pt]
      \hline
      ReLU$^c$                           & 0.0689 (2.121)  & 98.14 (67.84) & 98.14 (68.92) & 98.13 (67.84) & 98.13 (67.69) & 86              & 5.7             \\
      \hline\hline
    \end{tabular}
  }
  \begin{tablenotes}
    \scriptsize
    \item $^a$ All results are ensemble averages over 3 independent training and testing experiments
    performed on a \nvidia\ DGX \\ \quad \ac{HPC} cluster node with \nvidia\ A100 80GB PCIe \acp{GPU}.
    \item $^b$ The test results are given in parentheses. The first and second rows in each activation
    function entry correspond to \\ \quad the built-in and gated representations, respectively.
    \item $^c$ The ResNet-101 architecture (Ref.~\citenum{He:2016:770}) with the terminal bottleneck block of Fig.~\ref{FIG:RESNET101} is taken as baseline architecture.
    \item $^d$ Restarted trainings affected by the runtime limitations on the cluster node.
  \end{tablenotes}
\end{table*}
Individual training sessions involve replacing the first two \ac{RELU} activation functions 
in the final convolutional bottleneck block of the ResNet-101 baseline architecture 
(Fig.~\ref{FIG:RESNET101}) with one of their counterparts from Table \ref{TAB:PARAMS}.

The performance results in Table \ref{TAB:RESNET101} suggest that, similar to ShuffleNet-v2,
the ResNet-101 neural network also overfits the ImageNet-1k data. Choosing different activation
functions in ResNet101's baseline architecture does not significantly change the value of
validation performance metrics (\ie, accuracy, precision, recall and F1-score) with the 
exception of loss. In particular, variants of ResNet-101 neural network with gated hyperbolic 
tangent and vanilla Mish show the best validation loss (1.714) and accuracy (68.02\%), respectively.
Substituting the \ac{RELU} activation functions in the ResNet-101's baseline architecture 
with any of their counterparts from Table \ref{TAB:PARAMS} slightly deteriorates the validation
precision and recall macro-averages with the exception of built-in Mish which 
improves the validation recall by less than 1\%. The aforementioned improvement in the 
validation recall is also reflected in the value of F1-score corresponding to vanilla Mish
(67.76\%) compared with that of ResNet-101's baseline architecture (67.69\%).

Due to the runtime limitations on our \ac{HPC} node, we had to restart our computations
during training ResNet-101 with gated Swish-1 and built-in Mish activation functions. As such,
we expect that the reported values for processing rates and wall-clock times for the
two aforementioned cases to be affected by the disruption. Comparing the processing times in 
Table \ref{TAB:RESNET101} with those in Table \ref{TAB:SHUFFLENET} reveals that the 
processing times corresponding to ShuffleNet-v2 are much higher than those of ResNet-101 
which is consistent with the previous reports in literature.\cite{Ma:2018:122} However,
care must be taken as the adopted batch sizes for the two training sets are quite different 
(1024 for ShuffleNet-v2 vs. 128 for ResNet-101). The highest processing rates (of approximately
140 images/s) correspond to the ResNet-101 neural networks armed with built-in Sigmoid and
SoftSign activation functions. The shortest training runtime is around 2 days which belongs
to the ResNet-101 neural network with built-in SoftSign activation function. In comparison, the
total training runtime corresponding to the baseline ResNet-101 architecture is longer by 
approximately 3 days.

\section{Conclusion and future work}\label{SEC:CONCLUSION}
In this manuscript, we have presented a unified representation of some of the most
popular neural network activation functions of fixed-shape type. The proposed functional
form not only sheds light on the direct analytical connections between several
well-established activation functions in the literature but also allows for interpolating
between different functional forms through varying the gate function parameters. The
derivative of the gated activation function, defined in terms of Mittag-Leffler
functions, is closed under differentiation. This characteristic of gated representations
makes them a suitable candidate for training neural networks through gradient-based
methods of optimization. A unified representation of activation functions is also
beneficial to studies which use fixed-shape or trainable activation functions as
it can lead to large savings in terms of number of code lines compared to what is otherwise
required for individual implementation of activation functions in popular \acl{ML} frameworks
via inheritance and/or customized classes. Through training the classic LeNet-5, ShuffleNet-v2
and ResNet-101 neural networks on standard benchmark datasets such as \ac{MNIST}, CIFAR-10,
and ImageNet-1k, we have established that a unified implementation of activation functions
is possible without any sacrifice in validation performance and at an affordable computational 
cost. The use of one- and two-parameter Mittag-Leffler functions and their relation to 
other generalized and special functions\cite{Olver:2010:NISTBOOK} such as hypergeometric 
and Wright functions\cite{Mainardi:2020:1359} opens a door to a new and active area of 
research in fractional \ac{ANN} and backpropagation algorithms which is also under current
investigation by us.

\acresetall	
\appendix
\section*{Appendix}\label{SEC:APPENDIX}
\subsection{General formula for derivatives of Mittag-Leffler function}\label{APPSUBSEC:DERIVATIVES}
The differentials of the one- and two-parameter Mittag-Leffler Functions
can be expressed in terms of Mittag-Leffler function itself. The aforementioned
closeness property is computationally beneficial for an efficient implementation of 
the gradient descent-based backpropagation algorithms for training \acp{ANN}.
For a more in-depth discussion on differential and recurrence relations of
Mittag-Leffler functions of one-, two- and three-parameter(s), see 
Refs.~\citenum{Gorenflo:2020:BOOK,Garrappa:2018:129}.

Let $p \in \mathbb{N}$, where $\mathbb{N}$ denotes the set of natural numbers.
Then, the general derivatives of one-parameter Mittag-Leffler function can be given as
\begin{equation}\label{EQ:DMLF1A}
  \begin{gathered}
    \frac{d^{\ \! \! p}}{d z^p} E_p(z^p) = E_p(z^p),  \qquad  \text{and} \\
    \frac{d^{\ \! \! p}}{d z^p} E_{p/q}(z^{p/q}) = E_{p/q}(z^{p/q})
    + \sum_{k=1}^{q-1} \frac{z^{-k/q}}{\Gamma(1-k/q)} \qquad q = 2, 3, \dots .
  \end{gathered}
\end{equation}

Assuming $\al>0$ and $\be \in \be$, the first-derivative of the two-parameter 
Mittag-Leffler functions can be written as a sum of two instances of two-parameter
Mittag-Leffler functions as\cite{Garrappa:2018:129}
\begin{equation}\label{EQ:DMLF2}
  \frac{d}{d z}E_{\al,\be}(z) = \frac{E_{\al ,\al +\be -1}(z)+(1 - \be) E_{\al ,\al +\be}(z)}{\al}.
\end{equation}

In general, one can write
\begin{equation}
  \frac{d^{\ \! \! m}}{d z^m} E_{\al,\be}(z) = \frac{1}{\al^m} \sum_{k=0}^{m} 
  c^{(m)}_k E_{\al, \al m + \be-k}(z),  \qquad m \in \mathbb{N}
\end{equation}

where the $c_0^{(0)} = 1$ and the remaining coefficients for $k = 0, 1, 2, \dots$ can be computed
using the following recurrence relation

\begin{equation}
  c_k^{(m)} = 
  \begin{cases}
    \left[1-\be-\al(m-1)\right]c_0^{(m-1)},   \qquad k = 0,  \\
    c_{k-1}^{(m-1)} + \left[1-\be-\al(m-1)+k\right]c_k^{(m-1)}, \qquad  1 \leq k \leq m-1,  \\
    1,  \qquad k = m. 
  \end{cases}
\end{equation}

\acresetall	
\section*{Acknowledgements}\label{SEC:ACKNOWLEDGEMENTS}
The present work is funded by the National Science Foundation grant CHE-2136142.
The author would like to thank \nvidia\ Corporation for the generous Academic Hardware Grant 
and Virginia Tech for providing an institutional license to \mathematica\ 13.2.
The author also acknowledges the Advanced Research Computing (\url{https://arc.vt.edu}) at
Virginia Tech for providing computational resources and technical support that have 
contributed to the results reported within this manuscript.

\acresetall	

\begin{acronym}
    \acrodefplural{1-RDM}{one-electron \aclp{RDM}}
    \acrodefplural{2-RDM}{two-electron \aclp{RDM}}
    \acrodefplural{3-RDM}{three-electron \aclp{RDM}}
    \acrodefplural{4-RDM}{four-electron \aclp{RDM}}
    \acrodefplural{RDM}{reduced density matrices}
    \acro{1-HRDM}{one-hole \acl{RDM}}
    \acro{1-RDM}{one-electron \acl{RDM}}
    \acrodef{1H-PDFT}{one-parameter hybrid \acl{PDFT}}
    \acro{2-HRDM}{two-hole \acl{RDM}}
    \acro{2-RDM}{two-electron \acl{RDM}}
    \acro{3-RDM}{three-electron \acl{RDM}}
    \acro{4-RDM}{four-electron \acl{RDM}}
    \acro{ACI}{adaptive \acl{CI}}
    \acro{ACI-DSRG-MRPT2}{\acl{ACI}-\acl{DSRG} \acl{MR} \acl{PT2}}
    \acro{ACSE}{anti-Hermitian \acl{CSE}}
    \acro{ADAM}{adaptive momentum}
    \acro{AI}{artificial intelligence}
    \acrodef{AKEEPE}[$\AKEEe$]{absolute kinetic energy error per electron}
    \acrodef{AKEE}{absolute kinetic energy error}
    \acro{ANN}{artificial \acl{NN}}
    \acro{AO}{atomic orbital}
    \acro{AQCC}{averaged quadratic \acl{CC}}
    \acro{ATAC}{attentional activation}
    \acro{aug-cc-pVQZ}{augmented correlation-consistent polarized-valence quadruple-$\ze$}
    \acro{aug-cc-pVTZ}{augmented correlation-consistent polarized-valence triple-$\ze$}
    \acro{aug-cc-pwCV5Z}[aug-cc-pwCV5Z]{augmented correlation-consistent polarized weighted core-valence quintuple-$\ze$}
    \acro{B3LYP}{Becke-3-\acl{LYP}}
    \acro{BLA}{bond length alternation}
    \acro{BLYP}{Becke and \acl{LYP}}
    \acro{BN}{batch normalization}
    \acro{BO}{Born-Oppenheimer}
    \acro{BP86}{Becke 88 exchange and P86 Perdew-Wang correlation}
    \acro{BPSDP}{boundary-point \acl{SDP}}
    \acro{CAM}{Coulomb-attenuating method}
    \acro{CAM-B3LYP}{Coulomb-attenuating method \acl{B3LYP}}
    \acro{CAS-PDFT}{\acl{CAS} \acl{PDFT}}
    \acro{CASPT2}{\acl{CAS} \acl{PT2}}
    \acro{CASSCF}{\acl{CAS} \acl{SCF}}
    \acro{CAS}{complete active-space}
    \acro{cc-pVDZ}{correlation-consistent polarized-valence double-$\ze$}
    \acro{cc-pVTZ}{correlation-consistent polarized-valence triple-$\ze$}
    \acro{CCSDT}{coupled-cluster, singles doubles and triples}
    \acro{CCSD}{coupled-cluster with singles and doubles}
    \acro{CC}{coupled-cluster}
    \acro{CI}{configuration interaction}
    \acro{CNN}{convolutional \acl{NN}}
    \acro{CO}{constant-order}
    \acro{CPO}{correlated participating orbitals}
    \acro{CRELU}[CReLU]{concatrenated \acl{RELU}}
    \acro{CSF}{configuration state function}
    \acro{CS-KSDFT}{\acl{CS}-\acl{KSDFT}}
    \acro{CSE}{contracted \acl{SE}}
    \acro{CS}{constrained search}
    \acro{DC-DFT}{density corrected-\acl{DFT}}
    \acro{DC-KDFT}{density corrected-\acl{KDFT}}
    \acro{DE}{delocalization error}
    \acro{DFT}{density functional theory}
    \acro{DF}{density-fitting}
    \acro{DIIS}{direct inversion in the iterative subspace}
    \acro{DMRG}{density matrix renormalization group}
    \acro{DNN}{deep \acl{NN}}
    \acro{DOCI}{doubly occupied \acl{CI}}
    \acro{DSRG}{driven similarity renormalization group}
    \acro{EKT}{extended Koopmans theorem}
    \acro{ELU}{exponential linear unit}
    \acro{ERI}{electron-repulsion integral}
    \acro{EUE}{effectively unpaired electron}
    \acro{FC}{fractional calculus}
    \acro{FCI}{full \acl{CI}}
    \acro{FP-1}{frontier partition with one set of interspace excitations}
    \acro{FRELU}[FReLU]{flexible \acl{RELU}}
    \acro{FSE}{fractional \acl{SE}}
    \acro{ftBLYP}{fully \acl{tBLYP}}
    \acro{ftPBE}{fully \acl{tPBE}}
    \acro{ftSVWN3}{fully \acl{tSVWN3}}
    \acro{ft}{full translation}
    \acro{GASSCF}{generalized active-space \acl{SCF}}
    \acro{GELU}{Gaussian error linear unit}
    \acro{GGA}{generalized gradient approximation}
    \acro{GL}{Gr\"unwald-Letnikov}
    \acro{GMCPDFT}[G-MC-PDFT]{generalized \acl{MCPDFT}}
    \acro{GPU}{graphics processing unit}
    \acro{GTO}{Gaussian-type orbital}
    \acro{HF}{Hartree-Fock}
    \acro{HISS}{Henderson-Izmaylov-Scuseria-Savin}
    \acrodef{HK}{Hohenberg-Kohn}
    \acro{HOMO}{highest-occupied \acl{MO}}
    \acro{HONO}{highest-occupied \acl{NO}}
    \acro{HPC}{high-performance computing}
    \acro{HPDFT}{hybrid \acl{PDFT}}
    \acro{HRDM}{hole \acl{RDM}}
    \acro{HSE}{Heyd-Scuseria-Ernzerhof}
    \acro{HXC}{Hartree-\acl{XC} }
    \acro{IPEA}{ionization potential electron affinity}
    \acro{IPSDP}{interior-point \acl{SDP}}
    \acro{KDFT}{kinetic \acl{DFT}}
    \acro{KSDFT}[KS-DFT]{\acl{KS} \acl{DFT}}
    \acro{KS}{Kohn-Sham}
    \acro{LBFGS}[L-BFGS]{limited-memory Broyden-Fletcher-Goldfarb-Shanno}
    \acro{LC}{long-range corrected}
    \acro{LC-VV10}{\acl{LC} Vydrov-van Voorhis 10}
    \acro{l-DFVB}[$\la$-DFVB]{$\la$-density functional \acl{VB}}
    \acro{LEB}{local energy balance}
    \acro{LE}{localization error}
    \acrodef{lftBLYP}[$\la$-\acs{ftBLYP}]{$\la$-\acl{ftBLYP}}
    \acrodef{lftPBE}[$\la$-\acs{ftPBE}]{$\la$-\acl{ftPBE}}
    \acrodef{lftrevPBE}[$\la$-\acs{ftrevPBE}]{$\la$-\acl{ftrevPBE}}
    \acrodef{lftSVWN3}[$\la$-\acs{ftSVWN3}]{$\la$-\acl{ftSVWN3}}
    \acro{lMCPDFT}[$\la$-MC-PDFT]{\acl{MC} \acl{1H-PDFT}}
    \acro{LMF}{local mixing function}
    \acro{LO}{Lieb-Oxford}
    \acro{LP}{linear programming}
    \acro{LR}{long-range}
    \acro{LRELU}[LReLU]{leaky \acl{RELU}}
    \acro{LSDA}{local spin-density approximation}
    \acrodef{ltBLYP}[$\la$-\acs{tBLYP}]{$\la$-\acl{tBLYP}}
    \acrodef{ltPBE}[$\la$-\acs{tPBE}]{$\la$-\acl{tPBE}}
    \acrodef{ltrevPBE}[$\la$-\acs{trevPBE}]{$\la$-\acl{trevPBE}}
    \acrodef{ltSVWN3}[$\la$-\acs{tSVWN3}]{$\la$-\acl{tSVWN3}}
    \acro{LUMO}{lowest-unoccupied \acl{MO}}
    \acro{LUNO}{lowest-unoccupied \acl{NO}}
    \acro{LYP}{Lee-Yang-Parr}
    \acro{M06}{Minnesota 06}
    \acro{M06-2X}{\acl{M06} with double non-local exchange}
    \acro{M06-L}{\acl{M06} local}
    \acro{MAEPE}[$\MAEe$]{\acl{MAE} per electron}
    \acro{MAE}{mean absolute error}
    \acro{MAKEE}[$\MAKEE$]{mean absolute kinetic energy error per electron}
    \acro{MAX}{maximum absolute error}
    \acro{MC1H-PDFT}{\acl{MC} \acl{1H-PDFT}}
    \acro{MC1H}{\acl{MC} one-parameter hybrid \acl{PDFT}}
    \acro{MCHPDFT}{\acl{MC} hybrid-\acl{PDFT}}
    \acro{MCPDFT}[MC-PDFT]{\acl{MC} \acl{PDFT}}
    \acro{MCRSHPDFT}[$\mu\la$-MCPDFT]{\acl{MC} range-separated hybrid-\acl{PDFT}}
    \acro{MCSCF}{\acl{MC} \acl{SCF}}
    \acrodef{MC}{multiconfiguration}
    \acro{ML}{machine learning}
    \acro{MN15}{Minnesota 15}
    \acro{MNIST}{Modified National Institute of Standards and Technology}
    \acro{MOLSSI}[MolSSI]{Molecular Sciences Software Institute}
    \acro{MO}{molecular orbital}
    \acro{MP2}{second-order M\o ller-Plesset \acl{PT}}
    \acro{MR-AQCC}{\acl{MR}-averaged quadratic \acl{CC}}
    \acrodef{MR}{multireference}
    \acro{MS0}{MS0 meta-GGA exchange and revTPSS GGA correlation}
    \acro{NGA}{nonseparable gradient approximation}
    \acro{NIAD}{normed integral absolute deviation}
    \acro{NLRELU}[NLReLU]{natural logarithm \acl{RELU}}
    \acro{NN}{neural network}
    \acro{NO}{natural orbital}
    \acro{NOON}{\acl{NO} \acl{ON}}
    \acro{NPE}{non-parallelity error}
    \acro{NSF}{National Science Foundation}
    \acro{OEP}{optimized effective potential}
    \acro{oMCPDFT}[$\om$-MC-PDFT]{range-separated \acl{MC} \acl{1H-PDFT}}
    \acrodef{ON}{occupation number}
    \acro{ORMAS}{occupation-restricted multiple active-space}
    \acro{OTPD}{on-top pair-density}
    \acro{PBE0}{hybrid-\acs{PBE}}
    \acro{PBE}{Perdew-Burke-Ernzerhof}
    \acro{pCCD-lDFT}[pCCD-$\la$DFT]{\acl{pCCD} $\la$\acs{DFT}}
    \acro{pCCD}{pair coupled-cluster doubles}
    \acro{PDFT}{pair-\acl{DFT}}
    \acro{PEC}{potential energy curve}
    \acro{PES}{potential energy surface}
    \acro{PKZB}{Perdew-Kurth-Zupan-Blaha}
    \acro{pp-RPA}{particle-particle \acl{RPA}}
    \acro{PRELU}[PReLU]{parametric \acl{RELU}}
    \acro{PT}{perturbation theory}
    \acro{PT2}{second-order \acl{PT}}
    \acro{PW91}{Perdew-Wang 91}
    \acro{QTAIM}{quantum theory of atoms in molecules}
    \acro{RASSCF}{restricted active-space \acl{SCF}}
    \acro{RBM}{restricted Boltzmann machines}
    \acro{RDM}{reduced density matrix}
    \acro{RELU}[ReLU]{rectified linear unit}
    \acro{revPBE}{revised \acs{PBE}}
    \acro{RL}{Riemann-Liouville}
    \acro{RMSD}{root mean square deviation}
    \acro{RNN}{recurrent \acl{NN}}
    \acro{RPA}{random-phase approximation}
    \acro{RRELU}[RReLU]{randomized \acl{RELU}}
    \acro{RSH}{range-separated hybrid}
    \acro{SCAN}{strongly constrained and appropriately normed}
    \acro{SCF}{self-consistent field}
    \acro{SDP}{semidefinite programming}
    \acro{SE}{Schr\"odinger equation}
    \acro{SELU}{scaled \acl{ELU}}
    \acro{SF-CCSD}{\acl{SF}-\acl{CCSD}}
    \acro{SF}{spin-flip}
    \acro{SGD}{stochastic gradient descent}
    \acro{SIE}{self-interaction error}
    \acrodef{SI}{Supporting Information}
    \acro{SNIAD}{spherical \acl{NIAD}}
    \acro{SOGGA11}{second-order \acl{GGA}}
    \acro{SR}{short-range}
    \acro{SRELU}[SReLU]{S-shaped \acl{RELU}}
    \acro{STO}{Slater-type orbital}
    \acro{SVWN3}{Slater and Vosko-Wilk-Nusair random-phase approximation expression III}
    \acro{tBLYP}{translated \acl{BLYP}}
    \acro{TMAE}[$\MAE$]{total \acl{MAE} per electron}
    \acro{TNIAD}[$\NIAD$]{total normed integral absolute deviation}
    \acro{tPBE}{translated \acl{PBE}}
    \acro{TPSS}{Tao-Perdew-Staroverov-Scuseria}
    \acro{trevPBE}{translated \acs{revPBE}}
    \acro{tr}{conventional translation}
    \acro{tSVWN3}{translated \acl{SVWN3}}
    \acro{TS}{transition state}
    \acro{v2RDM-CASSCF-PDFT}{\acl{v2RDM} \acl{CASSCF} \acl{PDFT}}
    \acro{v2RDM-CASSCF}{\acl{v2RDM}-driven \acl{CASSCF}}
    \acro{v2RDM-CAS}{\acl{v2RDM}-driven \acl{CAS}}
    \acro{v2RDM-DOCI}{\acl{v2RDM}-\acl{DOCI}}
    \acro{v2RDM}{variational \acl{2-RDM}}
    \acro{VB}{valence bond}
    \acro{VO}{variable-order}
    \acro{wB97X}[$\omega$B97X]{$\omega$B97X}
    \acro{WFT}{wave function theory}
    \acro{WF}{wave function}
    \acro{XC}{exchange-correlation}
    \acro{ZPE}{zero-point energy}
    \acro{ZPVE}{zero-point vibrational energy}
    \end{acronym} 

\newpage

{\bf References}
\vspace{-28pt}

\bibliography{ms}

\end{document}


\author{Mohammad Mostafanejad}
\affiliation{Department of Chemistry, Virginia Tech, Blacksburg, Virginia 24061, USA}
\affiliation{Molecular Sciences Software Institute, Blacksburg, Virginia 24060, USA}

\title{Supplemental Materials: \\[5pt] Unification of popular artificial neural network activation functions}
\maketitle
\tableofcontents

\newpage
\section{Training LeNet-5 on the MNIST dataset}
Tables \ref{SITAB:MNISTA100} and \ref{SITAB:MNIST1650} show the performance results of training
and validation of LeNet-5 neural network on \ac{MNIST} dataset using \nvidia\ A100 80GB PCIe 
and GeForce GTX 1650 \acp{GPU}, respectively. Each training session is carried out for 10 epochs
and repeated 20 times with randomly initialized network parameters. We ensured each activation
function sees the same set of parameters in each round of trainings within the ensemble. 
The details of the experimental setup and the chosen hyperparameters can be found 
in the Computational details section of the manuscript.

\begin{table*}[!htbp]
    \centering
    \setlength{\tabcolsep}{3pt}
    \setlength{\extrarowheight}{1pt}
    \caption{The best performance metrics and timings for training and testing LeNet-5 neural 
             network on the MNIST dataset using various activation functions$^{a,b}$}
    \label{SITAB:MNISTA100}
    \resizebox{\textwidth}{!}{
        \begin{tabular}{lcccccc}
            \hline\hline
            Activation Function$^c$ & Loss            & Accuracy (\%) & Precision (\%) & Recall (\%)   & F1 Score (\%) & Wall-Clock Time (s) \\
            \hline
            \mr{2}{*}{Sigmoid}  & 0.0270 (0.0290) & 99.13 (99.13) & 99.13 (99.13) & 99.13 (99.12) & 99.13 (99.12) & 21.93               \\
                                & 0.0270 (0.0290) & 99.13 (99.13) & 99.13 (99.13) & 99.13 (99.12) & 99.13 (99.13) & 30.39               \\[3pt]
            \mr{2}{*}{Swish-1}  & 0.0411 (0.0295) & 98.73 (98.98) & 98.72 (98.98) & 98.71 (98.96) & 98.72 (98.97) & 28.38               \\
                                & 0.0422 (0.0295) & 98.69 (98.99) & 98.68 (98.98) & 98.68 (98.97) & 98.68 (98.98) & 30.60               \\[3pt]
            \mr{2}{*}{SoftSign} & 0.0138 (0.0324) & 99.56 (99.00) & 99.56 (99.00) & 99.55 (98.99) & 99.56 (98.99) & 28.39               \\
                                & 0.0137 (0.0315) & 99.56 (99.03) & 99.56 (99.03) & 99.56 (99.02) & 99.56 (99.02) & 28.61               \\[3pt]
            \mr{2}{*}{$\tanh$}  & 0.0146 (0.0329) & 99.52 (98.95) & 99.52 (98.94) & 99.52 (98.94) & 99.52 (98.94) & 20.77               \\
                                & 0.0126 (0.0325) & 99.58 (99.01) & 99.58 (99.00) & 99.58 (99.00) & 99.58 (99.00) & 29.72               \\[3pt]
            \mr{2}{*}{Mish}     & 0.0279 (0.0299) & 99.14 (99.07) & 99.14 (99.06) & 99.14 (99.06) & 99.14 (99.06) & 24.38               \\
                                & 0.0339 (0.0304) & 98.96 (99.03) & 98.96 (99.03) & 98.95 (99.02) & 98.96 (99.02) & 40.73               \\[3pt]
            \hline
            ReLU$^d$            & 0.0144 (0.0292) & 99.54 (99.16) & 99.54 (99.15) & 99.54 (99.15) & 99.54 (99.15) & 23.78               \\
            \hline\hline
        \end{tabular}
    }
    \begin{tablenotes}
        \scriptsize
        \item $^a$ All results are ensemble averages over 20 independent training and testing experiments.
        \item $^b$ All calculations are performed using a \nvidia\ A100 80GB PCIe \acs{GPU}.
        \item $^c$ The test results are given in parentheses. The first and second rows in each activation function 
                   entry correspond to the \\ \quad built-in and gated representations, respectively.
        \item $^d$ The LeNet-5 neural network architecture with \acs{RELU} activation functions is 
                   taken as base architecture.
    \end{tablenotes}
\end{table*}

\begin{table*}[!htbp]
    \centering
    \setlength{\tabcolsep}{3pt}
    \setlength{\extrarowheight}{1pt}
    \caption{The Best performance metrics and timings for training and testing LeNet-5 neural 
             network on the MNIST dataset using various activation functions$^{a,b}$}
    \label{SITAB:MNIST1650}
    \resizebox{\textwidth}{!}{
        \begin{tabular}{lcccccc}
            \hline\hline
            Activation Function$^c$ & Loss            & Accuracy (\%) & Precision (\%) & Recall (\%)   & F1 Score (\%) & Wall-Clock Time (s) \\
            \hline
            \mr{2}{*}{Sigmoid}  & 0.0270 (0.0290) & 99.13 (99.13) & 99.13 (99.13)  & 99.13 (99.12) & 99.13 (99.12) & 60.18               \\
                                & 0.0270 (0.0290) & 99.13 (99.13) & 99.13 (99.13)  & 99.13 (99.12) & 99.13 (99.13) & 70.23               \\[3pt]
            \mr{2}{*}{Swish-1}  & 0.0411 (0.0294) & 98.72 (98.99) & 98.72 (98.99)  & 98.71 (98.97) & 98.71 (98.98) & 69.51               \\
                                & 0.0383 (0.0294) & 98.81 (99.00) & 98.80 (98.99)  & 98.80 (98.98) & 98.80 (98.99) & 74.38               \\[3pt]
            \mr{2}{*}{SoftSign} & 0.0122 (0.0316) & 99.60 (99.02) & 99.60 (99.02)  & 99.60 (99.01) & 99.60 (99.01) & 65.32               \\
                                & 0.0134 (0.0319) & 99.56 (99.04) & 99.56 (99.03)  & 99.56 (99.03) & 99.56 (99.03) & 64.09               \\[3pt]
            \mr{2}{*}{$\tanh$}  & 0.0165 (0.0327) & 99.46 (98.94) & 99.46 (98.93)  & 99.46 (98.93) & 99.46 (98.93) & 58.72               \\
                                & 0.0142 (0.0326) & 99.54 (98.97) & 99.54 (98.96)  & 99.54 (98.96) & 99.54 (98.96) & 71.09               \\[3pt]
            \mr{2}{*}{Mish}     & 0.0323 (0.0301) & 99.01 (99.05) & 99.00 (99.05)  & 99.00 (99.04) & 99.00 (99.04) & 59.73               \\
                                & 0.0306 (0.0300) & 99.06 (99.04) & 99.05 (99.04)  & 99.05 (99.03) & 99.05 (99.03) & 91.29               \\[3pt]
            \hline
            ReLU$^d$            & 0.0137 (0.0291) & 99.56 (99.16) & 99.56 (99.16)  & 99.56 (99.15) & 99.56 (99.15) & 63.77               \\
            \hline\hline
        \end{tabular}
    }
    \begin{tablenotes}
        \scriptsize
        \item $^a$ All results are ensemble averages over 20 independent training and testing experiments.
        \item $^b$ All calculations are performed using a \nvidia\ GeForce GTX 1650 \acs{GPU}.
        \item $^c$ The test results are given in parentheses. The first and second rows in each activation function 
                   entry correspond to the \\ \quad built-in and gated representations, respectively.
        \item $^d$ The LeNet-5 neural network architecture with \acs{RELU} activation functions is 
                   taken as base architecture.
    \end{tablenotes}
\end{table*}
\pagebreak
\section{Training LeNet-5 on the CIFAR-10 Dataset}
Tables \ref{SITAB:CIFAR10A100} and \ref{SITAB:CIFAR101650} show the performance results of 
training and validation of LeNet-5 neural network on CIFAR-10 dataset using \nvidia\ A100
80GB PCIe and GeForce GTX 1650 \acp{GPU}, respectively. Each training is carried out 
for 20 epochs and repeated 20 times with randomly initialized network parameters. We ensured
each activation function sees the same set of parameters in each round of trainings over the
ensemble. The details of the experimental setup and the chosen hyperparameters can be found 
in the Computational details section of the manuscript.

\begin{table*}[!htbp]
    \centering
    \setlength{\tabcolsep}{3pt}
    \setlength{\extrarowheight}{1pt}
    \caption{The best performance metrics and timings for training and testing LeNet-5 neural 
    network on the CIFAR-10 dataset using various activation functions$^{a,b}$}
    \label{SITAB:CIFAR10A100}
    \resizebox{\textwidth}{!}{
        \begin{tabular}{lcccccc}
            \hline\hline
            Activation Function$^c$ & Loss            & Accuracy (\%) & Precision (\%) & Recall (\%)   & F1 Score (\%) & Wall-Clock Time (s) \\
            \hline
            \mr{2}{*}{Sigmoid}  & 0.7353 (1.0151) & 74.79 (65.39) & 74.68 (65.27) & 74.79 (65.39) & 74.73 (65.24) & 44.26               \\
                                & 0.7353 (1.0152) & 74.78 (65.38) & 74.68 (65.25) & 74.78 (65.38) & 74.72 (65.22) & 54.97               \\[3pt]
            \mr{2}{*}{Swish-1}  & 0.7323 (0.9016) & 74.55 (69.39) & 74.44 (69.34) & 74.55 (69.39) & 74.48 (69.13) & 54.44               \\
                                & 0.7322 (0.9018) & 74.55 (69.40) & 74.44 (69.35) & 74.55 (69.40) & 74.49 (69.13) & 60.77               \\[3pt]
            \mr{2}{*}{SoftSign} & 0.6380 (0.9297) & 77.89 (68.64) & 77.77 (68.70) & 77.89 (68.64) & 77.82 (68.55) & 53.09               \\
                                & 0.6280 (0.9313) & 78.26 (68.56) & 78.15 (68.63) & 78.26 (68.56) & 78.20 (68.49) & 54.52               \\[3pt]
            \mr{2}{*}{$\tanh$}  & 0.8129 (0.9766) & 71.63 (66.79) & 71.41 (66.75) & 71.63 (66.79) & 71.51 (66.44) & 45.58               \\
                                & 0.8015 (0.9738) & 72.05 (66.90) & 71.84 (66.90) & 72.05 (66.90) & 71.93 (66.57) & 56.09               \\[3pt]
            \mr{2}{*}{Mish}     & 0.6842 (0.8914) & 76.18 (69.40) & 76.08 (69.55) & 76.18 (69.40) & 76.12 (69.26) & 48.37               \\
                                & 0.6841 (0.8913) & 76.18 (69.44) & 76.08 (69.58) & 76.18 (69.44) & 76.12 (69.30) & 78.09               \\[3pt]
            \hline
            ReLU$^d$            & 0.7299 (0.9279) & 74.57 (68.24) & 74.43 (68.56) & 74.57 (68.24) & 74.49 (67.99) & 45.08              \\
            \hline\hline
        \end{tabular}
    }
    \begin{tablenotes}
        \scriptsize
        \item $^a$ All results are ensemble averages over 20 independent training and testing experiments.
        \item $^b$ All calculations are performed using a \nvidia\ A100 80GB PCIe \acs{GPU}.
        \item $^c$ The test results are given in parentheses. The first and second rows in each activation function 
                   entry correspond to the \\ \quad built-in and gated representations, respectively.
        \item $^d$ The LeNet--5 neural network architecture with \acs{RELU} activation functions is 
                   taken as base architecture.
    \end{tablenotes}
  \end{table*}

\begin{table*}[!htbp]
    \centering
    \setlength{\tabcolsep}{3pt}
    \setlength{\extrarowheight}{1pt}
    \caption{The best performance metrics and timings for training and testing LeNet-5 neural 
    network on the CIFAR-10 dataset using various activation functions$^{a,b}$}
    \label{SITAB:CIFAR101650}
    \resizebox{\textwidth}{!}{
        \begin{tabular}{lcccccc}
            \hline\hline
            Activation Function & Loss            & Accuracy (\%) & Precision (\%) & Recall (\%)   & F1 Score (\%) & Wall-Clock Time (s) \\
            \hline
            \mr{2}{*}{Sigmoid}  & 0.7353 (1.0152) & 74.77 (65.37) & 74.66 (65.24)  & 74.77 (65.37) & 74.70 (65.21) & 112.10               \\
                                & 0.7352 (1.0152) & 74.77 (65.37) & 74.67 (65.25)  & 74.77 (65.37) & 74.71 (65.22) & 134.99               \\[3pt]
            \mr{2}{*}{Swish-1}  & 0.7323 (0.9013) & 74.55 (69.43) & 74.44 (69.36)  & 74.55 (69.43) & 74.48 (69.15) & 132.38               \\
                                & 0.7323 (0.9015) & 74.56 (69.40) & 74.44 (69.35)  & 74.56 (69.40) & 74.49 (69.13) & 145.52               \\[3pt]
            \mr{2}{*}{SoftSign} & 0.6332 (0.9284) & 78.03 (68.72) & 77.91 (68.78)  & 78.03 (68.72) & 77.96 (68.64) & 130.23               \\
                                & 0.6378 (0.9303) & 77.92 (68.64) & 77.80 (68.71)  & 77.92 (68.64) & 77.85 (68.55) & 131.39               \\[3pt]
            \mr{2}{*}{$\tanh$}  & 0.8238 (0.9739) & 71.23 (66.83) & 71.00 (66.78)  & 71.23 (66.83) & 71.10 (66.47) & 109.93               \\
                                & 0.8244 (0.9782) & 71.22 (66.69) & 70.99 (66.64)  & 71.23 (66.69) & 71.10 (66.31) & 136.13               \\[3pt]
            \mr{2}{*}{Mish}     & 0.6841 (0.8914) & 76.17 (69.42) & 76.07 (69.59)  & 76.17 (69.42) & 76.11 (69.29) & 115.97               \\
                                & 0.6840 (0.8910) & 76.17 (69.42) & 76.07 (69.57)  & 76.17 (69.42) & 76.11 (69.29) & 194.53               \\[3pt]
            \hline
            ReLU$^c$            & 0.7379 (0.9358) & 74.23 (67.81) & 74.09 (68.16)  & 74.23 (67.81) & 74.15 (67.56) & 54.90               \\
            \hline\hline
        \end{tabular}
    }
    \begin{tablenotes}
        \scriptsize
        \item $^a$ All results are ensemble averages over 20 independent training and testing experiments.
        \item $^b$ All calculations are performed using a \nvidia\ GeForce GTX 1650 \acs{GPU}.
        \item $^c$ The test results are given in parentheses. The first and second rows in each activation function 
                   entry correspond to the \\ \quad built-in and gated representations, respectively.
        \item $^d$ The LeNet--5 neural network architecture with \acs{RELU} activation functions is 
                   taken as base architecture.
    \end{tablenotes}
\end{table*}
\pagebreak
\section{Plots of Training and Validation Losses: MNIST Dataset}
Here, we present plots of training and validation losses pertinent to the individual
training sessions for training LeNet-5 neural network on the \ac{MNIST} dataset. Each plot
includes the loss values of twenty training experiments each running over 10 epochs
using a \nvidia\ A100 80GB PCIe \ac{GPU}. Note that the loss axis is in logarithmic scale.

\begin{figure}[!tbph]
    \centering     
    \includegraphics[scale=0.85]{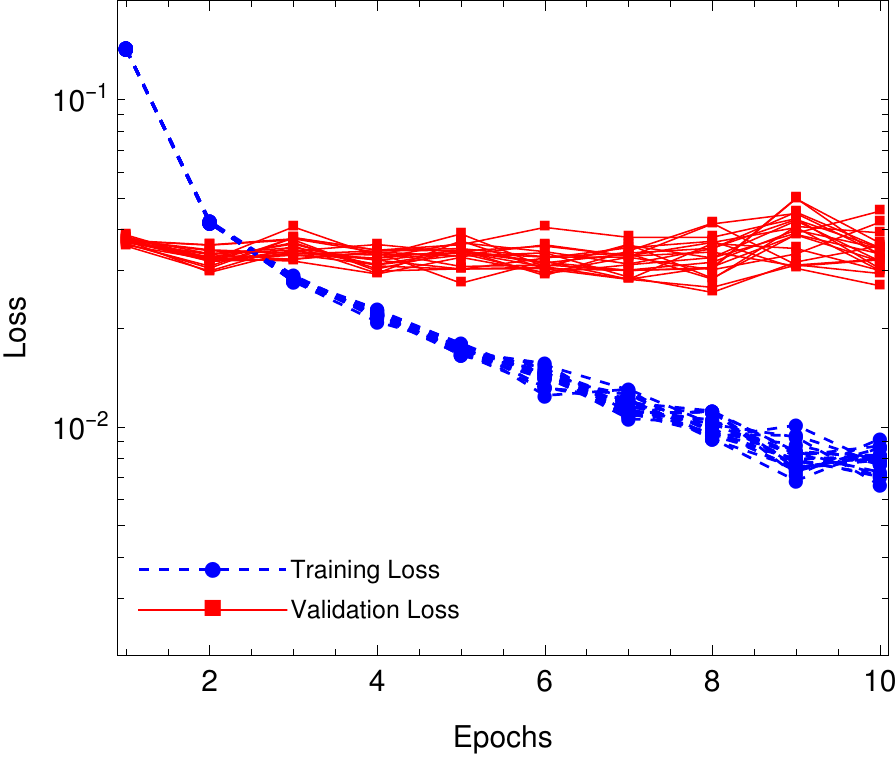}\label{SIFIG:LOSSRELU}
    \caption{Plot of training/validation losses with built-in \acs{RELU} activation function}
\end{figure}

\begin{figure}[!tbph]
    \centering     
    \subfigure[\hspace{-25pt}]{\includegraphics[scale=0.85]{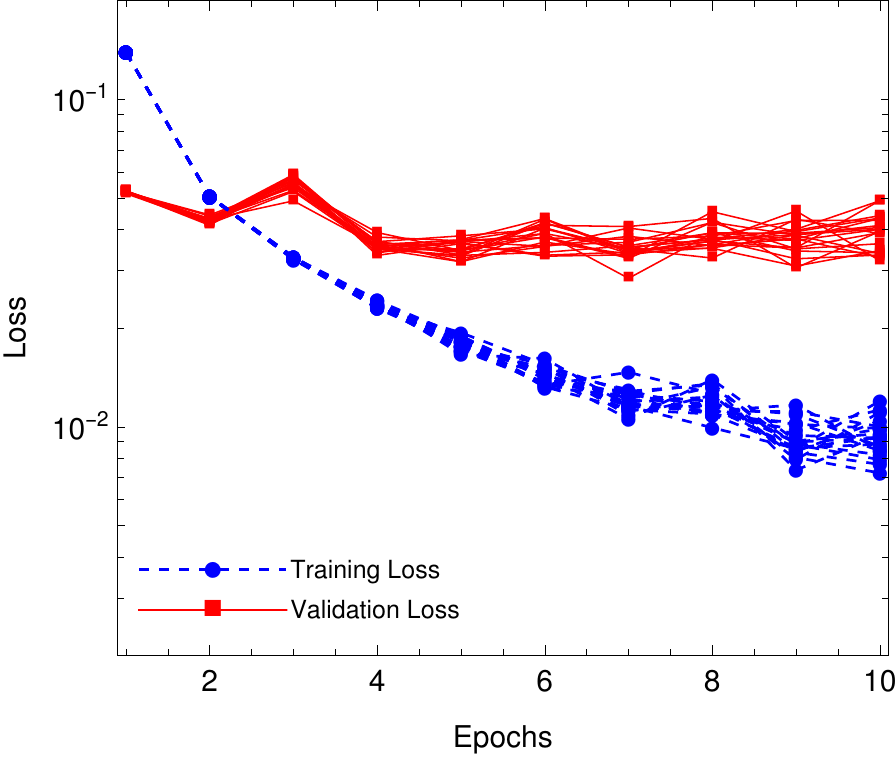}\label{SIFIG:LOSSTANH}}
    \hspace{3pt}
    \subfigure[\hspace{-25pt}]{\includegraphics[scale=0.85]{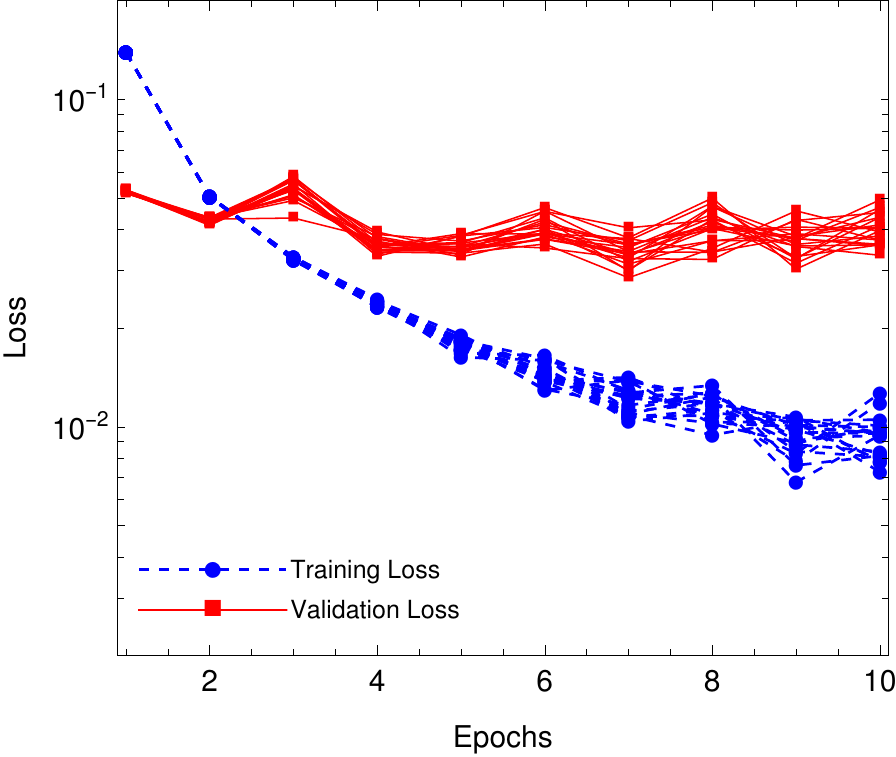}\label{SIFIG:LOSSFTANH}} 
    \\
    \subfigure[\hspace{-25pt}]{\includegraphics[scale=0.85]{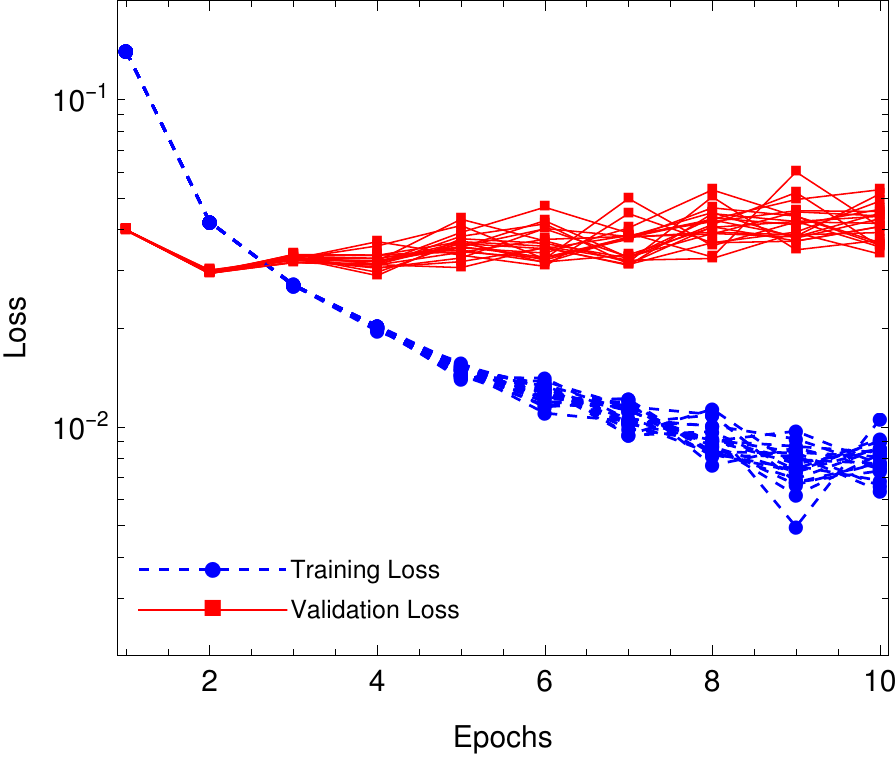}\label{SIFIG:LOSSSWISH1}}
    \hspace{3pt}
    \subfigure[\hspace{-25pt}]{\includegraphics[scale=0.85]{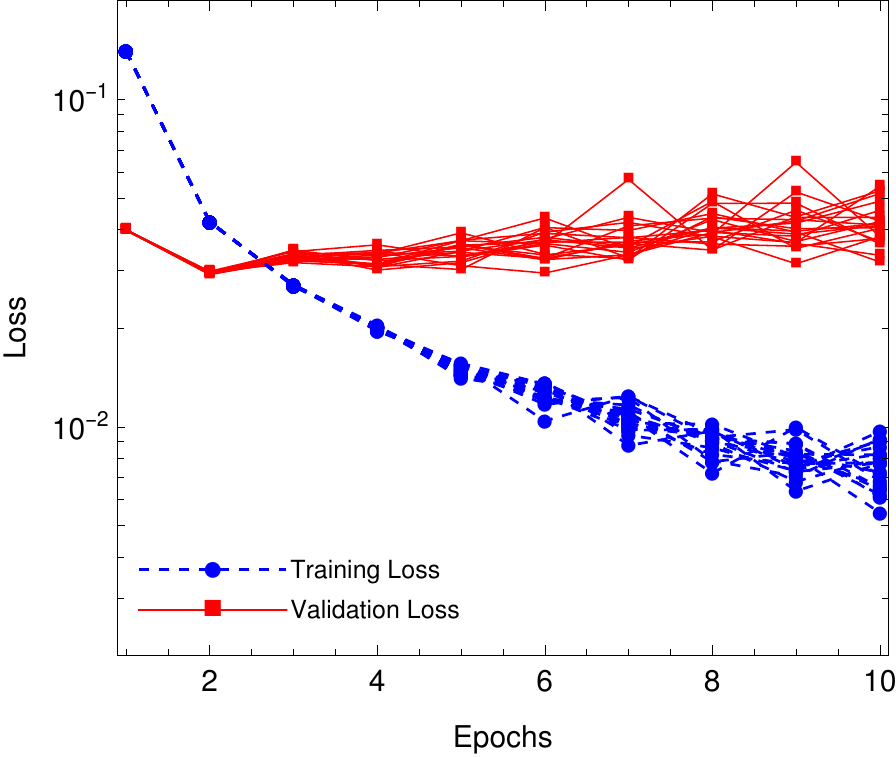}\label{SIFIG:LOSSFSWISH1}}
    \\
    \subfigure[\hspace{-25pt}]{\includegraphics[scale=0.85]{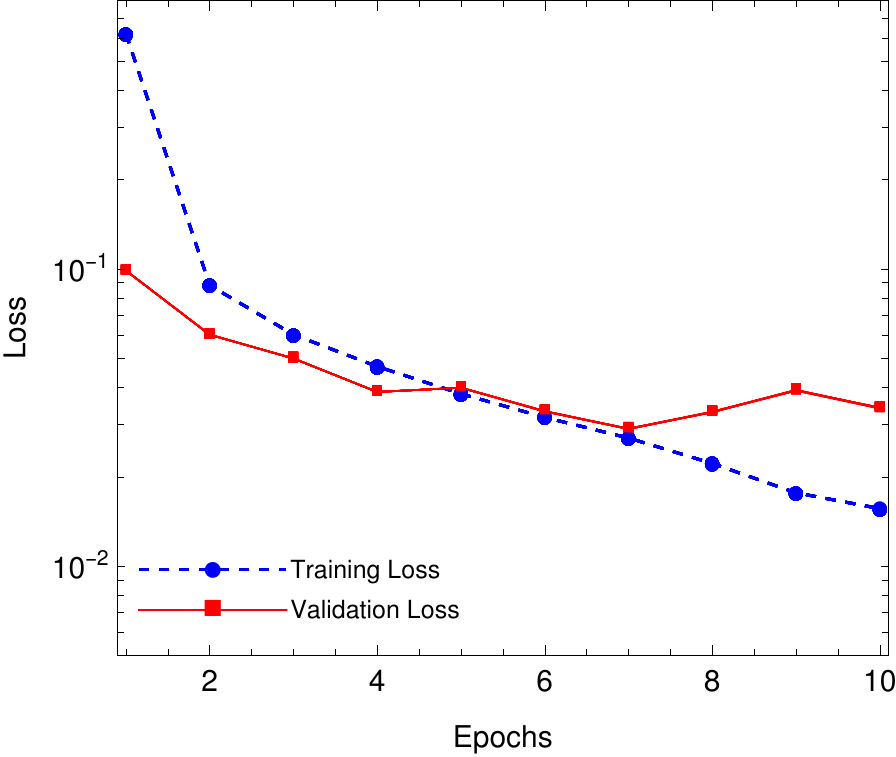}\label{SIFIG:LOSSSIGMOID}}
    \hspace{3pt}
    \subfigure[\hspace{-25pt}]{\includegraphics[scale=0.85]{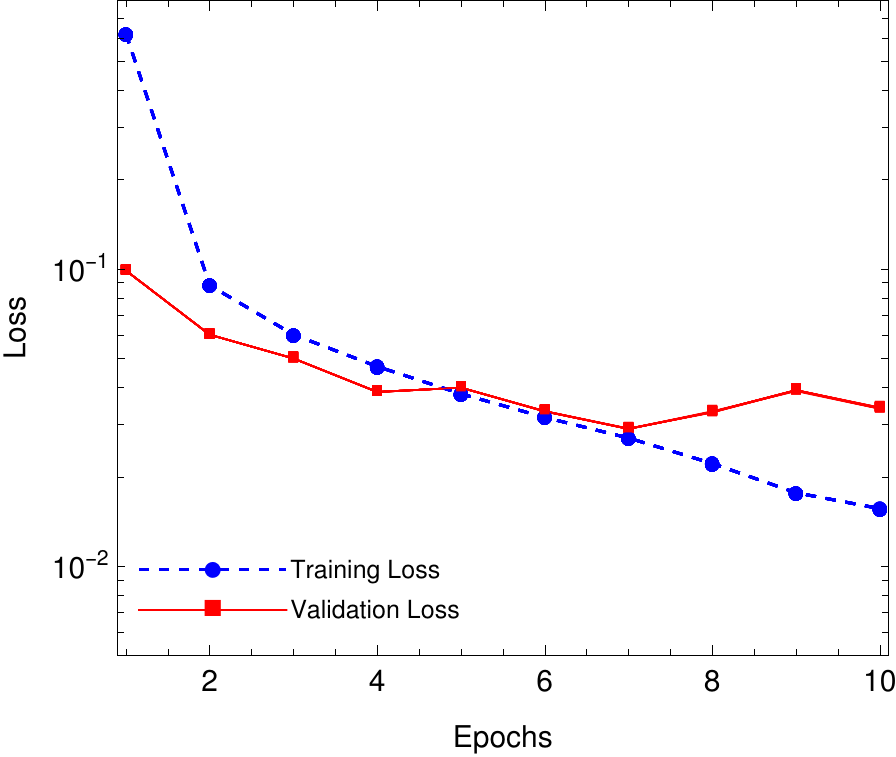}\label{SIFIG:LOSSFSIGMOID}}
    \caption{Plot of training/validation losses with built-in (a), (c) and (e) and gated 
    representations (b), (d) and (f) of $\tanh$, Swish-1 and sigmoid activation functions, respectively}
\end{figure}

\begin{figure}[!tbph]
    \centering     
    \subfigure[\hspace{-25pt}]{\includegraphics[scale=0.85]{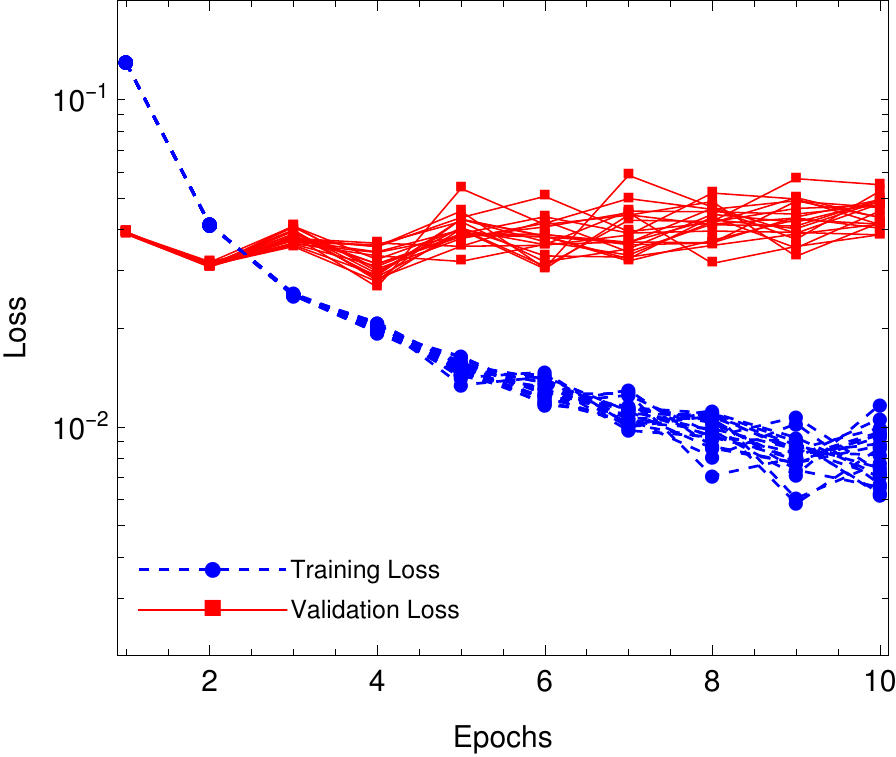}\label{SIFIG:LOSSMISH}}
    \hspace{3pt}
    \subfigure[\hspace{-25pt}]{\includegraphics[scale=0.85]{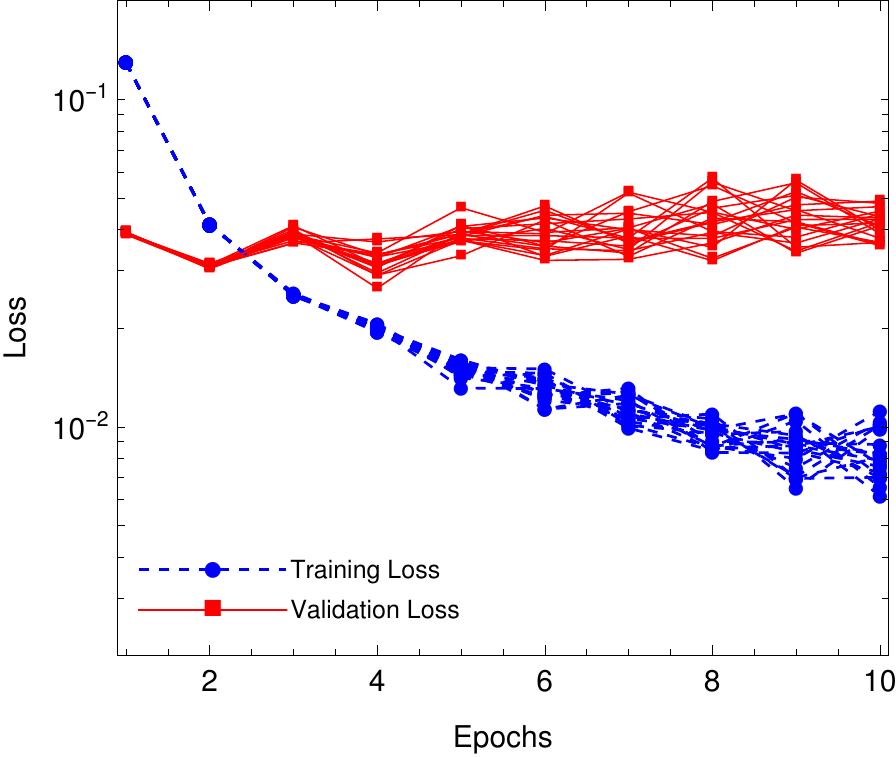}\label{SIFIG:LOSSFMISH}}
    \\
    \subfigure[\hspace{-25pt}]{\includegraphics[scale=0.85]{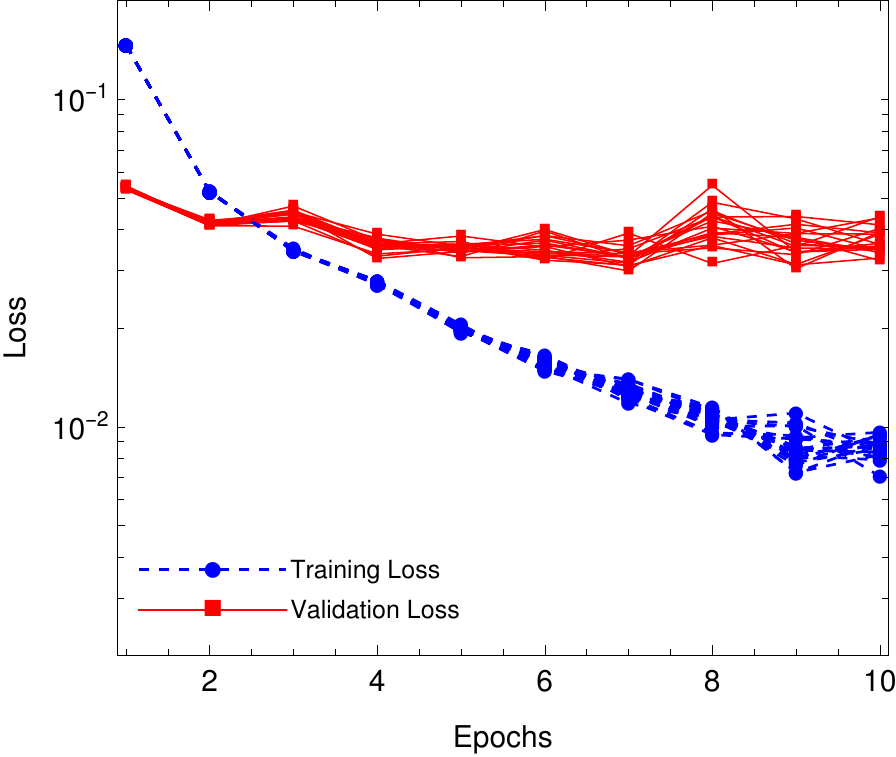}\label{SIFIG:LOSSSOFTSIGN}}
    \hspace{3pt}
    \subfigure[\hspace{-25pt}]{\includegraphics[scale=0.85]{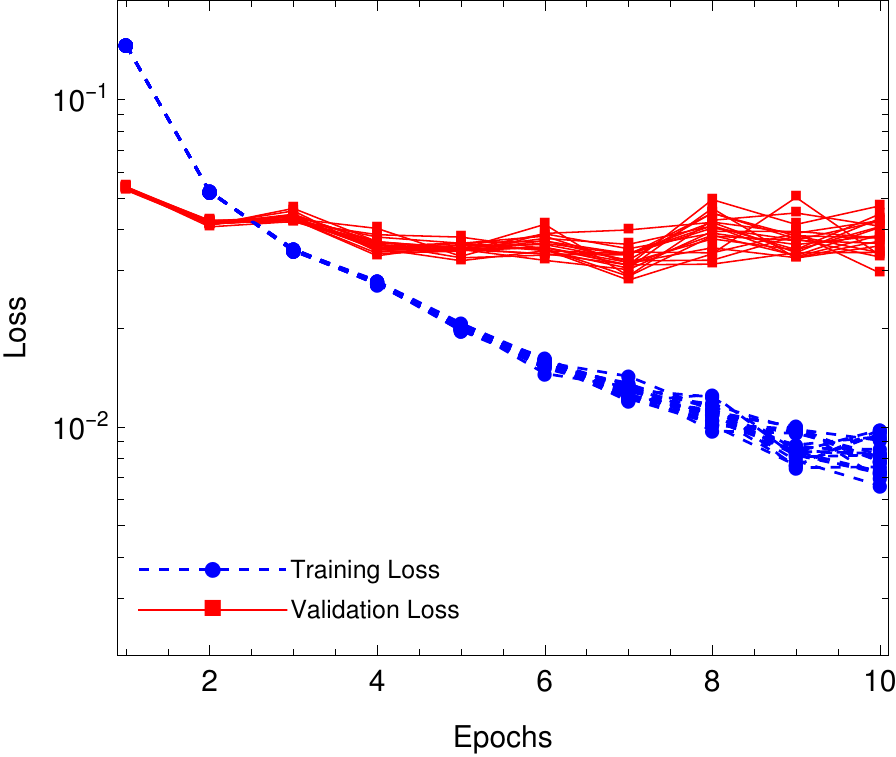}\label{SIFIG:LOSSFSOFTSIGN}}
    \caption{Plot of training/validation losses with built-in (a) and (c) and gated representations (b) 
    and (d) of Mish and Softsign activation functions, respectively}
\end{figure}
\newpage \ \newpage
\section{Plots of Training and Validation Accuracies: MNIST Dataset}
In this section, we illustrate the plots of training and validation accuracies 
corresponding to the individual training sessions for training LeNet-5 neural network
on the \ac{MNIST} dataset. Each plot includes the accuracy values for twenty training 
experiments each running over 10 epochs using a \nvidia\ A100 80GB PCIe \ac{GPU}.

\begin{figure}[!tbph]
    \centering     
    \includegraphics[scale=0.85]{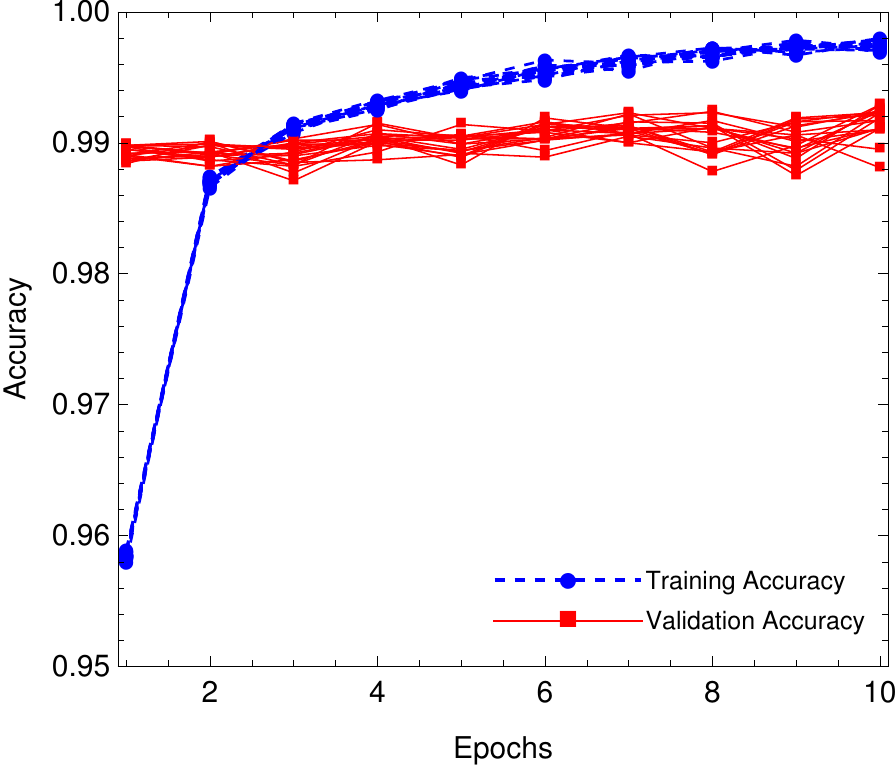}\label{SIFIG:ACCRELU}
    \caption{Plot of training/validation accuracies with built-in \acs{RELU} activation function}
\end{figure}

\begin{figure}[!tbph]
    \centering     
    \subfigure[\hspace{-25pt}]{\includegraphics[scale=0.85]{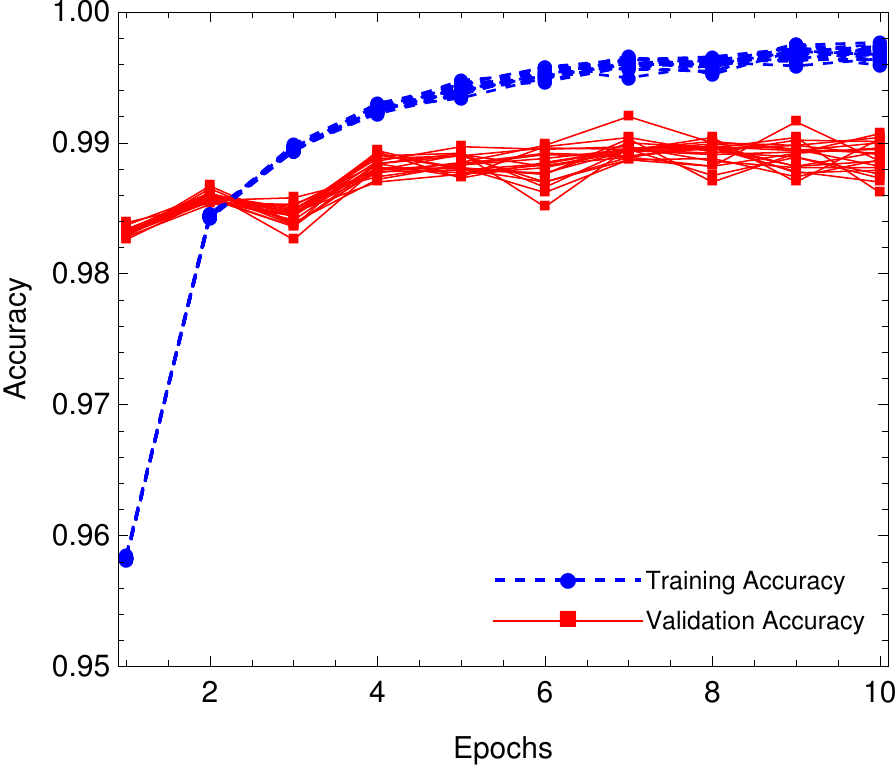}\label{SIFIG:ACCTANH}} 
    \hspace{3pt}
    \subfigure[\hspace{-25pt}]{\includegraphics[scale=0.85]{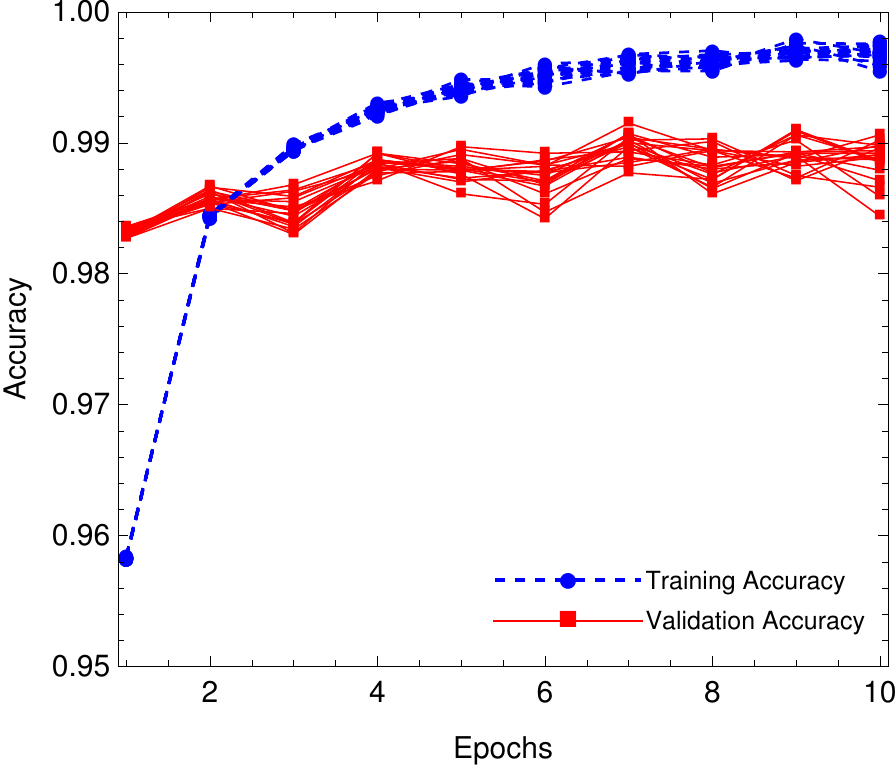}\label{SIFIG:ACCFTANH}} 
    \\
    \subfigure[\hspace{-25pt}]{\includegraphics[scale=0.85]{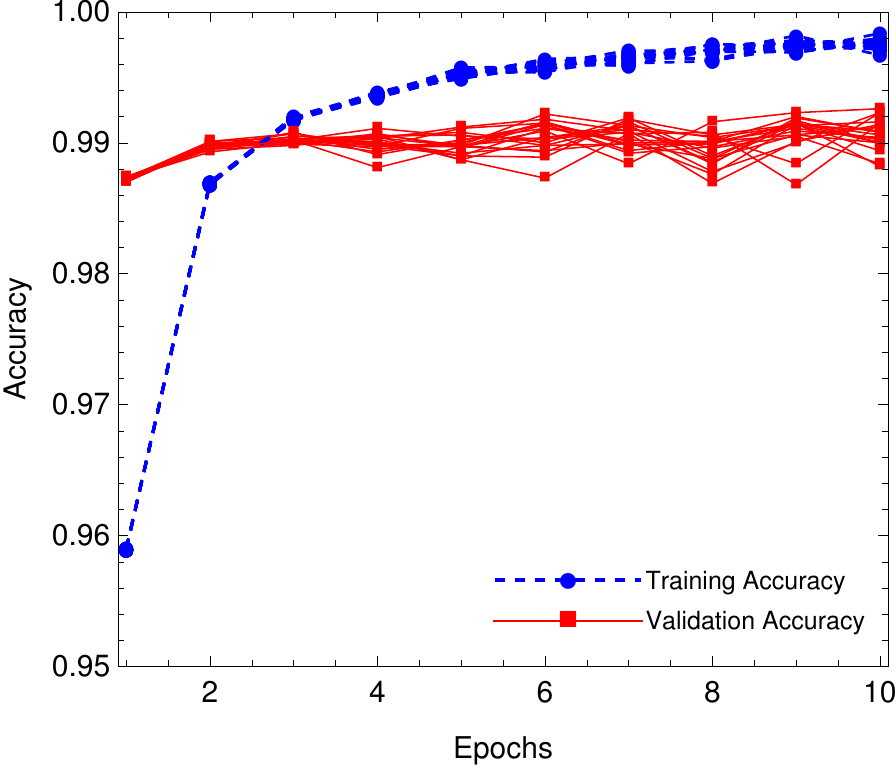}\label{SIFIG:ACCSWISH1}}
    \hspace{3pt}
    \subfigure[\hspace{-25pt}]{\includegraphics[scale=0.85]{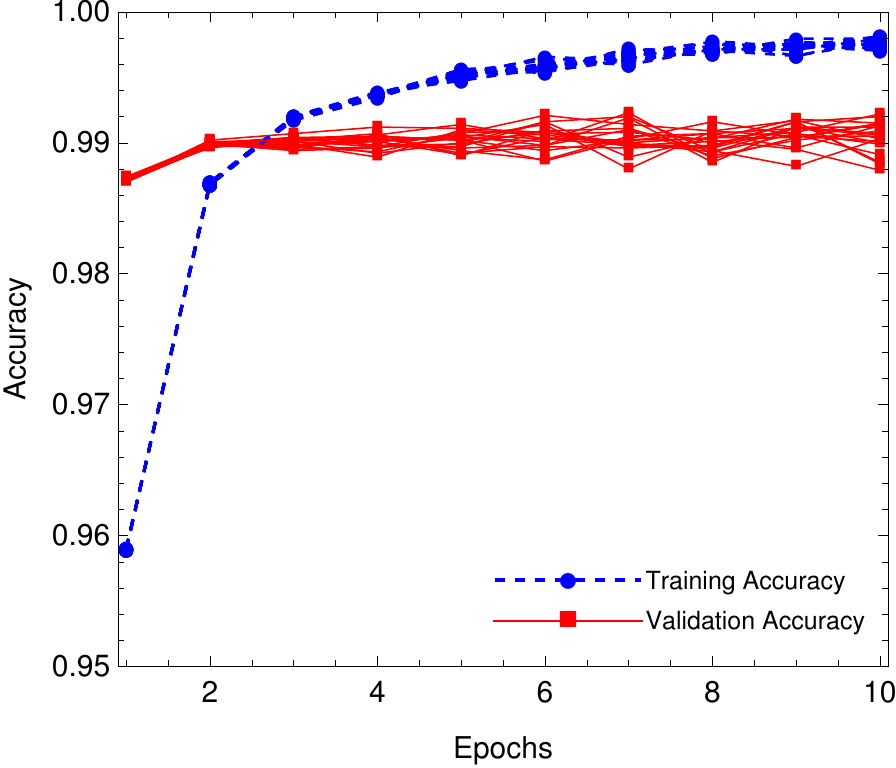}\label{SIFIG:ACCFSWISH1}}
    \\
    \subfigure[\hspace{-25pt}]{\includegraphics[scale=0.85]{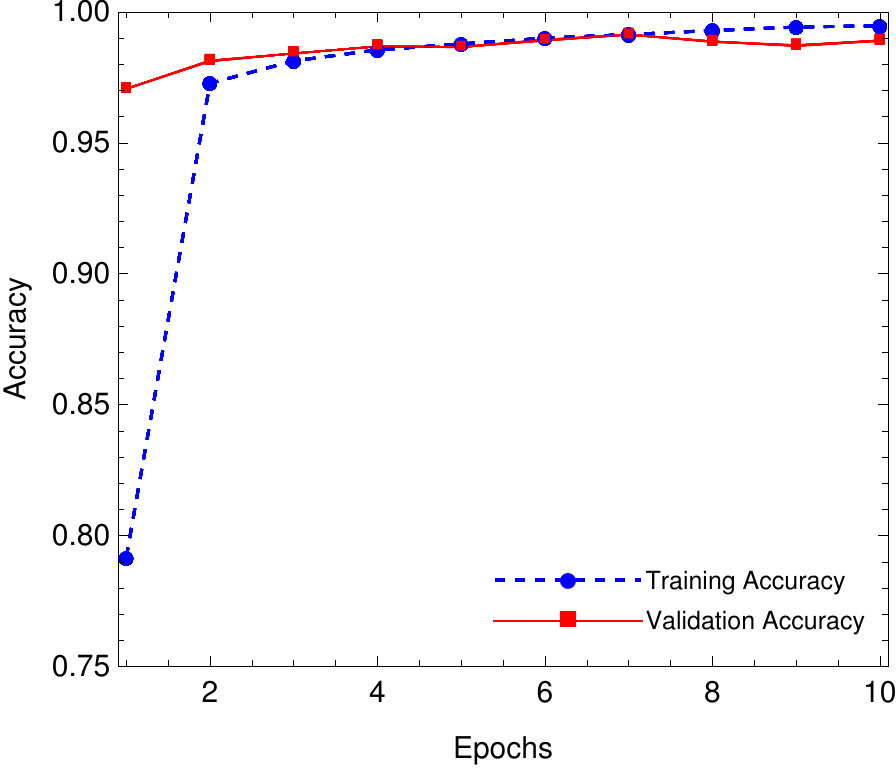}\label{SIFIG:ACCSIGMOID}}
    \hspace{3pt}
    \subfigure[\hspace{-25pt}]{\includegraphics[scale=0.85]{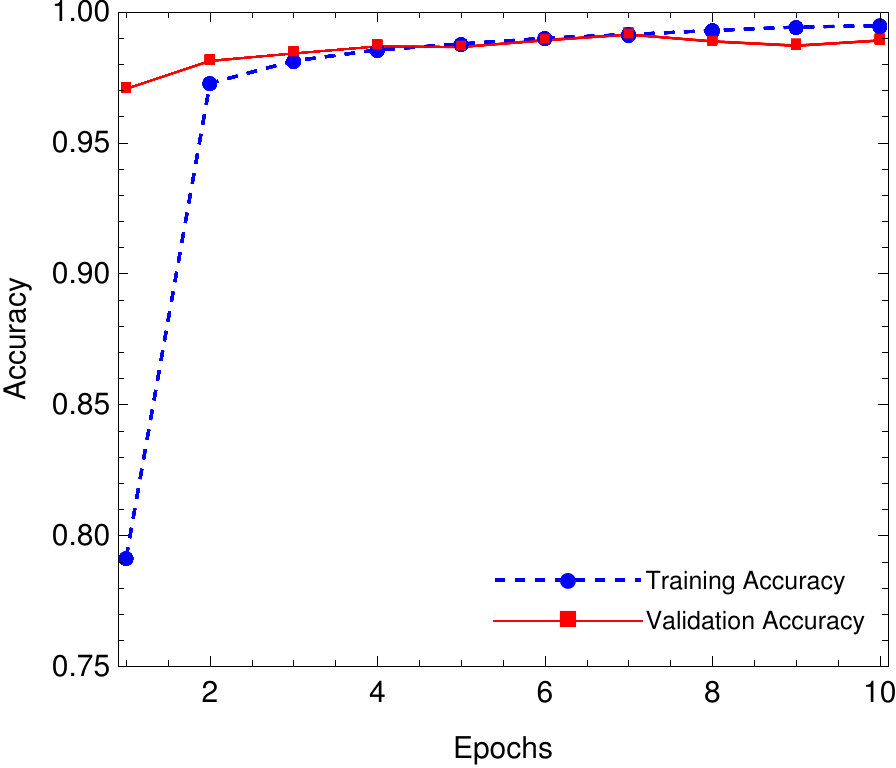}\label{SIFIG:ACCFSIGMOID}}
    \caption{Plot of training/validation accuracies with built-in (a), (c) and (e) and gated 
    representations (b), (d) and (f) of $\tanh$, Swish-1 and sigmoid activation functions, respectively}
\end{figure}

\begin{figure}[!tbph]
    \centering     
    \subfigure[\hspace{-25pt}]{\includegraphics[scale=0.85]{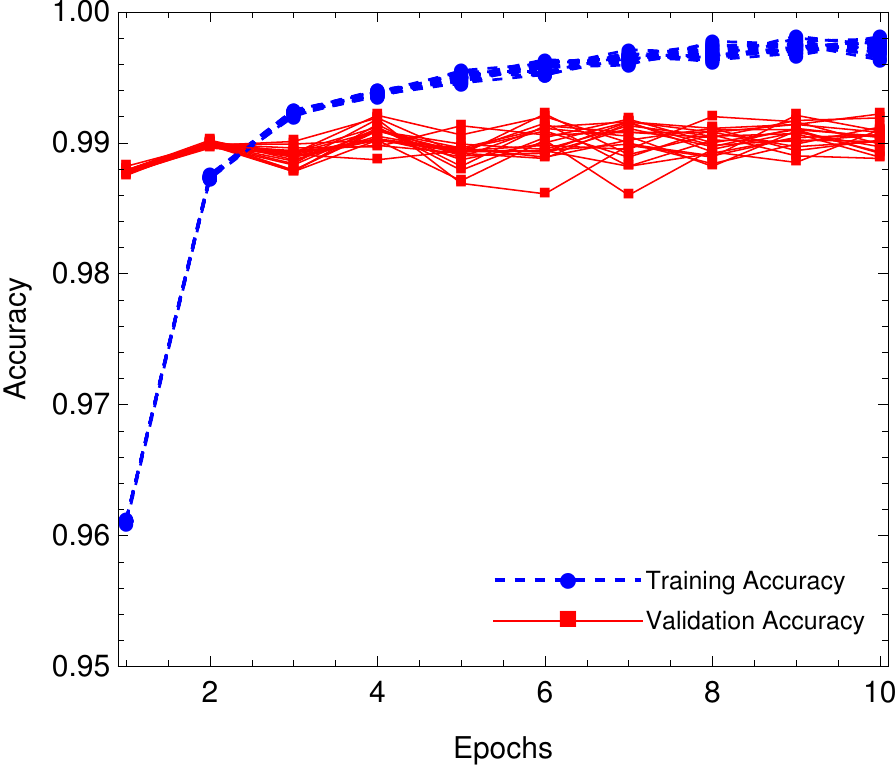}\label{SIFIG:ACCMISH}}
    \hspace{3pt}
    \subfigure[\hspace{-25pt}]{\includegraphics[scale=0.85]{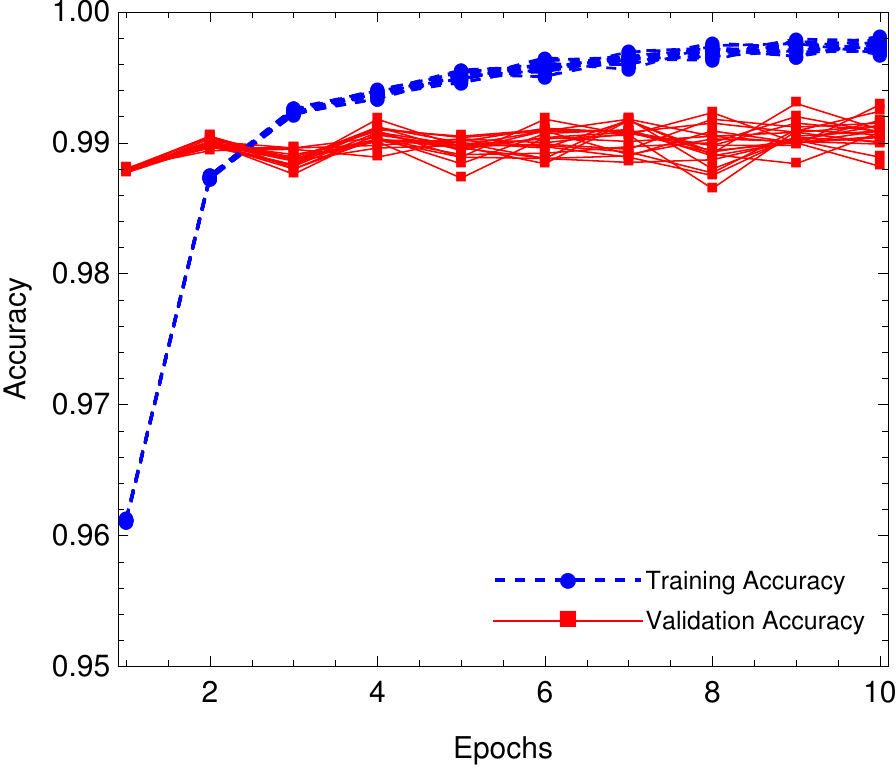}\label{SIFIG:ACCFMISH}}
    \\
    \subfigure[\hspace{-25pt}]{\includegraphics[scale=0.85]{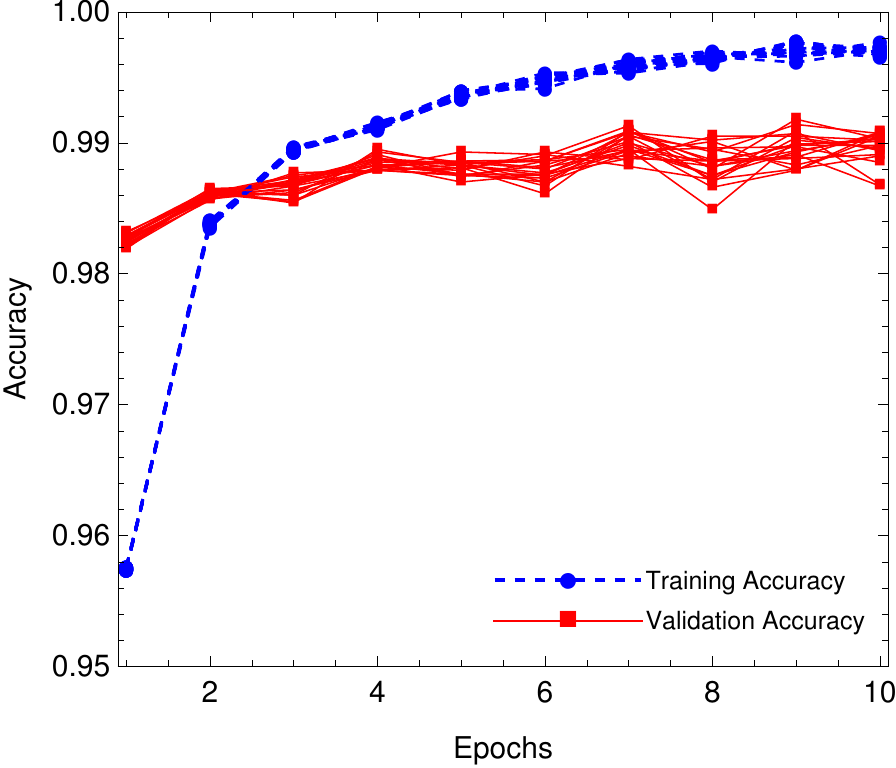}\label{SIFIG:ACCSOFTSIGN}}
    \hspace{3pt}
    \subfigure[\hspace{-25pt}]{\includegraphics[scale=0.85]{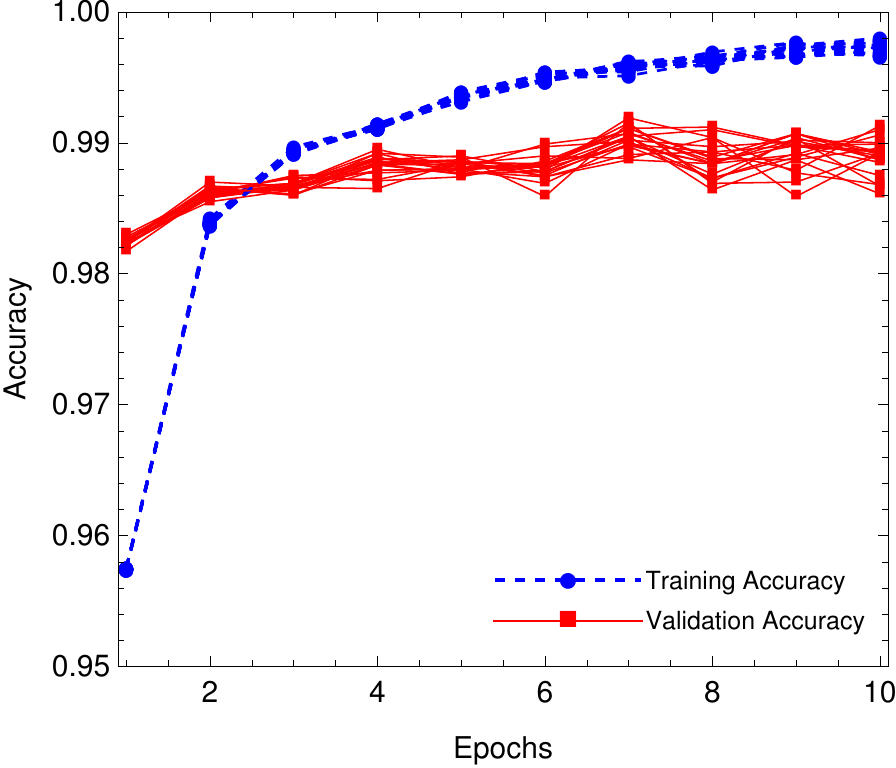}\label{SIFIG:ACCFSOFTSIGN}}
    \caption{Plot of training/validation accuracies with built-in (a) and (c) and gated representations (b) 
    and (d) of Mish and Softsign activation functions, respectively}
\end{figure}
\newpage \ \newpage
\section{Plots of Training and Validation Losses: CIFAR-10 Dataset}
Here, we present plots of training and validation losses pertinent to the individual
training sessions for training LeNet-5 neural network on the CIFAR-10 dataset. Each plot
includes the loss values of twenty training experiments each running over 20 epochs
using a \nvidia\ A100 80GB PCIe \ac{GPU}. Note that the loss axis is in logarithmic scale.

\begin{figure}[!tbph]
    \centering     
    \includegraphics[scale=0.85]{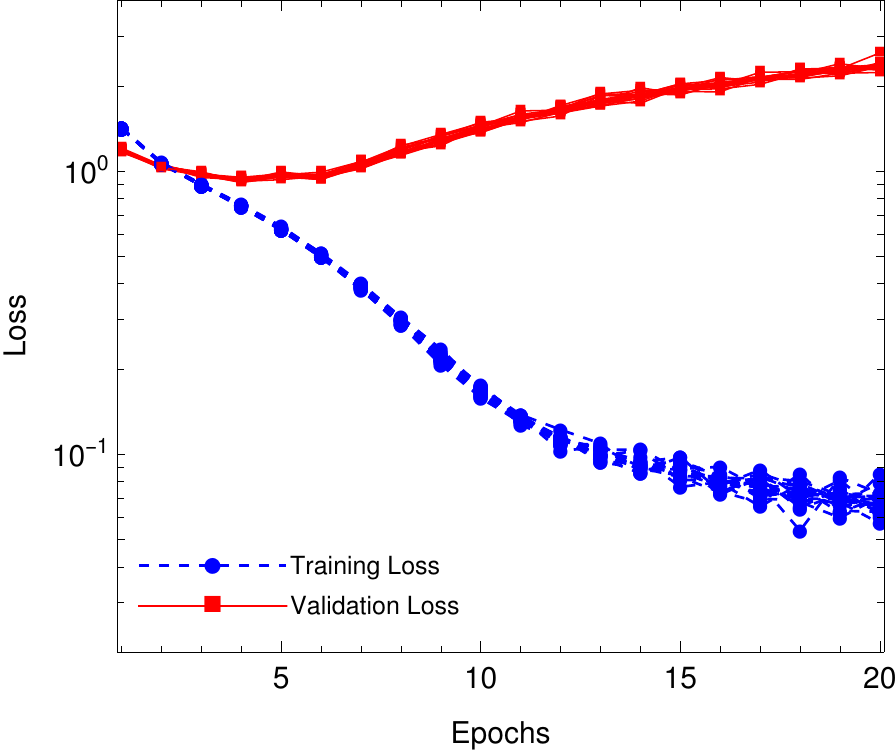}\label{SIFIG:LOSSRELU2}
    \caption{Plot of training/validation losses with built-in \acs{RELU} activation function}
\end{figure}

\begin{figure}[!tbph]
    \centering     
    \subfigure[\hspace{-25pt}]{\includegraphics[scale=0.85]{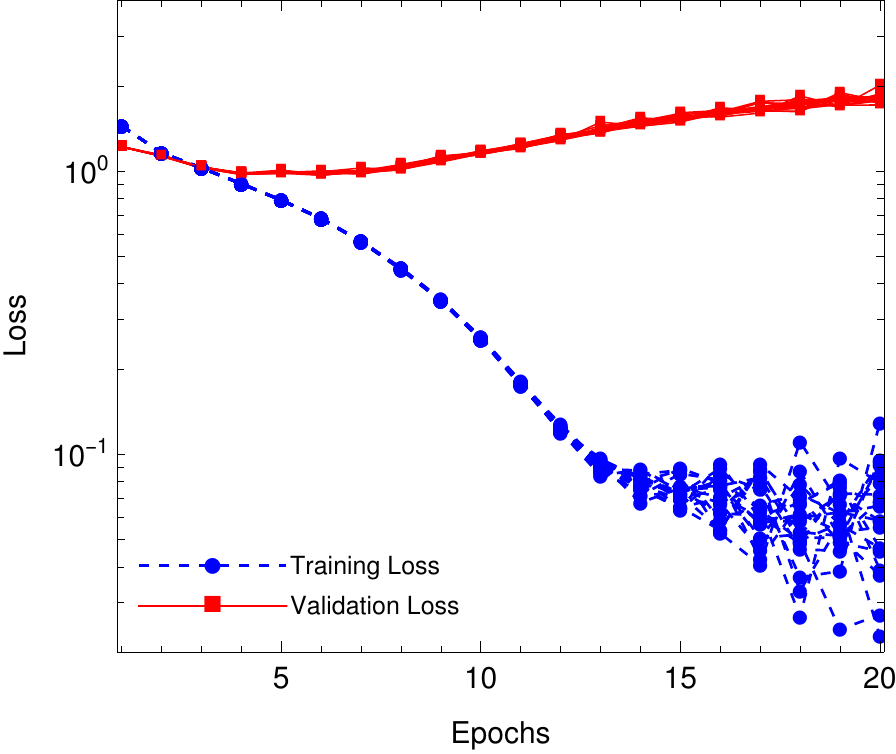}\label{SIFIG:LOSSTANH2}}
    \hspace{3pt}
    \subfigure[\hspace{-25pt}]{\includegraphics[scale=0.85]{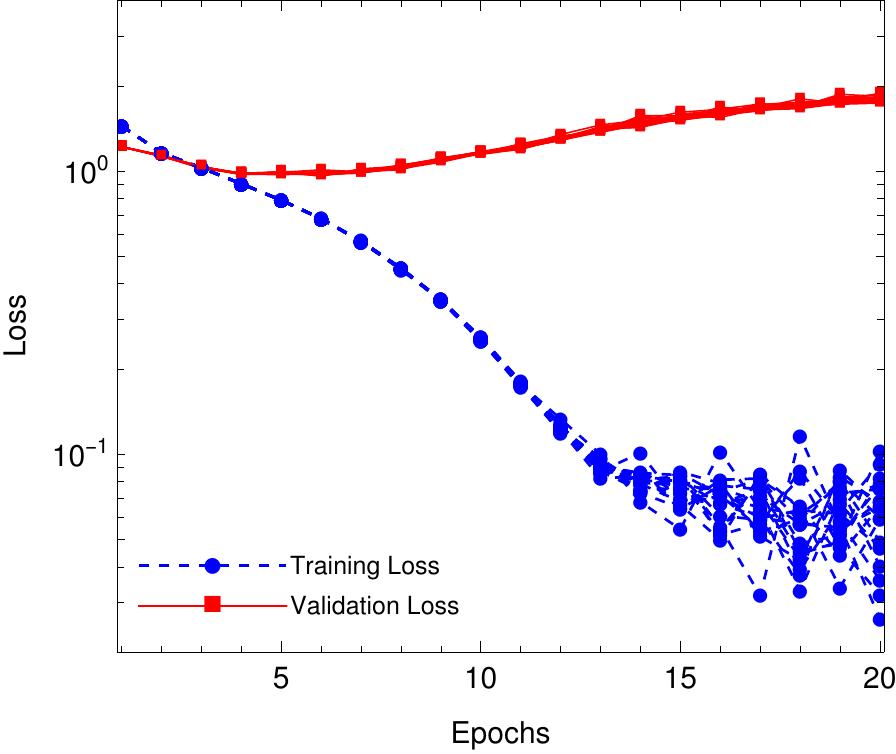}\label{SIFIG:LOSSFTANH2}} 
    \\
    \subfigure[\hspace{-25pt}]{\includegraphics[scale=0.85]{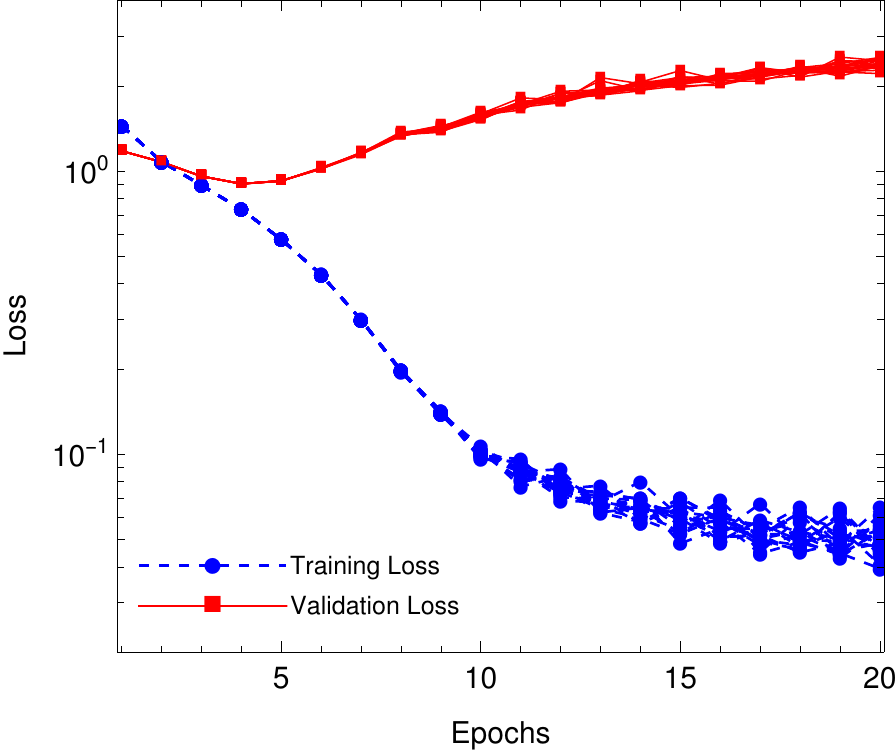}\label{SIFIG:LOSSSWISH12}}
    \hspace{3pt}
    \subfigure[\hspace{-25pt}]{\includegraphics[scale=0.85]{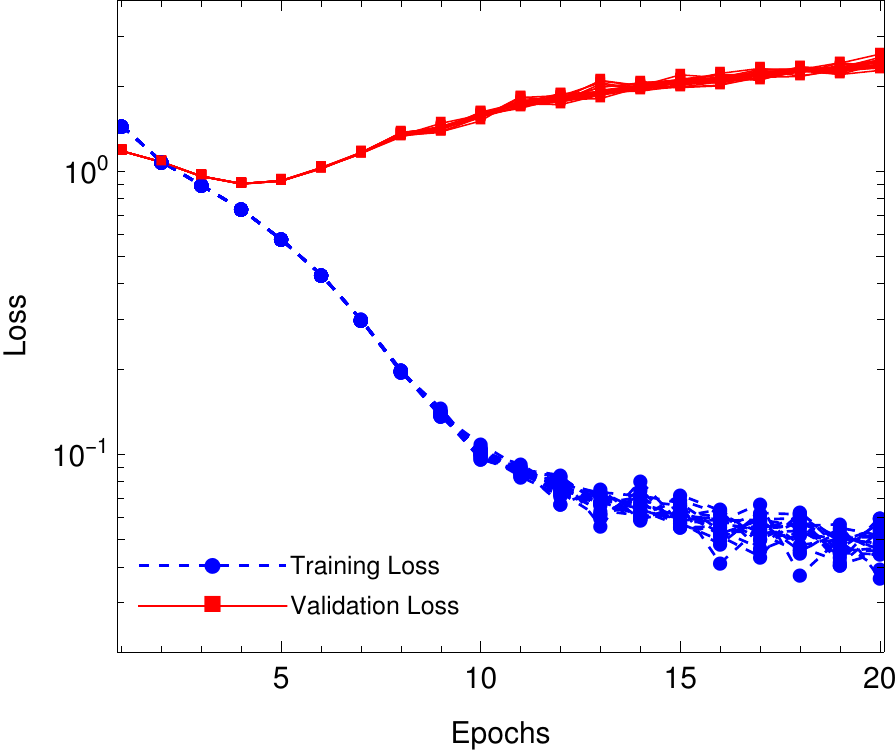}\label{SIFIG:LOSSFSWISH12}}
    \\
    \subfigure[\hspace{-25pt}]{\includegraphics[scale=0.85]{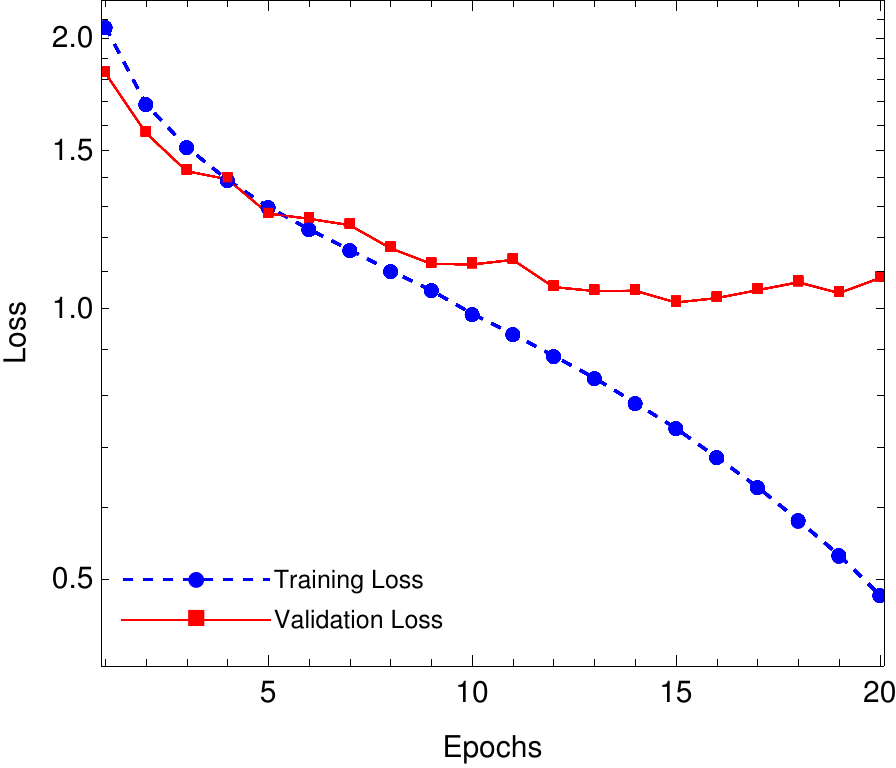}\label{SIFIG:LOSSSIGMOID2}}
    \hspace{3pt}
    \subfigure[\hspace{-25pt}]{\includegraphics[scale=0.85]{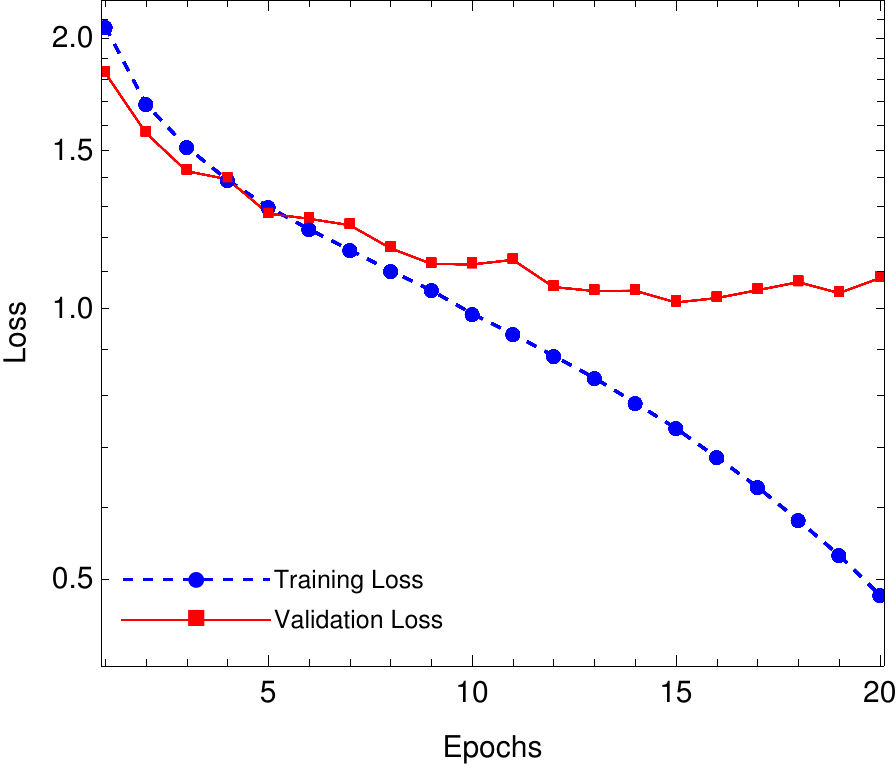}\label{SIFIG:LOSSFSIGMOID2}}
    \caption{Plot of training/validation losses with built-in (a), (c) and (e) and gated 
    representations (b), (d) and (f) of $\tanh$, Swish-1 and sigmoid activation functions, respectively}
\end{figure}

\begin{figure}[!tbph]
    \centering     
    \subfigure[\hspace{-25pt}]{\includegraphics[scale=0.85]{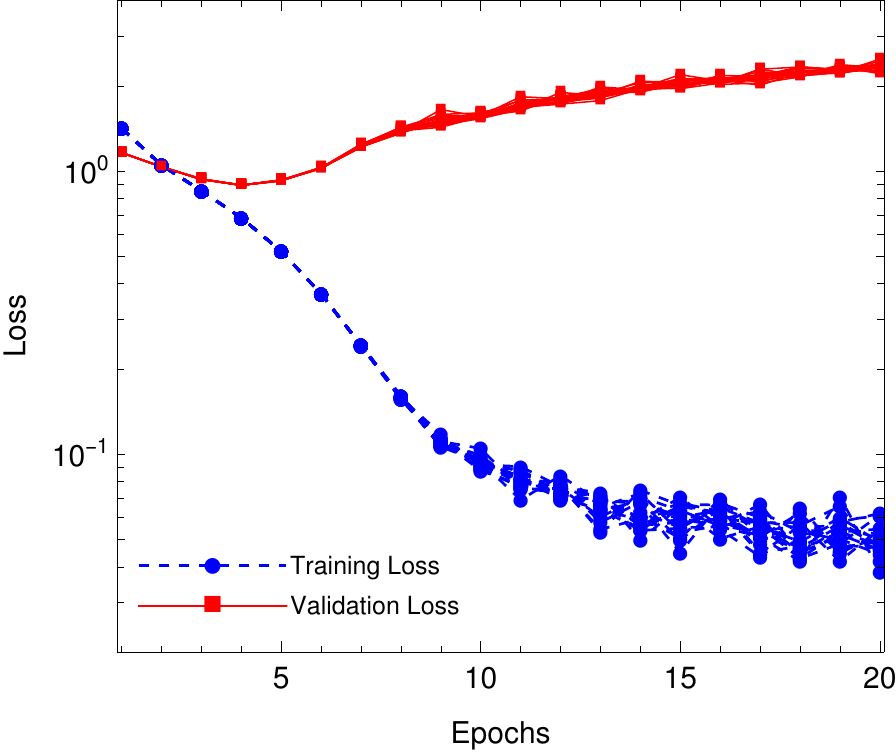}\label{SIFIG:LOSSMISH2}}
    \hspace{3pt}
    \subfigure[\hspace{-25pt}]{\includegraphics[scale=0.85]{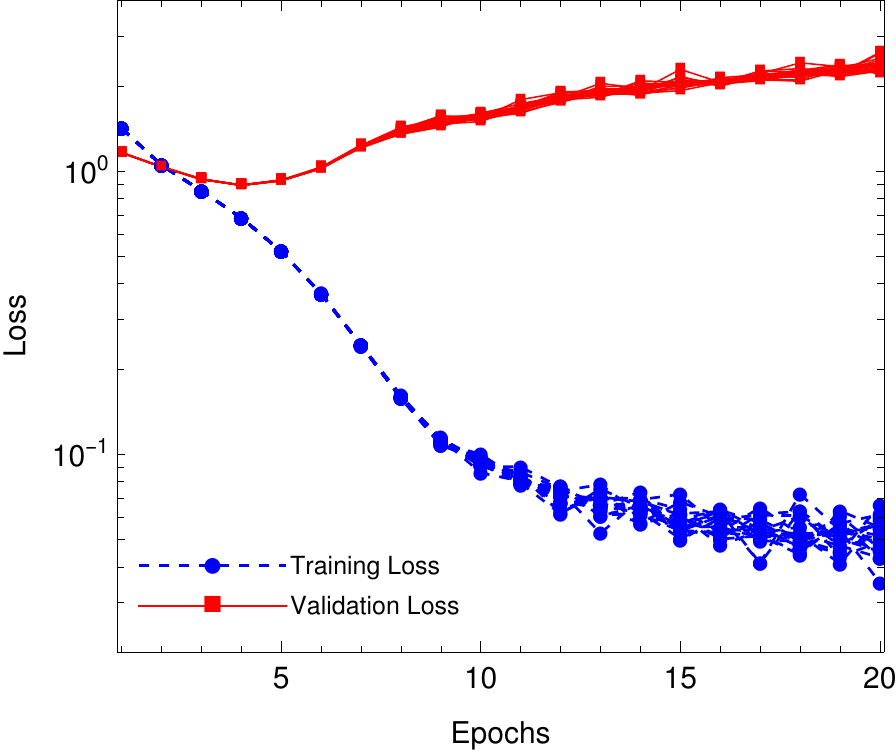}\label{SIFIG:LOSSFMISH2}}
    \\
    \subfigure[\hspace{-25pt}]{\includegraphics[scale=0.85]{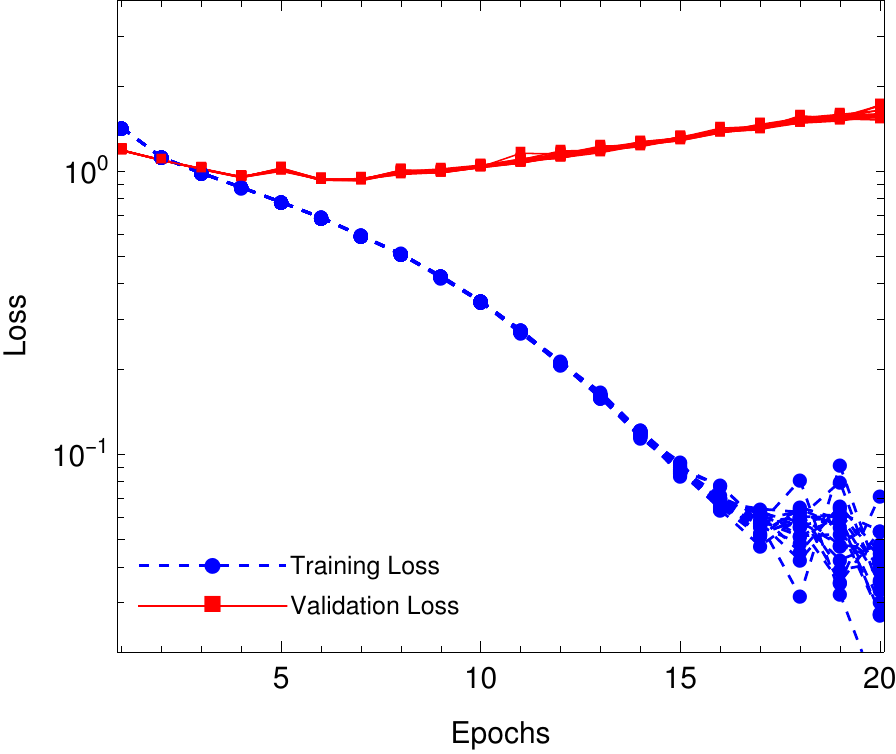}\label{SIFIG:LOSSSOFTSIGN2}}
    \hspace{3pt}
    \subfigure[\hspace{-25pt}]{\includegraphics[scale=0.85]{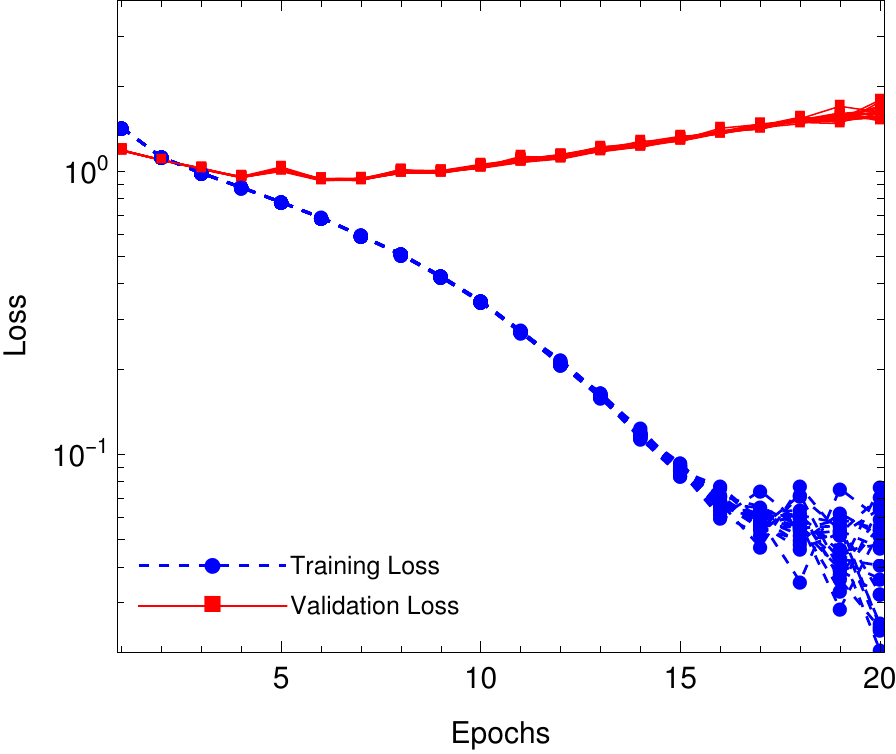}\label{SIFIG:LOSSFSOFTSIGN2}}
    \caption{Plot of training/validation losses with built-in (a) and (c) and gated representations (b) 
    and (d) of Mish and Softsign activation functions, respectively}
\end{figure}
\newpage \ \newpage
\section{Plots of Training and Validation Accuracies: CIFAR-10 Dataset}
In this section, we present plots of training and validation accuracies 
corresponding to the individual training sessions for training LeNet-5 neural network on 
the CIFAR-10 dataset. Each plot includes the accuracy values for twenty training experiments 
each running over 20 epochs using a \nvidia\ A100 80GB PCIe \ac{GPU}.

\begin{figure}[!tbph]
    \centering     
    \includegraphics[scale=0.85]{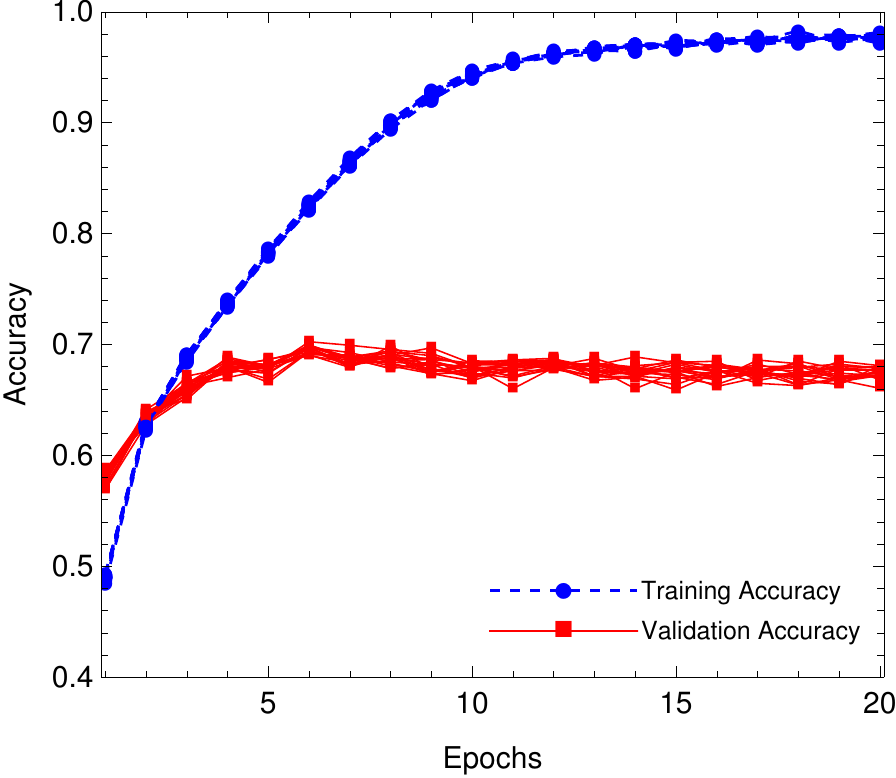}\label{SIFIG:ACCRELU2}
    \caption{Plot of training/validation accuracies with built-in \acs{RELU} activation function}
\end{figure}

\begin{figure}[!tbph]
    \centering     
    \subfigure[\hspace{-25pt}]{\includegraphics[scale=0.85]{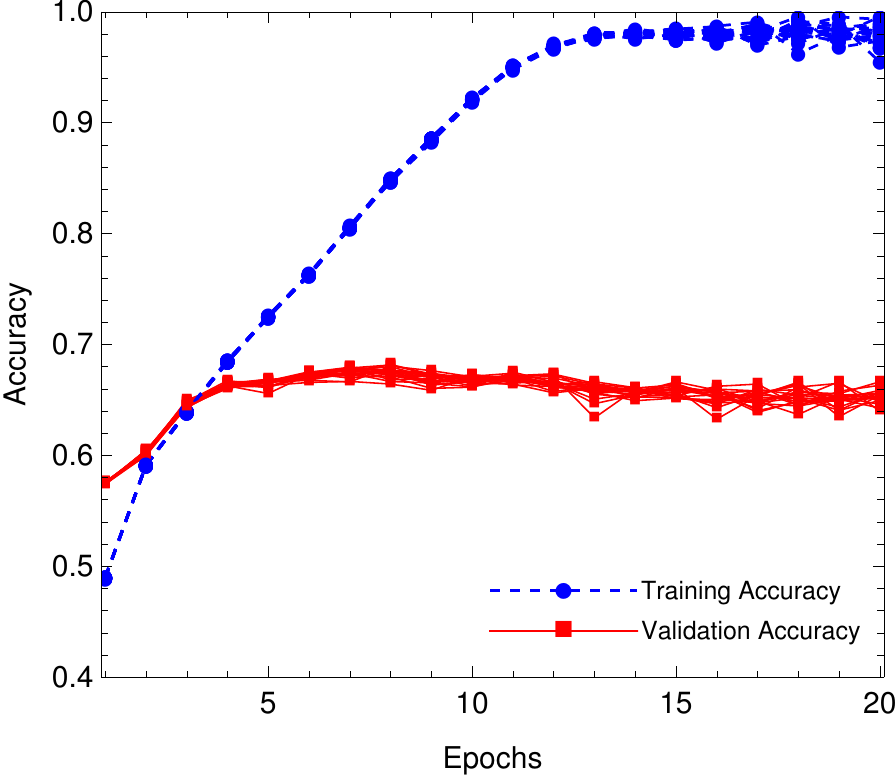}\label{SIFIG:ACCTANH2}} 
    \hspace{3pt}
    \subfigure[\hspace{-25pt}]{\includegraphics[scale=0.85]{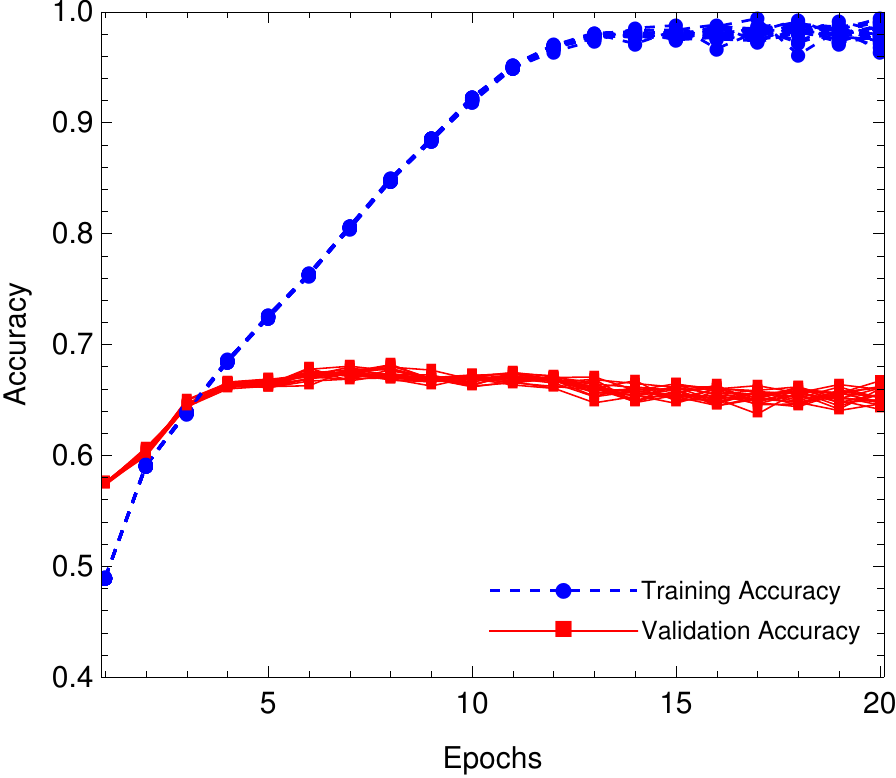}\label{SIFIG:ACCFTANH2}} 
    \\
    \subfigure[\hspace{-25pt}]{\includegraphics[scale=0.85]{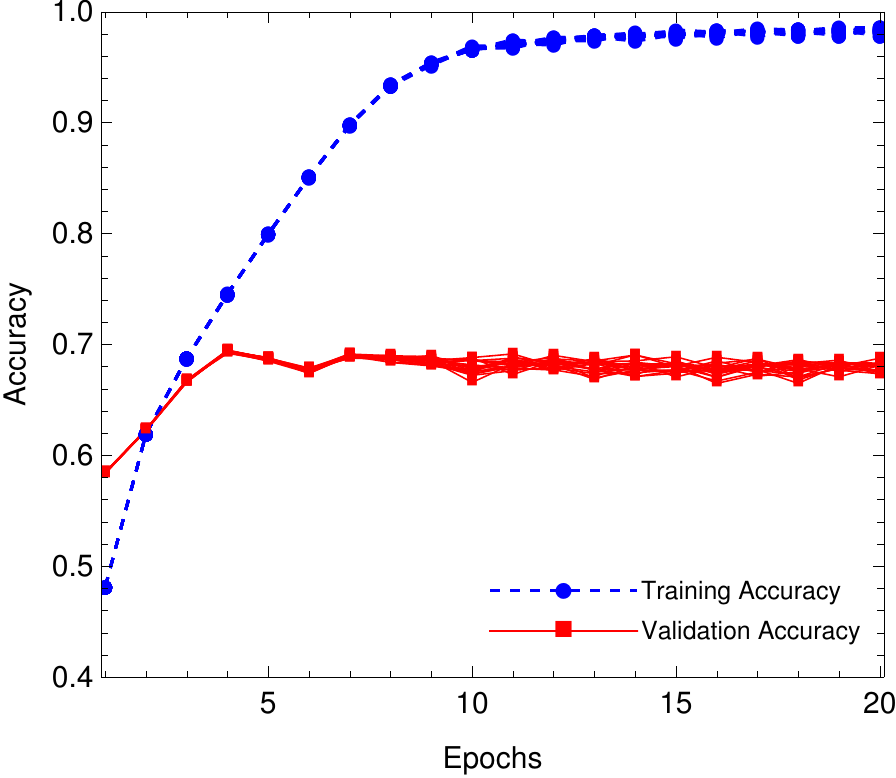}\label{SIFIG:ACCSWISH12}}
    \hspace{3pt}
    \subfigure[\hspace{-25pt}]{\includegraphics[scale=0.85]{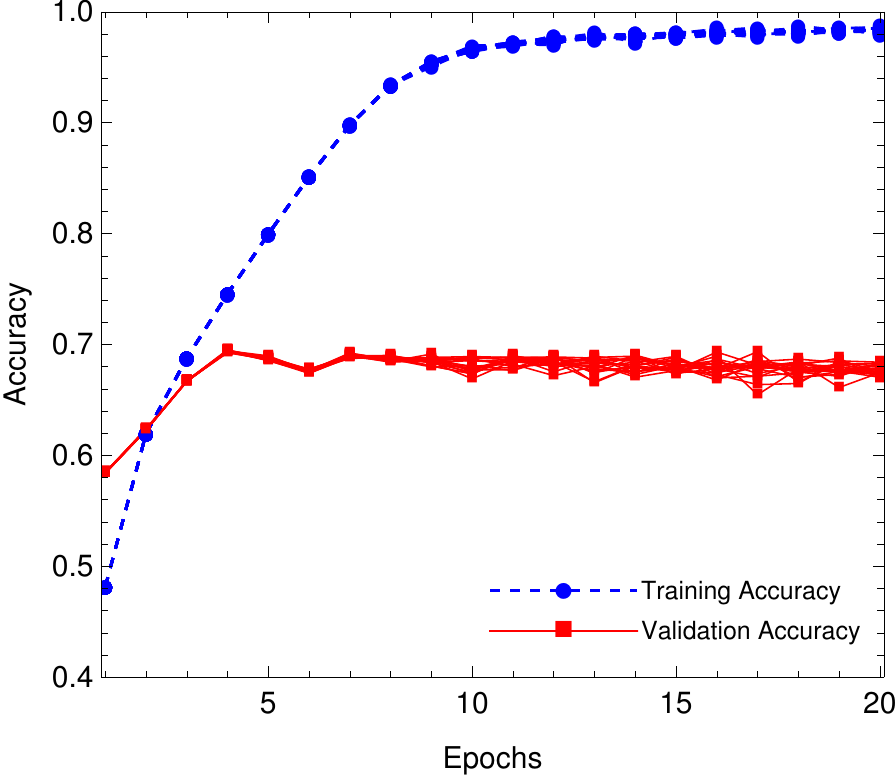}\label{SIFIG:ACCFSWISH12}}
    \\
    \subfigure[\hspace{-25pt}]{\includegraphics[scale=0.85]{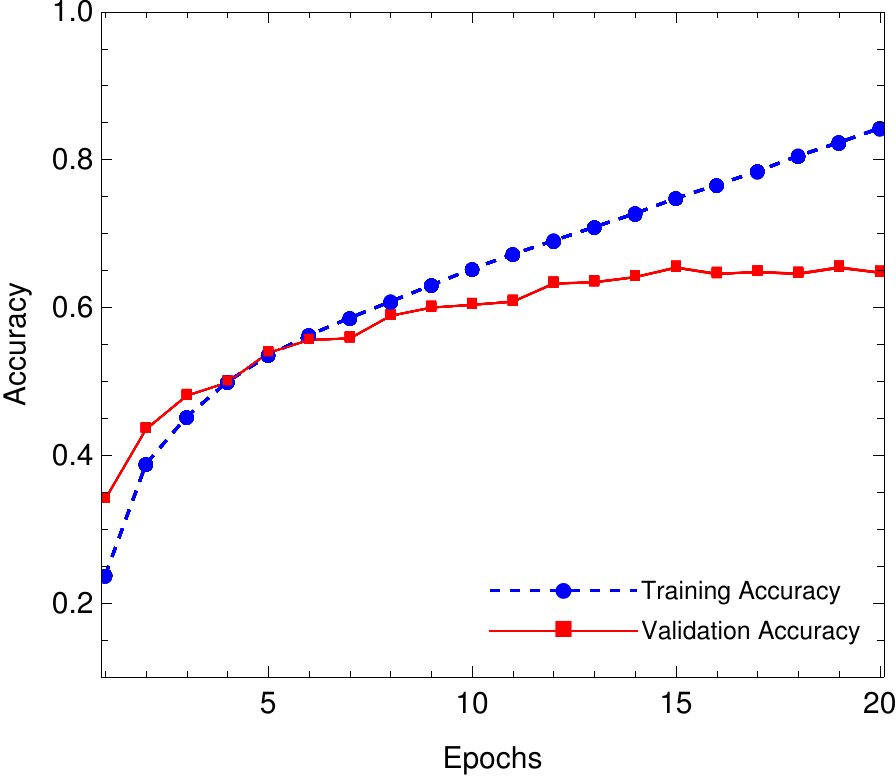}\label{SIFIG:ACCSIGMOID2}}
    \hspace{3pt}
    \subfigure[\hspace{-25pt}]{\includegraphics[scale=0.85]{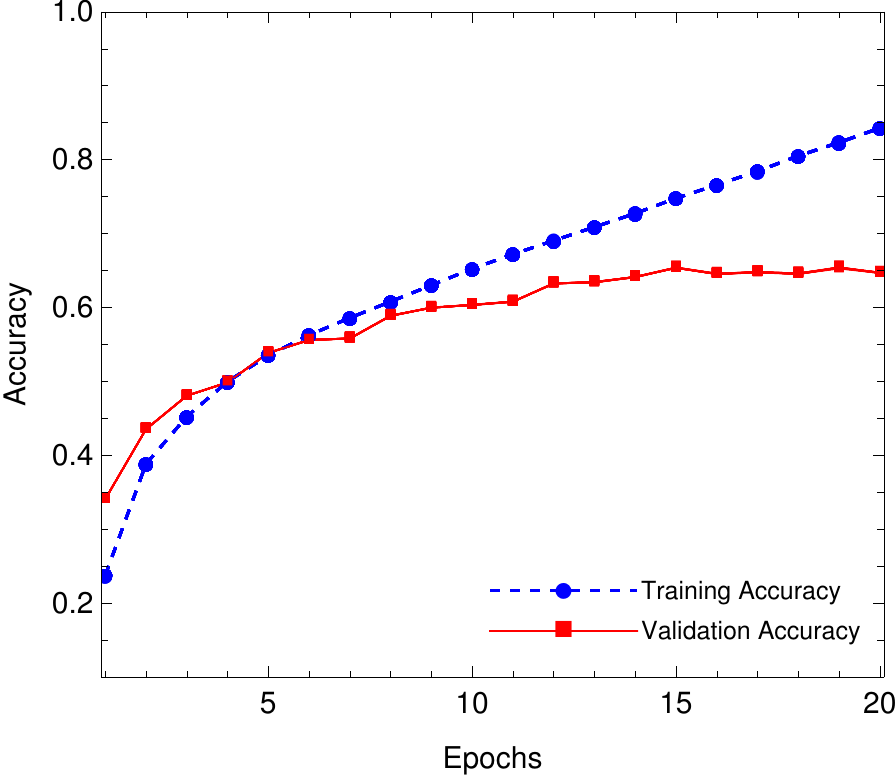}\label{SIFIG:ACCFSIGMOID2}}
    \caption{Plot of training/validation accuracies with built-in (a), (c) and (e) and gated 
    representations (b), (d) and (f) of $\tanh$, Swish-1 and sigmoid activation functions, respectively}
\end{figure}

\begin{figure}[!tbph]
    \centering     
    \subfigure[\hspace{-25pt}]{\includegraphics[scale=0.85]{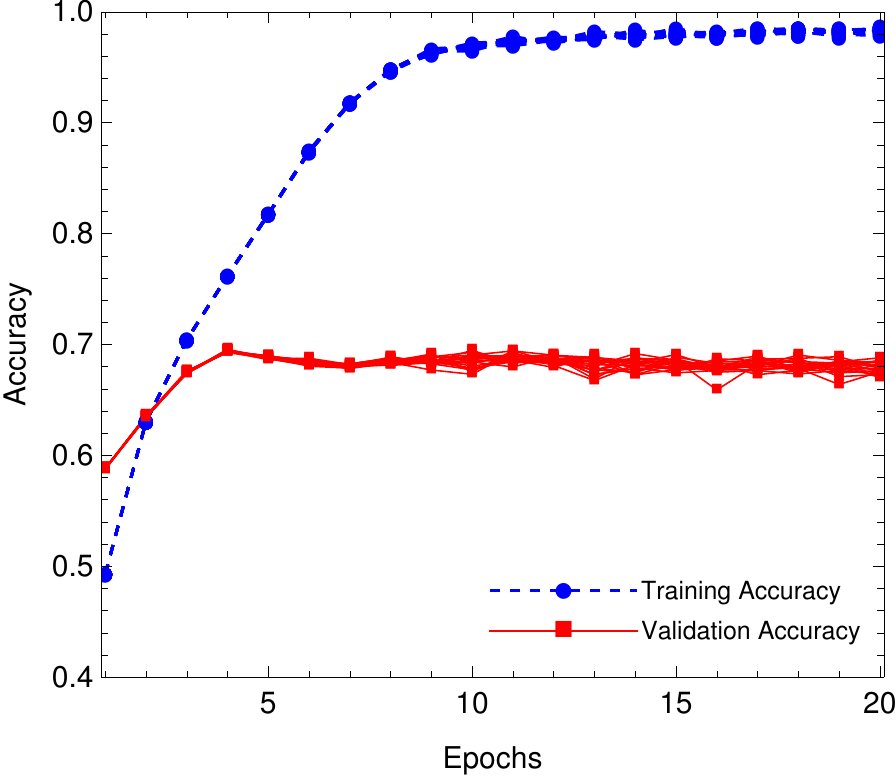}\label{SIFIG:ACCMISH2}}
    \hspace{3pt}
    \subfigure[\hspace{-25pt}]{\includegraphics[scale=0.85]{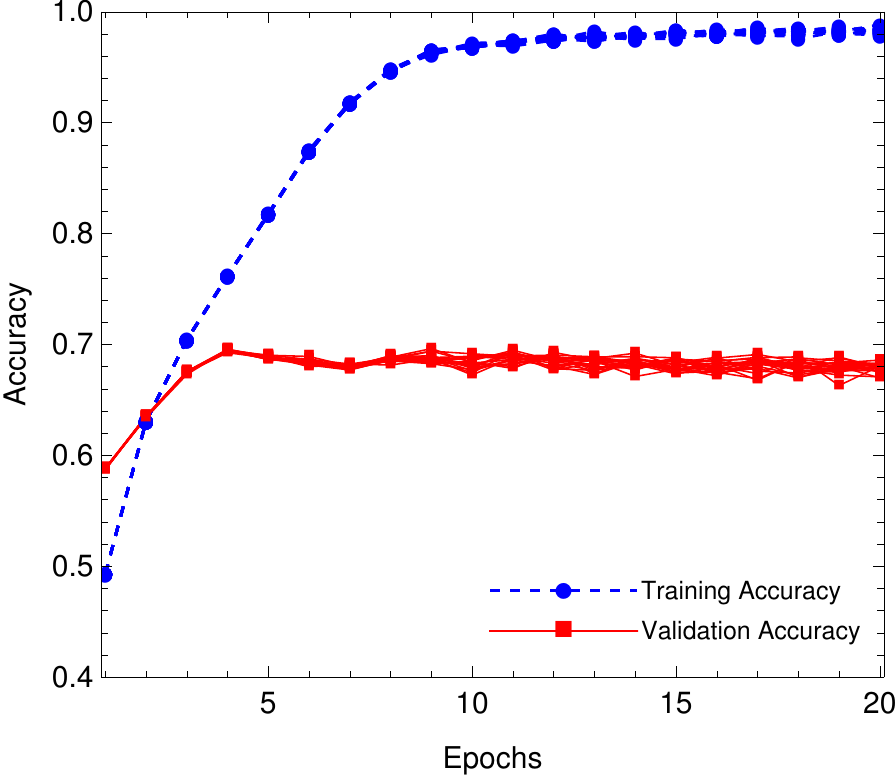}\label{SIFIG:ACCFMISH2}}
    \\
    \subfigure[\hspace{-25pt}]{\includegraphics[scale=0.85]{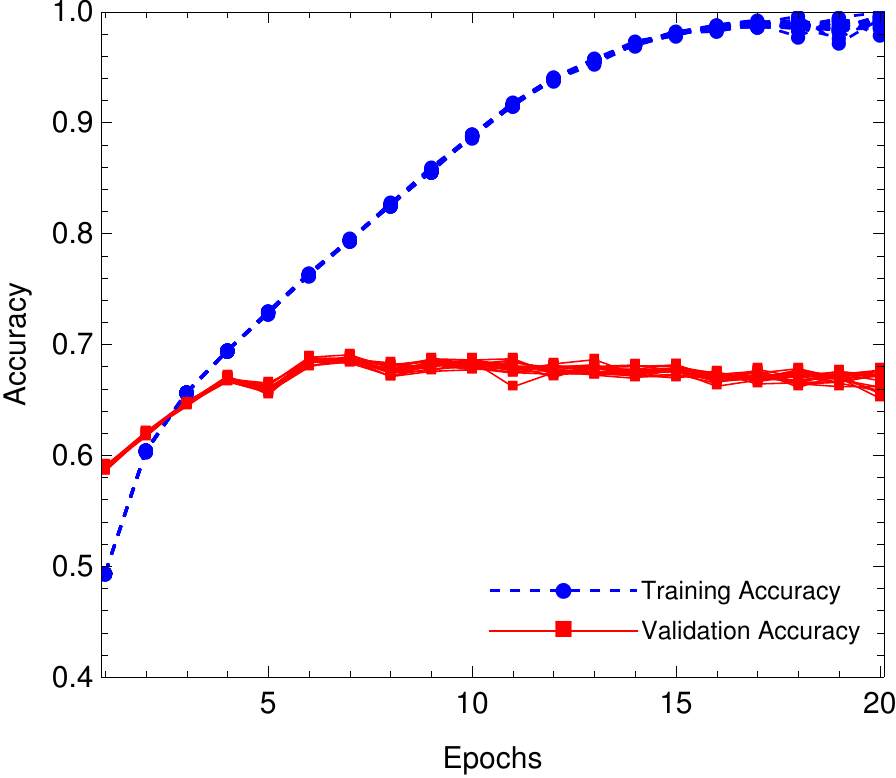}\label{SIFIG:ACCSOFTSIGN2}}
    \hspace{3pt}
    \subfigure[\hspace{-25pt}]{\includegraphics[scale=0.85]{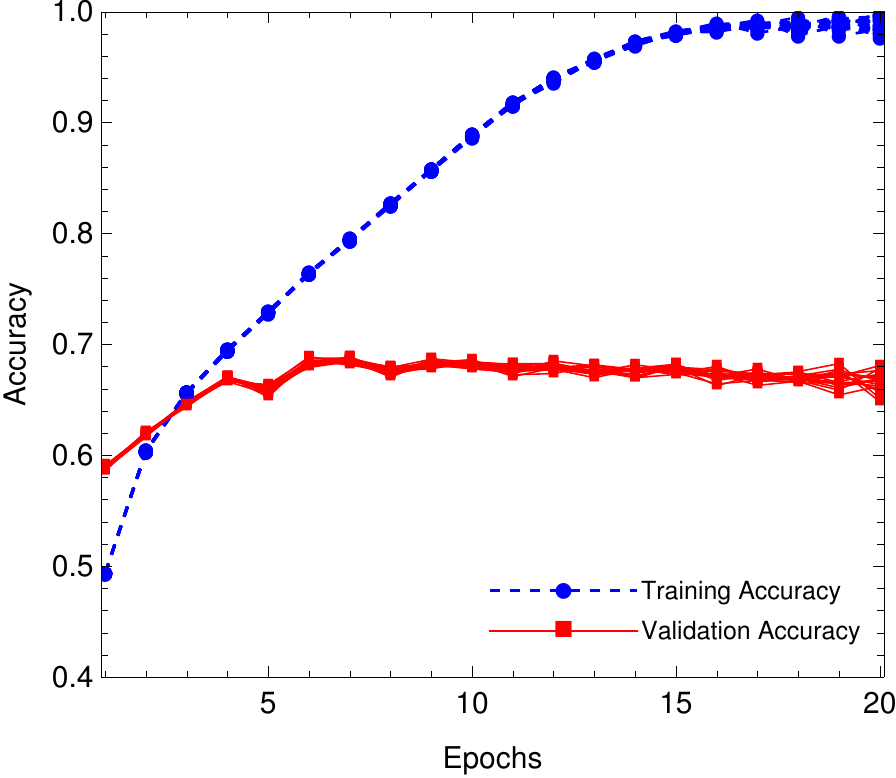}\label{SIFIG:ACCFSOFTSIGN2}}
    \caption{Plot of training/validation accuracies with built-in (a) and (c) and gated representations (b) 
    and (d) of Mish and Softsign activation functions, respectively}
\end{figure}

\begin{acronym}
    \acrodefplural{1-RDM}{one-electron \aclp{RDM}}
    \acrodefplural{2-RDM}{two-electron \aclp{RDM}}
    \acrodefplural{3-RDM}{three-electron \aclp{RDM}}
    \acrodefplural{4-RDM}{four-electron \aclp{RDM}}
    \acrodefplural{RDM}{reduced density matrices}
    \acro{1-HRDM}{one-hole \acl{RDM}}
    \acro{1-RDM}{one-electron \acl{RDM}}
    \acrodef{1H-PDFT}{one-parameter hybrid \acl{PDFT}}
    \acro{2-HRDM}{two-hole \acl{RDM}}
    \acro{2-RDM}{two-electron \acl{RDM}}
    \acro{3-RDM}{three-electron \acl{RDM}}
    \acro{4-RDM}{four-electron \acl{RDM}}
    \acro{ACI}{adaptive \acl{CI}}
    \acro{ACI-DSRG-MRPT2}{\acl{ACI}-\acl{DSRG} \acl{MR} \acl{PT2}}
    \acro{ACSE}{anti-Hermitian \acl{CSE}}
    \acro{ADAM}{adaptive momentum}
    \acro{AI}{artificial intelligence}
    \acrodef{AKEEPE}[$\AKEEe$]{absolute kinetic energy error per electron}
    \acrodef{AKEE}{absolute kinetic energy error}
    \acro{ANN}{artificial \acl{NN}}
    \acro{AO}{atomic orbital}
    \acro{AQCC}{averaged quadratic \acl{CC}}
    \acro{ATAC}{attentional activation}
    \acro{aug-cc-pVQZ}{augmented correlation-consistent polarized-valence quadruple-$\ze$}
    \acro{aug-cc-pVTZ}{augmented correlation-consistent polarized-valence triple-$\ze$}
    \acro{aug-cc-pwCV5Z}[aug-cc-pwCV5Z]{augmented correlation-consistent polarized weighted core-valence quintuple-$\ze$}
    \acro{B3LYP}{Becke-3-\acl{LYP}}
    \acro{BLA}{bond length alternation}
    \acro{BLYP}{Becke and \acl{LYP}}
    \acro{BO}{Born-Oppenheimer}
    \acro{BP86}{Becke 88 exchange and P86 Perdew-Wang correlation}
    \acro{BPSDP}{boundary-point \acl{SDP}}
    \acro{CAM}{Coulomb-attenuating method}
    \acro{CAM-B3LYP}{Coulomb-attenuating method \acl{B3LYP}}
    \acro{CAS-PDFT}{\acl{CAS} \acl{PDFT}}
    \acro{CASPT2}{\acl{CAS} \acl{PT2}}
    \acro{CASSCF}{\acl{CAS} \acl{SCF}}
    \acro{CAS}{complete active-space}
    \acro{cc-pVDZ}{correlation-consistent polarized-valence double-$\ze$}
    \acro{cc-pVTZ}{correlation-consistent polarized-valence triple-$\ze$}
    \acro{CCSDT}{coupled-cluster, singles doubles and triples}
    \acro{CCSD}{coupled-cluster with singles and doubles}
    \acro{CC}{coupled-cluster}
    \acro{CI}{configuration interaction}
    \acro{CNN}{convolutional \acl{NN}}
    \acro{CO}{constant-order}
    \acro{CPO}{correlated participating orbitals}
    \acro{CRELU}[CReLU]{concatrenated \acl{RELU}}
    \acro{CSF}{configuration state function}
    \acro{CS-KSDFT}{\acl{CS}-\acl{KSDFT}}
    \acro{CSE}{contracted \acl{SE}}
    \acro{CS}{constrained search}
    \acro{DC-DFT}{density corrected-\acl{DFT}}
    \acro{DC-KDFT}{density corrected-\acl{KDFT}}
    \acro{DE}{delocalization error}
    \acro{DFT}{density functional theory}
    \acro{DF}{density-fitting}
    \acro{DIIS}{direct inversion in the iterative subspace}
    \acro{DMRG}{density matrix renormalization group}
    \acro{DNN}{deep \acl{NN}}
    \acro{DOCI}{doubly occupied \acl{CI}}
    \acro{DSRG}{driven similarity renormalization group}
    \acro{EKT}{extended Koopmans theorem}
    \acro{ELU}{exponential linear unit}
    \acro{ERI}{electron-repulsion integral}
    \acro{EUE}{effectively unpaired electron}
    \acro{FC}{fractional calculus}
    \acro{FCI}{full \acl{CI}}
    \acro{FP-1}{frontier partition with one set of interspace excitations}
    \acro{FRELU}[FReLU]{flexible \acl{RELU}}
    \acro{FSE}{fractional \acl{SE}}
    \acro{ftBLYP}{fully \acl{tBLYP}}
    \acro{ftPBE}{fully \acl{tPBE}}
    \acro{ftSVWN3}{fully \acl{tSVWN3}}
    \acro{ft}{full translation}
    \acro{GASSCF}{generalized active-space \acl{SCF}}
    \acro{GELU}{Gaussian error linear unit}
    \acro{GGA}{generalized gradient approximation}
    \acro{GL}{Gr\"unwald-Letnikov}
    \acro{GMCPDFT}[G-MC-PDFT]{generalized \acl{MCPDFT}}
    \acro{GPU}{graphics processing unit}
    \acro{GTO}{Gaussian-type orbital}
    \acro{HF}{Hartree-Fock}
    \acro{HISS}{Henderson-Izmaylov-Scuseria-Savin}
    \acrodef{HK}{Hohenberg-Kohn}
    \acro{HOMO}{highest-occupied \acl{MO}}
    \acro{HONO}{highest-occupied \acl{NO}}
    \acro{HPDFT}{hybrid \acl{PDFT}}
    \acro{HRDM}{hole \acl{RDM}}
    \acro{HSE}{Heyd-Scuseria-Ernzerhof}
    \acro{HXC}{Hartree-\acl{XC} }
    \acro{IPEA}{ionization potential electron affinity}
    \acro{IPSDP}{interior-point \acl{SDP}}
    \acro{KDFT}{kinetic \acl{DFT}}
    \acro{KSDFT}[KS-DFT]{\acl{KS} \acl{DFT}}
    \acro{KS}{Kohn-Sham}
    \acro{LBFGS}[L-BFGS]{limited-memory Broyden-Fletcher-Goldfarb-Shanno}
    \acro{LC}{long-range corrected}
    \acro{LC-VV10}{\acl{LC} Vydrov-van Voorhis 10}
    \acro{l-DFVB}[$\la$-DFVB]{$\la$-density functional \acl{VB}}
    \acro{LEB}{local energy balance}
    \acro{LE}{localization error}
    \acrodef{lftBLYP}[$\la$-\acs{ftBLYP}]{$\la$-\acl{ftBLYP}}
    \acrodef{lftPBE}[$\la$-\acs{ftPBE}]{$\la$-\acl{ftPBE}}
    \acrodef{lftrevPBE}[$\la$-\acs{ftrevPBE}]{$\la$-\acl{ftrevPBE}}
    \acrodef{lftSVWN3}[$\la$-\acs{ftSVWN3}]{$\la$-\acl{ftSVWN3}}
    \acro{lMCPDFT}[$\la$-MC-PDFT]{\acl{MC} \acl{1H-PDFT}}
    \acro{LMF}{local mixing function}
    \acro{LO}{Lieb-Oxford}
    \acro{LP}{linear programming}
    \acro{LR}{long-range}
    \acro{LRELU}[LReLU]{leaky \acl{RELU}}
    \acro{LSDA}{local spin-density approximation}
    \acrodef{ltBLYP}[$\la$-\acs{tBLYP}]{$\la$-\acl{tBLYP}}
    \acrodef{ltPBE}[$\la$-\acs{tPBE}]{$\la$-\acl{tPBE}}
    \acrodef{ltrevPBE}[$\la$-\acs{trevPBE}]{$\la$-\acl{trevPBE}}
    \acrodef{ltSVWN3}[$\la$-\acs{tSVWN3}]{$\la$-\acl{tSVWN3}}
    \acro{LUMO}{lowest-unoccupied \acl{MO}}
    \acro{LUNO}{lowest-unoccupied \acl{NO}}
    \acro{LYP}{Lee-Yang-Parr}
    \acro{M06}{Minnesota 06}
    \acro{M06-2X}{\acl{M06} with double non-local exchange}
    \acro{M06-L}{\acl{M06} local}
    \acro{MAEPE}[$\MAEe$]{\acl{MAE} per electron}
    \acro{MAE}{mean absolute error}
    \acro{MAKEE}[$\MAKEE$]{mean absolute kinetic energy error per electron}
    \acro{MAX}{maximum absolute error}
    \acro{MC1H-PDFT}{\acl{MC} \acl{1H-PDFT}}
    \acro{MC1H}{\acl{MC} one-parameter hybrid \acl{PDFT}}
    \acro{MCHPDFT}{\acl{MC} hybrid-\acl{PDFT}}
    \acro{MCPDFT}[MC-PDFT]{\acl{MC} \acl{PDFT}}
    \acro{MCRSHPDFT}[$\mu\la$-MCPDFT]{\acl{MC} range-separated hybrid-\acl{PDFT}}
    \acro{MCSCF}{\acl{MC} \acl{SCF}}
    \acrodef{MC}{multiconfiguration}
    \acro{ML}{machine learning}
    \acro{MN15}{Minnesota 15}
    \acro{MNIST}{Modified National Institute of Standards and Technology}
    \acro{MOLSSI}[MolSSI]{Molecular Sciences Software Institute}
    \acro{MO}{molecular orbital}
    \acro{MP2}{second-order M\o ller-Plesset \acl{PT}}
    \acro{MR-AQCC}{\acl{MR}-averaged quadratic \acl{CC}}
    \acrodef{MR}{multireference}
    \acro{MS0}{MS0 meta-GGA exchange and revTPSS GGA correlation}
    \acro{NGA}{nonseparable gradient approximation}
    \acro{NIAD}{normed integral absolute deviation}
    \acro{NLRELU}[NLReLU]{natural logarithm \acl{RELU}}
    \acro{NN}{neural network}
    \acro{NO}{natural orbital}
    \acro{NOON}{\acl{NO} \acl{ON}}
    \acro{NPE}{non-parallelity error}
    \acro{NSF}{National Science Foundation}
    \acro{OEP}{optimized effective potential}
    \acro{oMCPDFT}[$\om$-MC-PDFT]{range-separated \acl{MC} \acl{1H-PDFT}}
    \acrodef{ON}{occupation number}
    \acro{ORMAS}{occupation-restricted multiple active-space}
    \acro{OTPD}{on-top pair-density}
    \acro{PBE0}{hybrid-\acs{PBE}}
    \acro{PBE}{Perdew-Burke-Ernzerhof}
    \acro{pCCD-lDFT}[pCCD-$\la$DFT]{\acl{pCCD} $\la$\acs{DFT}}
    \acro{pCCD}{pair coupled-cluster doubles}
    \acro{PDFT}{pair-\acl{DFT}}
    \acro{PEC}{potential energy curve}
    \acro{PES}{potential energy surface}
    \acro{PKZB}{Perdew-Kurth-Zupan-Blaha}
    \acro{pp-RPA}{particle-particle \acl{RPA}}
    \acro{PRELU}[PReLU]{parametric \acl{RELU}}
    \acro{PT}{perturbation theory}
    \acro{PT2}{second-order \acl{PT}}
    \acro{PW91}{Perdew-Wang 91}
    \acro{QTAIM}{quantum theory of atoms in molecules}
    \acro{RASSCF}{restricted active-space \acl{SCF}}
    \acro{RBM}{restricted Boltzmann machine}
    \acro{RDM}{reduced density matrix}
    \acro{RELU}[ReLU]{rectified linear unit}
    \acro{revPBE}{revised \acs{PBE}}
    \acro{RL}{Riemann-Liouville}
    \acro{RMSD}{root mean square deviation}
    \acro{RNN}{recurrent \acl{NN}}
    \acro{RPA}{random-phase approximation}
    \acro{RRELU}[RReLU]{randomized \acl{RELU}}
    \acro{RSH}{range-separated hybrid}
    \acro{SCAN}{strongly constrained and appropriately normed}
    \acro{SCF}{self-consistent field}
    \acro{SDP}{semidefinite programming}
    \acro{SE}{Schr\"odinger equation}
    \acro{SELU}{scaled \acl{ELU}}
    \acro{SF-CCSD}{\acl{SF}-\acl{CCSD}}
    \acro{SF}{spin-flip}
    \acro{SGD}{stochastic gradient descent}
    \acro{SIE}{self-interaction error}
    \acrodef{SI}{Supporting Information}
    \acro{SNIAD}{spherical \acl{NIAD}}
    \acro{SOGGA11}{second-order \acl{GGA}}
    \acro{SR}{short-range}
    \acro{SRELU}[SReLU]{S-shaped \acl{RELU}}
    \acro{STO}{Slater-type orbital}
    \acro{SVWN3}{Slater and Vosko-Wilk-Nusair random-phase approximation expression III}
    \acro{tBLYP}{translated \acl{BLYP}}
    \acro{TMAE}[$\MAE$]{total \acl{MAE} per electron}
    \acro{TNIAD}[$\NIAD$]{total normed integral absolute deviation}
    \acro{tPBE}{translated \acl{PBE}}
    \acro{TPSS}{Tao-Perdew-Staroverov-Scuseria}
    \acro{trevPBE}{translated \acs{revPBE}}
    \acro{tr}{conventional translation}
    \acro{tSVWN3}{translated \acl{SVWN3}}
    \acro{TS}{transition state}
    \acro{v2RDM-CASSCF-PDFT}{\acl{v2RDM} \acl{CASSCF} \acl{PDFT}}
    \acro{v2RDM-CASSCF}{\acl{v2RDM}-driven \acl{CASSCF}}
    \acro{v2RDM-CAS}{\acl{v2RDM}-driven \acl{CAS}}
    \acro{v2RDM-DOCI}{\acl{v2RDM}-\acl{DOCI}}
    \acro{v2RDM}{variational \acl{2-RDM}}
    \acro{VB}{valence bond}
    \acro{VO}{variable-order}
    \acro{wB97X}[$\omega$B97X]{$\omega$B97X}
    \acro{WFT}{wave function theory}
    \acro{WF}{wave function}
    \acro{XC}{exchange-correlation}
    \acro{ZPE}{zero-point energy}
    \acro{ZPVE}{zero-point vibrational energy}
\end{acronym}


